%% file: egpaper_for_review.tex
\documentclass[10pt,twocolumn,letterpaper]{article}

\usepackage{cvpr}
\usepackage{times}
\usepackage{epsfig}
\usepackage{graphicx}
\usepackage{amsmath}
\usepackage{amssymb}

\usepackage[utf8]{inputenc} %
\usepackage[T1]{fontenc}    %
\usepackage{url}            %
\usepackage{booktabs}       %
\usepackage{amsfonts}       %
\usepackage{nicefrac}       %
\usepackage{microtype}      %
\usepackage{multirow}
\usepackage{graphicx}
\usepackage{subfig}
\usepackage{floatrow}
\usepackage{adjustbox}

\usepackage{amsmath,amssymb} %
\def\bal#1\eal{\begin{align}#1\end{align}} %

\def\transp{\mathsf{T}} %
\def\m{\mathbf}

\DeclareMathOperator*{\argmin}{arg\,min}

\newcommand {\bbmtx}{\begin{bmatrix}} %
\newcommand {\ebmtx}{\end{bmatrix}} %

\DeclareMathOperator{\vect}{vec}
\newcommand{\vectorize}[1]{\ensuremath{\vect(#1)}}

\usepackage[normalem]{ulem}
\useunder{\uline}{\ul}{}

\usepackage[pagebackref=true,breaklinks=true,letterpaper=true,colorlinks,bookmarks=false]{hyperref}

\cvprfinalcopy %

\def\cvprPaperID{4316} %

\begin{document}

\title{Learning  monocular 3D reconstruction of articulated categories from motion}

\author{Filippos Kokkinos \\
University College London\\
{\tt\small filippos.kokkinos@ucl.ac.uk} \\
\and
Iasonas Kokkinos \\
University College London, Snap Inc. \\
{\tt\small i.kokkinos@cs.ucl.ac.uk}
}

\maketitle

\begin{abstract}
Monocular 3D reconstruction of articulated object categories is challenging due to the lack of training data and the inherent ill-posedness of the problem.
In this work we use video self-supervision, forcing the consistency of consecutive 3D reconstructions by a motion-based cycle loss. This largely improves both optimization-based and learning-based 3D mesh reconstruction. We further introduce an interpretable model of 3D template deformations that controls a 3D surface through the displacement of a small number of local, learnable handles. We formulate this operation as a structured layer relying on mesh-laplacian regularization and show that it can be trained in an end-to-end manner. 
We finally introduce a per-sample numerical optimisation approach that jointly optimises over mesh displacements and cameras within a video, boosting accuracy both for training and also as test time post-processing. 

While relying exclusively on a small set of videos collected per category for supervision, we obtain state-of-the-art reconstructions with diverse shapes, viewpoints and textures for multiple articulated object categories. Supplementary materials, code, and videos are provided on the project page: \url{https://fkokkinos.github.io/video_3d_reconstruction/}.

\end{abstract}

\newcommand{\mycomment}[1]{}

\section{Introduction}
Monocular 3D reconstruction of general articulated categories is a task that humans perform routinely, but remains challenging for current computer vision systems. 
The breakthroughs achieved for humans \cite{smplify,kanazawa_humans,holopose,monocap,SMPL-X:2019,kolotouros2019spin,texturepose,eft,Choutas20}  have relied on expressive articulated shape priors \cite{LoperM0PB15} and mocap recordings to provide strong supervision in the form of 3D joint locations. Still, for general articulated categories, such as horses or cows, 
the problem remains in its infancy due to both the lack of strong supervision \cite{ZuffiKBB19} and the inherent challenge of representing and learning articulated deformations for general categories. 

Recent works have started tackling this problem by relying on minimal, 2D-based supervision  such as manual keypoint annotations or masks \cite{vicente2014reconstructing} and learning morphable model priors~\cite{vicente2014reconstructing,kar2015category,cmrKanazawa18, ucmrGoel20} or hand-crafted mesh segmentations~\cite{kulkarni2019csm}. 
In this work we leverage the rich information available in videos, and use networks trained for the 2D tasks of object detection, semantic segmentation, and optical flow to complement (optional) 2D keypoint-level supervision. 
 
We make three contributions towards pushing the envelope of monocular 3D object category reconstruction, by injecting ideas from structure-from-motion (SFM), geometry processing and bundle adjustment in the task of monocular 3D articulated reconstruction.

\begin{figure}
    \centering
    \includegraphics[width=\textwidth]{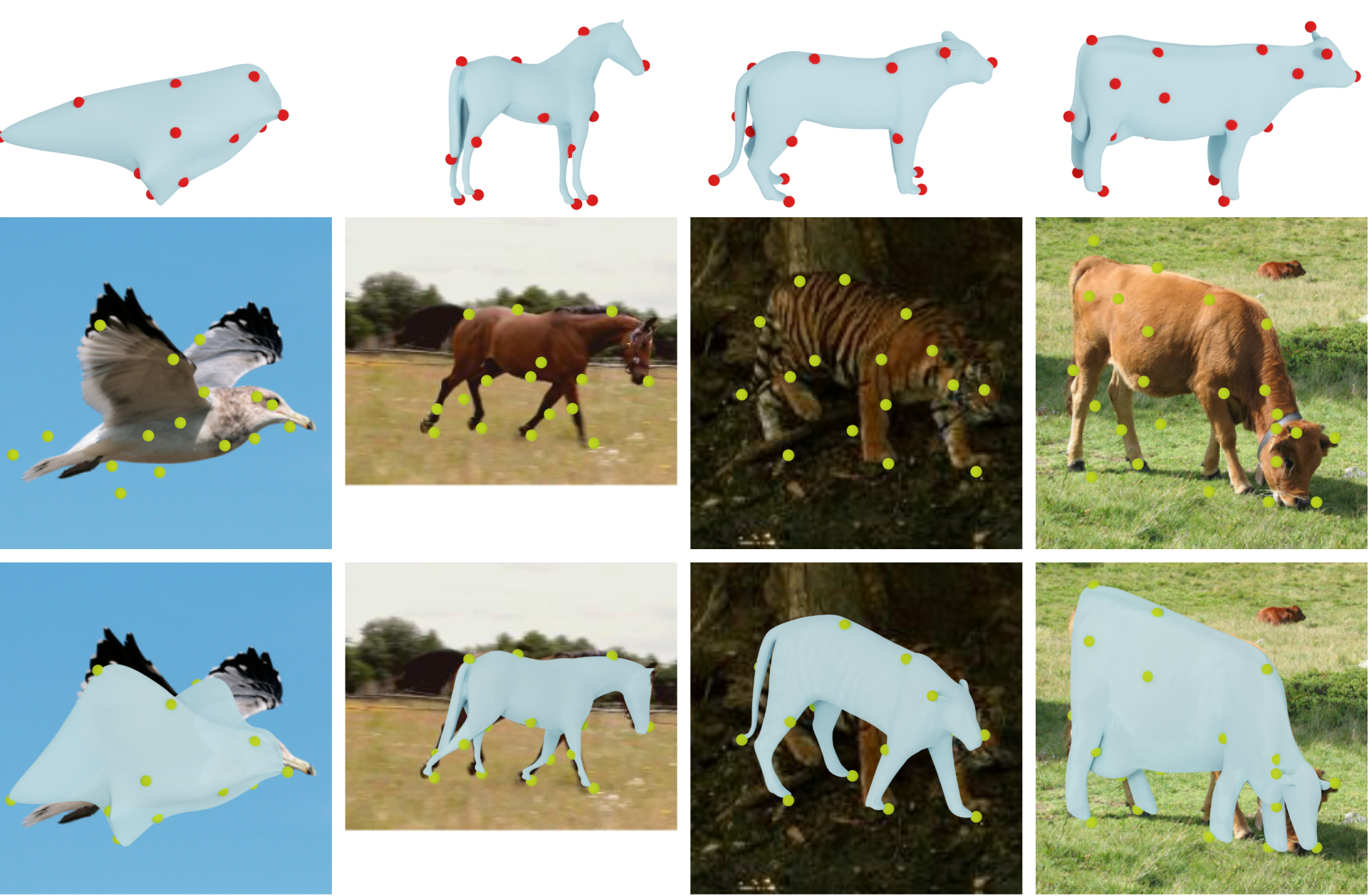}
    \caption{We tackle the problem of monocular 3D reconstruction for articulated object categories by guiding the deformation of a mesh template (top) through a sparse set of 3D control points regressed by a network (middle). Despite using only weak supervision in the form of keypoints, masks and video-based correspondence our approach is able to capture broad articulations, such as  opening wings, as well as motion of the lower limbs and neck (bottom).   \vspace{-0.5cm}}
    \label{fig:my_label}
\end{figure}

\begin{figure*}[t]
  \includegraphics[width=\textwidth]{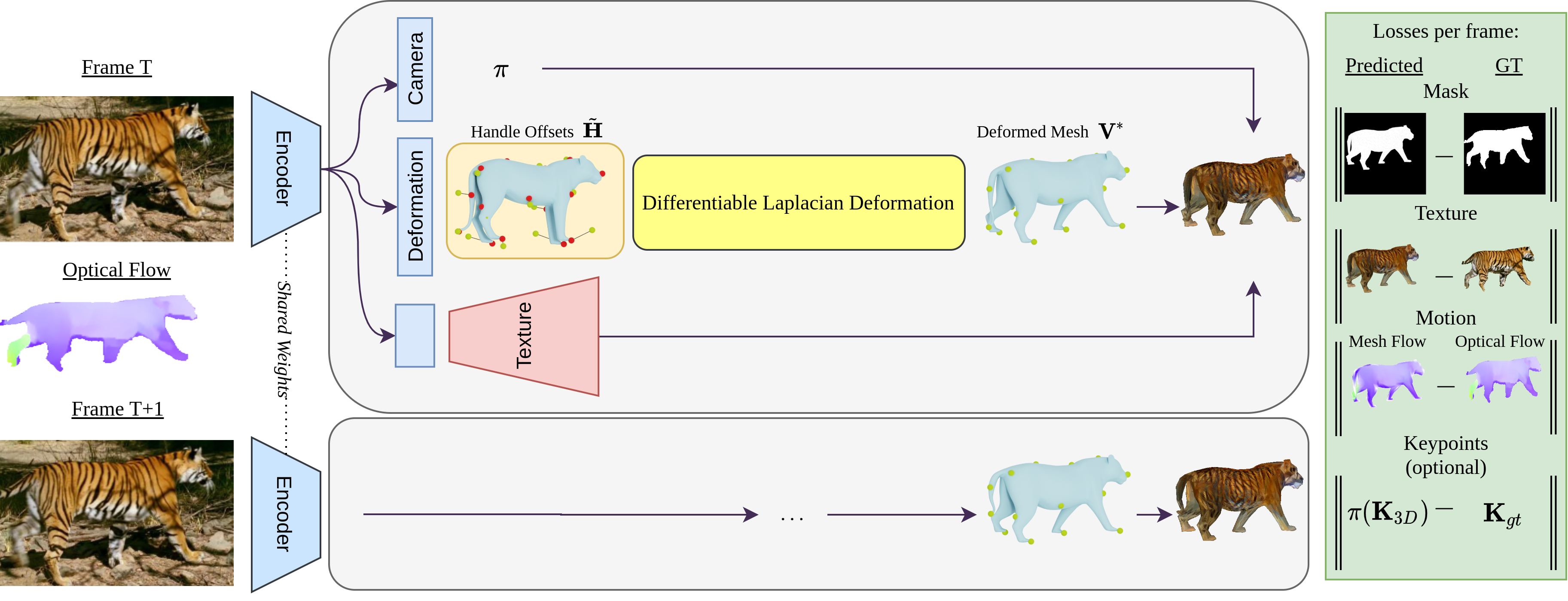}
  \vspace{-0.55cm} 
  \caption{\textbf{Training overview:} Two consecutive frames are separately processed by a network that estimates the camera pose, deformation and UV texture parameters. The network regresses per frame a  mesh $\m V^*$ by estimating offsets to the handles $\m H$ of the template shape and consequently solving the respective Laplacian optimization problem. The  predictions are supervised by  per-frame losses on  masks,  appearance, and optionally keypoints as well as a novel, intra-frame, motion-based loss that compares the predictions of an optical flow network to the mesh-based prediction of pixel displacements (`mesh flow').  \vspace{-0.2cm}}
\end{figure*}

Firstly, we draw inspiration from 3D vision which has traditionally relied on motion information for SFM~\cite{tomasi1992shape, Hartley2004}, SLAM~\cite{ klein2007parallel, newcombe2011dtam} or Non-Rigid SFM~\cite{torresani2003learning,GargRA13,fragkiadaki2013pose}. These category-agnostic techniques interpret 2D point trajectories in terms of an underlying 3D scene and a moving camera. In this work we use the same principle to supervise monocular 3D category reconstruction, effectively allowing us to leverage video as a source of self-supervision.
In particular we establish dense correspondences between consecutive video frames through optical flow and force the back projections of the respective 3D reconstructions to be consistent with the optical flow results. This loss can be back-propagated through the 3D lifting pipeline, allowing us to supervise both the camera pose estimation and mesh reconstruction modules through video. Beyond coming for free, this supervision also ensures that the resulting models will exhibit a smaller amount of jitter and be more flexible when processing videos, since the motion-based loss can penalize  inconsistencies across consecutive frames and failure to co-vary with moving object parts. 

Secondly, we introduce a model for regularised mesh deformations that allows for learnable, part-level mesh control and is back-propagateable, providing us with a drop-in replacement to the common morphable model paradigm adopted in~\cite{cmrKanazawa18}. For this we rely on the Laplacian surface deformation algorithm~\cite{sorkine2004laplacian}, commonly used in geometry processing to deform a template mesh through a set of control points (`handles’) while preserving the surface structure and details. We observe that the result of this optimization-based algorithm is differentiable in its inputs, i.e. can be used as a structured layer, while incurring no additional cost at test time since the expression for the optimum can be folded within a linear layer. We incorporate this operation as the top layer of a deep network tasked with regressing the position of the control points given an RGB image. Our results show that we can learn meaningful control points that allow us to capture limb articulations while also providing a human-interpretable interface that enables  manual post-processing and  refinement using any available 3D software.

Thirdly, we adopt an optimization-based approach to 3D reconstruction that is inspired from bundle adjustment \cite{modern}: given a video, we use the `bottom-up’ reconstructions of consecutive frames delivered by our CNN in terms of cameras and handle positions as the initialisation for a numerical optimisation algorithm. We then jointly optimise the per-frame mask and/or keypoint reprojection losses, and video-level motion consistency losses with respect to the cameras and handle variables, giving a `top-down’ refinement of our solution that better matches the image evidence. We show that this  improves the results at test-time based on whatever image evidence can be obtained without manual annotation.

We evaluate our approach on 3D shape, pose and texture reconstruction on a range of  object categories that exhibit challenging articulations.  Our ablation highlights the importance of the employed self-supervised losses and the tolerance of our method to the number of learnable handles, while both our qualitative and quantitative results indicate that our method largely outperforms recent approaches.

\section{Related Work}
\textbf{Pose, Texture and Articulation Prediction}
Our work addresses the task of inferring the camera pose, articulation and texture corresponding to an input image. Recent works have addressed several aspects of this problem~\cite{cmrKanazawa18, kulkarni2019csm, kulkarni2020articulation} with varying forms of supervision. Earlier approaches like CMR~\cite{cmrKanazawa18} treat the problem of 3D reconstruction from single images using known masks and manually labelled keypoints from single viewpoint image collections. Closer to our work is  Canonical Surface Mapping (CSM)~\etal ~\cite{kulkarni2019csm, kulkarni2020articulation} which produces a 3D representation in the form of a rigid or articulated template using a 2D-to-3D cycle-consistency loss. The articulated variant of CSM~\cite{kulkarni2020articulation} achieves non-rigid deformation by explicitly segmenting 3D parts of the template shape manually set prior to training the method.  Finally, a line of recent research works~\cite{LAE, attila,Wu_2020_CVPR} focus on the disentanglement of images into 3D surfaces with  simultaneous camera, lighting and texture prediction without any ground-truth supervision, but are limited to categories of moderate shape variability, such as faces, cats or symmetric objects in general. By contrast to the works above, our method successfully learns models of highly articulated deformable objects without requiring any special preprocessing, such as manual part segmentation, or strong assumptions, such as symmetry. 

\textbf{Surface Deformation}
Recent works on monocular 3D reconstruction~\cite{cmrKanazawa18, ucmrGoel20} treat deformation as offsets added to mesh vertices, regressed by image-driven CNNs. However regressing vertices can result in surface distortions or corrupt features, while being opaque to a human modeller. 
By contrast we rely on geometry processing methods
~\cite{sorkine2004laplacian, sorkine2007rigid, jacobson2011bounded, jacobson2012fast}, and in particular 
focus on the Laplacian Deformation method ~\cite{sorkine2004laplacian, sorkine2007rigid} 
which uses 
a sparse set of control points to achieve a detail-preserving mesh deformation. We realise that the associated optimization problem can be used as a differentiable, structured layer and use it to both  learn the control points and efficiently regress their 3D position.

\textbf{Video-based supervision}
 Video has been commonly used as a source of weak supervision in  the context of dense labelling tasks such as semantic segmentation \cite{Tokmakov16a} or Densepose estimation \cite{dpslim}. 
 Drawing on the classical use of motion for 3D reconstruction, e.g. \cite{tomasi1992shape, Hartley2004, klein2007parallel, newcombe2011dtam, torresani2003learning,GargRA13,fragkiadaki2013pose} many recent works~\cite{NIPS2017_7108,  thiemo_optical_flow, vijayanarasimhan2017sfm} have also incorporated optical flow information to supervise 3D reconstruction networks. Both in the category-specific \cite{NIPS2017_7108, thiemo_optical_flow} and agnostic \cite{zhou2017unsupervised,UZUMIDB17,vijayanarasimhan2017sfm} setting,
 optical flow provides detailed point correspondences inside the object silhouette which can aid the prediction of object articulations and the reconstruction of the underlying 3D geometry. More recent works have leveraged videos for monocular 3D human reconstruction \cite{texturepose} or  sparsely-supervised  hand-object interactions \cite{hasson2020leveraging} based on photometric losses. In this work we show that motion is a particularly effective source of supervision for our case, where we jointly learn the category-specific shape prior and the 3D reconstructions.  We also rely on robust, occlusion-sensitive optical flow networks \cite{zhao2020maskflownet} which provide a stronger source of supervision than photometric consistency, since they are  trained to both handle the aperture effect in the interior of objects to recover large displacement vectors when appropriate. 
 
\textbf{Cycle Consistency} Our approach is reminiscent of the principle of cycle consistency \cite{Huang_consistentshape,ZhouKAHE16,CycleGAN2017}, where the composition of two maps is meant to result in the identity mapping. Our motion-based approach is in a sense the dual of  \cite{ZhouKAHE16}, where 3D synthetic data were used to learn dense correspondences between categories; here we use correspondences from a pre-trained optical flow network to learn about 3D object categories.

\section{Method Description}
Given an image our target is to infer the 3D shape,  camera pose, and texture of the depicted object. During training we only have at our disposal a single representative mesh for the category ('template'),  a set of videos, and 2D-level supervision from pre-trained models for semantic segmentation; we can optionally also use ground truth for 2D joints and/or segmentation.

In our approach we use single-frame  networks and exploit temporal information only for supervision. At test time we can deploy the learned networks on a per-frame level, but can also exploit temporal information to improve the accuracy of our results through a bundle adjustment-type joint optimization. 

In this section we detail our method. We start by introducing our novel representation of an articulated object's 3D shape  
in terms of a differentiable deformation model in Section~\ref{sec:articulation}.
We then turn to the use of motion as a source of supervision, introducing  our motion-consistency loss in Section~\ref{sec:video}.
In Section~\ref{sec:optimization} we introduce a fine-tuning approach to refine our bottom-up network predictions with a more careful, sample-based optimization and also present other forms of weak supervision used in training.

\newcommand{\temp}[0]{\m T}

\begin{figure}[t]
  \includegraphics[width=\linewidth]{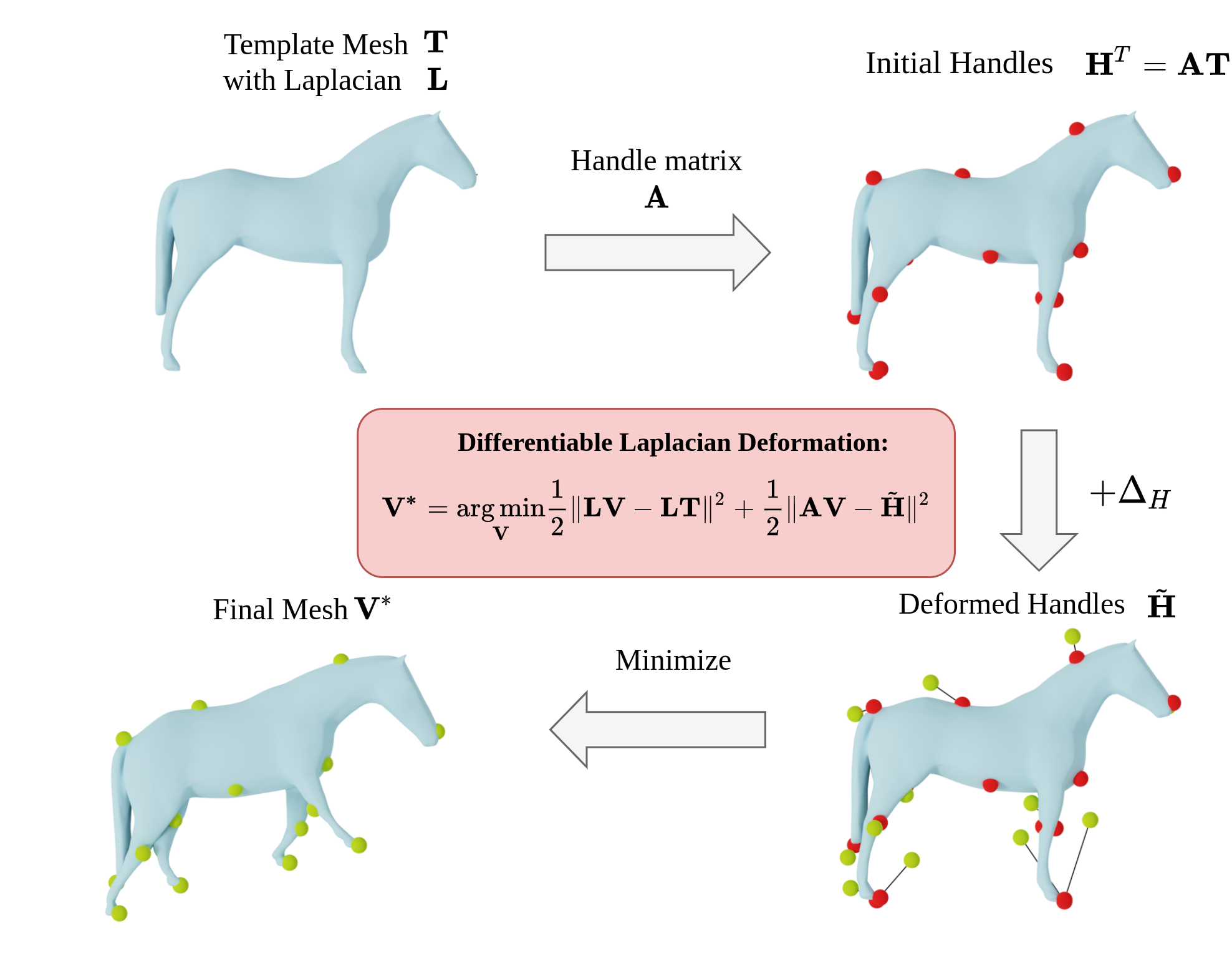}
  \vspace{-1cm}
  \caption{\textbf{Learnable Deformation Layer:} The deformed mesh $\m V^*$ is the result of an optimization scheme forcing $\m V^*$ to retain the surface details of the template mesh while also minimizing  constraints imposed by learnable handles. The optimization solution comes in a closed form, and can be backpropagated through, providing us with a new layer.}
\end{figure}

\subsection{Articulated Mesh Prediction} \label{sec:articulation}
Our aim  is to synthesise the shape of an articulated object category by a neural network. While in broad terms we adopt the deformable template paradigm adopted by most recent works \cite{cmrKanazawa18, ucmrGoel20,umr2020}, we deviate from the morphable model-based \cite{blanz1999morphable} modeling of shape adopted in  \cite{cmrKanazawa18, ucmrGoel20, umr2020}. In those works a shape estimate $\m V$ is obtained in terms of offsets $\m \Delta_{V}$ to a template shape $\temp$, yielding $\m V = \m \Delta_{V} + \temp$, where 
$\m \Delta_{V}$ is delivered by the last, linear, layer of a shape decoder branch, effectively modeling shape variability as an expansion on a linear basis. Such global basis models are well-suited to categories  such as faces or cars, but for objects with part-based articulation such as quadrupeds we argue that a part-level model of deformation is more appropriate - which is also  the approach routinely taken in rigged modeling in graphics. Furthermore, the linear synthesis model is non-interpretable or controllable by humans and requires careful regularization during training to recover plausible meshes.

We propose instead a deformation model where a set of learnable, network-driven control points (or `handles') deform a given template while preserving its shape, as captured by its curvature. For this we build on Laplacian surface editing ~\cite{sorkine2004laplacian}. This model is controllable, interpretable, and regularized by design, while our experiments show that it yields systematically more accurate mesh reconstructions. 

In particular we represent the 3D shape of a category as a triangular mesh $M = (V,F)$ with vertices $\m V \in \mathbb{R}^{N \times 3}$ and fixed edges $F \in \mathbb{Z}^{N_f \times 3}$.
Our deformation approach relies on the cotangent-based discretization $\m L \in \mathbb{R}^{N\times N}$ of the continuous Laplace-Beltrami operator used to calculate the curvature at each vertex of a mesh \cite{taubin1995signal}. 

Instead of manually determining a set of handles, we propose to obtain $K$ 3D handles through a learnable dependency matrix $\m A \in \mathbb{R}_{+} ^{K \times N}$ that is right-stochastic:
\begin{equation}
\m H = \m A \m V, \quad \mathrm{where} \sum_{v} \m A_{k, v}=1,
\end{equation}
forcing each handle to lie in the convex hull of the mesh vertices.
For a given image we obtain the  target handle positions $\m{\tilde{H}}$ by adding a network-driven update $\m \Delta_H$ to the template handles $\m A \m T$: $\m{\tilde{H}} = \m A \m T + \m \Delta_H$.
Based on $\m{\tilde{H}}$, we obtain the deformed mesh $\m{V^*}$ as the minimum of the following quadratic loss:
\begin{equation}
\label{lpl_eq}
\m{\m V^*} = \argmin_{\m V} \frac{1}{2} \left\lVert \m{L}\m{V} - \m L \m \temp \right\rVert ^2 + \frac{1}{2} \left\lVert \m{A}\m{V} - \m{\tilde{H}}\right\rVert^2,
\end{equation}

\noindent where as in \cite{sorkine2004laplacian} the first term forces the solution to respect the curvature of the template mesh, $\m L \m \temp$,
ensuring that salient, high-curvature details of the template shape are preserved, while the second term forces the location of the handles according to $\m{V}$ to be close to the target location, $\m{\tilde{H}}$.

The stationary point of \eqref{lpl_eq} can be found by solving the following linear system:
\begin{equation}
\label{grad_of_lpl_eq}
(\m{L}^\transp \m{L} + \m{A}^\transp \m{A}) \m{V^*} = \m{L}^\transp \boldsymbol{\m L \m T} + \m{A}^\transp \tilde{\m{H}} 
\end{equation}

Given that $(\m{L}^\transp \m{L} + \m{A}^\transp \m{A})$ is symmetric, positive semi-definite and sparse, 
the solution $\m{V ^*}$ can be  efficiently computed with conjugate gradients or sparse Cholesky factorization.
We rely on efficient solvers that cannot be currently handled by automatic differentiation for backpropagating through the linear system solution, and therefore provide the explicit gradient expression in the supplemental material.

Backpropagating gradients through the Laplacian solver allows us to both learn the association of the vertices to the handles via the matrix $\m A$ and also provide gradients back to the handle position $\tilde{ \m H}$ regressor.  As such our method is end-to-end differentiable and no manual annotation, segmentation or rigging of the template mesh is required to achieve part-based articulations.

In practice we initialize the dependency matrix $\m A$ based on Farthest Point Sampling (FPS) \cite{eldar1992irregular} of the mesh, shortlisting a set of vertices $\{v_k\},k=1\ldots K$ that are approximately equidistant.
For each vertex $v_k$ we initialize the $k-$ th row of $\m A$ based on the geodesic distance of the vertices to $v_k$: 
\begin{equation}
    \m A[i,k] = \frac{\exp(1 / d_{i,v_k})}{\sum_j \exp(1 / d_{j,v_k})}
    \end{equation}

We note that at test time we have a constant affinity matrix, $\m{A}$. Combined with the fixed values of $\m{L}$ and $\m{T}$, we can fold the solution of the linear system in  Eq.~\ref{grad_of_lpl_eq} into a linear layer:

\begin{equation}
\label{linear}
\m{V ^*} = \m{C} +  \m{D} \tilde{\m{H}},
\end{equation}
with $\m{C}$ and $\m{D}$ being constant matrices obtained by multiplying both sides of Eq.~\ref{grad_of_lpl_eq} by the inverse of
$\m{L}^\transp \m{L} + \m{A}^\transp \m{A}$.

 We can thus interpret our method as using at training time  template-driven regularization to solve the ill-posed problem of monocular 3D reconstruction, but being as simple and fast as a linear layer at test-time. 

\subsection{Motion-based 3D supervision}
\label{sec:video}
Having described our deformation model, we turn to the use of video information for network training. We rely on optical flow \cite{zhao2020maskflownet}  to deliver pixel-level correspondences  between consecutive object-centered crops.
Unlike traditional 3D vision which relies on category-agnostic point trajectories for 3D lifting, e.g. through factorization \cite{tomasi1992shape}, we use the flow-based correspondences to constrain the mesh-level predictions of our network in consecutive frames.

In particular, our network takes as input a frame at time $t$ and estimates a mesh $\m V_t$ and 
a weak perspective camera $\m C_t$. A mesh vertex $i$ that is visible in both frames $t$ and $t+1$ will project to two image points $\m p_{i,t} = \pi(\m V_{i,t},\m C_t)$ and $\m p_{i,t+1}= \pi(\m V_{i,t+1},\m C_{t+1})$ where $\pi$ amounts to weak perspective projection. As such the displacement of point $\m p_{i,t}$ according to our network will be  $\tilde{\m u}_i = \m p_{i,t+1}- \m p_{i,t}$.

This prediction is compared to the optical flow value $u_i$ delivered at $\m p_{i,t}$  by a pretrained network \cite{zhao2020maskflownet} that we treat as  ground-truth. We limit our supervision to image positions in the interior to the object masks and vertices visible in both frames; vertex visibility is recovered by z-buffering, available in any differentiable renderer. We denote 
the vertices that are eligible for supervision in terms of a binary visibility mask $\boldsymbol \gamma: \{1,\ldots,\Gamma\} \rightarrow \{0,1\}$.

We combine these terms in a `motion re-projection' loss expressed as follows:
\begin{equation}
\label{eq:motion_error}
L_{\text {motion}}=\frac{1}{\sum_{i=1}^\Gamma {\boldsymbol \gamma}_i} \sum_{i = 1}^\Gamma {\boldsymbol \gamma}_i \left\| \m u_i -\tilde{\m u}_i\right\|_{1} 
\end{equation}
where we use the $\ell_1$ distance between the flow vectors for robustness and average over the number of visible vertices to avoid pose-specific value fluctuations.
Since $\tilde{\m u}_i = \pi(\m V_{i,t+1},\m C_{t+1}) - \pi(\m V_{i,t},\m C_t)$ continuously depends on the camera and mesh predictions of our network, we see that this loss can be used to supervise both the camera and mesh regression tasks.

This loss  penalizes the cases where limb articulation observed in the image domain is not reflected in the 3D reconstructions, effectively forcing the 3D reconstructions to become more `agile' by deforming the mesh more actively. Interestingly, we observed that beyond this expected behaviour this loss has an equally important effect on the camera prediction, by forcing the backprojected mesh to `stand still' in the object interior: 
even though different camera poses could potentially backproject to the same object in a single image, a change in the camera across frames will cause large 2D displacements for the corresponding 3D vertices. These are picked up when compared to the predictions of an optical flow system that regresses small displacements in the object interior.

\subsection{Optimization-based learning and refinement}
\label{sec:optimization}
The objective function for our 3D reconstruction task combines  motion supervision with other common losses in a joint objective function:
\begin{equation}
L_{\text {total}} = L_{\text {motion}} + L_{\text {kp }} + L_{\text {pixel}} +  L_{\text {rigid }}+  L_{\text {mask }} +  L_{\text {boundary }},
\end{equation}
capturing keypoint, pixel-level appearance, rigidity priors, as well as mask- and boundary- level supervision for the shape; the forms of the losses are provided in Sec~\ref{sec:losses}, while we omit the empirically-determined loss scaling for simplicity. 

In principle a neural network could successfully minimize the sum of these losses and learn the correct 3D reconstruction of the scene. In practice we are asking  the network to both recover and learn the solution to an ill-posed problem for multiple training samples, which has many local minima. 

This has been observed even in strongly-supervised human pose estimation, where careful per-sample numerical optimization \cite{smplify} was shown to yield substantial performance improvements in  \cite{holopose,kolotouros2019spin,eft,Kulon_2020_CVPR}.
In our weakly-supervised case the local minima problem is even more pronounced.

We use focused, per-sample numerical optimization to refine the network's `bottom-up' predictions so as to better match the image evidence by minimizing $L_{\text {total}}$ with respect to the per-frame handles and camera poses; if the object were rigid this would amount to bundle adjustment, but in our case we also allow the handles to deform per frame. Our approach also applies to both videos and individual frames, where in the latter case we omit the motion-based loss. At test-time, as in  the `synergistic refinement' approach of \cite{holopose}, once the network has delivered its prediction for a test sample (frame/video), we start a numerical `top-down’ refinement of its estimate by minimizing $L_{total}$ using only masks delivered by an instance segmentation network and flow computed from the video if applicable. The approach comes with a computational overhead due to the need for forward-backward passes over the differentiable renderer for every gradient computation.

Further attesting to the importance of per-sample optimization, we note that we have also found a careful initialization of the camera predictions to be critical to the success of our system: as detailed in the supplemental material we train our system by building on the camera multiplex technique~\cite{ucmrGoel20}, that we extend further with the handle deformations.

\subsubsection{Loss terms}
\label{sec:losses}
\textbf{Keypoint reprojection loss}, as in ~\cite{kulkarni2019csm}, penalizes the $\ell_1$ distance between surface-based predictions and ground truth keypoints, when available:
\begin{equation}
L_{\text {kp }}=\sum_{i}\left\|\m k_{i}-\pi\left(\m K_i \m V, \m C\right)\right\|_{1}, \nonumber
\end{equation}
 where $\m K_i$ is a fixed vector that regresses the $i-$th semantic keypoint in 3D from the 3D mesh.

\textbf{Texture Loss} compares the  mesh-based texture and the image appearance in terms of the perceptual similarity metric of~\cite{zhang2018perceptual} after masking by the silhoutette $S$:
\begin{equation}
L_{\text {pixel}}=\operatorname{dist}\left(\tilde{I} \odot S, I \odot S\right). \nonumber
\end{equation}
As in  ~\cite{cmrKanazawa18} we enforce symmetric texture predictions by using a bilateral symmetric viewpoint.

\textbf{Local Rigidity Loss}, as in \cite{kar2015category} aims at preserving the Euclidean  distances between  vertices
 in the extended neighborhood $\mathcal{N}(\mathbf{u})$ of a point $\mathbf{u}$: 
 \begin{equation}
L_{\text {rigid }}=\underset{\mathbf{u} \in V }{\mathbb{E}}  \underset{\mathbf{u}^{\prime} \in \mathcal{N}(\mathbf{u})}{\mathbb{E}}\left|\left\|V\left(\mathbf{u}\right)-V\left(\mathbf{u}^{\prime}\right)\right\|-\left\|\bar{V}(\mathbf{u})-\bar{V}\left(\mathbf{u}^{\prime}\right)\right\|\right| \nonumber
\end{equation}

\textbf{Region similarity loss} compares the object support computed from the mesh by  
a differentiable renderer~\cite{ravi2020pytorch3d} to instance segmentations $S$ provided either by manual annotations or pretrained CNNs using their absolute distance:
\begin{equation}
L_{\text {mask }}=\sum_i \|{S_i}-f_{\text {render}}(V_i, \pi_i)\| \nonumber
\end{equation}

\textbf{Chamfer-based loss} penalizes smaller areas that are hard to align, like hooves or tails: 
\begin{equation}
L_{\text {boundary }}=\underset{\mathbf{u} \in V}{\mathbb{E}} \mathcal{C}_{f g}(\pi(\mathbf{u}))+\underset{\mathbf{b} \in {\mathcal{B}}_{f g}}{\mathbb{E}} \min _{\mathbf{u} \in V}\|\pi(\mathbf{u})-\mathbf{b}\|_2^2,  \nonumber
\end{equation}
where as in ~\cite{fragkiadaki2013pose, kar2015category} the first term  penalizes points  of the predicted
shape that project outside of the foreground mask using the Chamfer distance to it
while the second term penalizes mask under-coverage by ensuring every point on the silhouette boundary has a mesh vertex projecting close to it.

\begin{table}[t]
\centering
\begin{tabular}{l|cc}
\hline
Method & \multicolumn{1}{l}{mIoU} & \multicolumn{1}{l}{PCK} \\ \hline
CMR~\cite{cmrKanazawa18} & 0.703 & 81.2 \\
CSM~\cite{kulkarni2019csm} & 0.622 & 68.5 \\
A-CSM~\cite{kulkarni2020articulation} & 0.705 & 72.4 \\ \hline
Ours &  &  \\
\multicolumn{1}{c|}{8} & 0.64 & 84.6 \\
\multicolumn{1}{c|}{16} & 0.676 & 89.8 \\
\multicolumn{1}{c|}{32} & 0.688 & 89.7 \\
\multicolumn{1}{c|}{64} & \textbf{0.711} & \textbf{91.5} \\ \hline
\end{tabular}
\caption{\textbf{Ablation of deformation layer on CUB:} Even with only 8 points, our handle-based approach outperforms all competing methods in terms of PCK, while with more handles both the mIoU and PCK scores  improve further.\label{tbl_cub_pck}\vspace{-0.2cm}}
\end{table}

\section{Experiments}
\subsection{Model architecture}
We use a similar architecture to CMR~\cite{cmrKanazawa18}, using a Resnet18 encoder and three decoders -one each for predicting articulations, camera pose and texture. The articulation prediction module is a set of 2 fully connected layers with $\mathbb{R}^{K\times 3}$ outputs. In particular for texture prediction, we directly predict the RGB pixel values of the UV image through a residual decoder~\cite{ucmrGoel20}. The texture head is a set of residual upsampling convolution layers that take as input the encoded features of ResNet18 and provide the color-valued UV image; we use Pytorch3D~\cite{ravi2020pytorch3d} as differentiable renderer. A more thorough description of the individual blocks can be found in the supplemental material.

\begin{table*}

\begin{minipage}{.55\linewidth}

\resizebox{\textwidth}{!}{%
\begin{tabular}{ccccllccc}
\multicolumn{1}{c|}{Method} & \multicolumn{3}{c|}{Supervision} & \multicolumn{2}{l|}{\begin{tabular}[c]{@{}l@{}}Training \\ Dataset\end{tabular}} & \multicolumn{2}{c}{Horse} &  Tiger \\ \hline
\multicolumn{1}{c|}{} & \multicolumn{1}{l}{KP} & \multicolumn{1}{l}{Mask} & \multicolumn{1}{c|}{Motion} & \multicolumn{2}{l|}{} & TigDog & Pascal  & TigDog \\ \hline
\multicolumn{1}{l|}{CSM} & \checkmark & \checkmark & \multicolumn{1}{c|}{} & \multicolumn{2}{l|}{\textit{P + I}} & 59.0 & 46.4  & - \\
\multicolumn{1}{l|}{ACSM} & \checkmark & \checkmark & \multicolumn{1}{c|}{} & \multicolumn{2}{l|}{\textit{P + I}} & 57.8 & 57.3  & - \\
\multicolumn{1}{l|}{ACSM} & \checkmark & \checkmark & \multicolumn{1}{c|}{} & \multicolumn{2}{l|}{\textit{TD}} & 68.7 & 44.4  & 36.2 \\

\multicolumn{1}{l|}{Ours, inference} & \checkmark & \checkmark & \multicolumn{1}{c|}{\checkmark} & \multicolumn{2}{l|}{\textit{TD}} & 74.7 & 57.2  & 51.9 \\
\multicolumn{1}{l|}{Ours, refinement} & \checkmark & \checkmark & \multicolumn{1}{c|}{\checkmark} & \multicolumn{2}{l|}{\textit{TD}} & \textbf{83.1} & \textbf{69.5}  & \textbf{55.7} \\ \hline
\multicolumn{1}{l|}{CSM} &  & \checkmark & \multicolumn{1}{c|}{} & \multicolumn{2}{l|}{\textit{P + I}} & 44.7 & 49.7  & - \\
\multicolumn{1}{l|}{ACSM} &  & \checkmark & \multicolumn{1}{c|}{} & \multicolumn{2}{l|}{\textit{P + I}} & 58.1 & 54.2  & - \\
\multicolumn{1}{l|}{ACSM} &  & \checkmark & \multicolumn{1}{c|}{} & \multicolumn{2}{l|}{\textit{TD + YV}} & 26.7  &33.3  & 15.1 \\
\multicolumn{1}{l|}{Ours, inference} &  & \checkmark & \multicolumn{1}{c|}{\checkmark} & \multicolumn{2}{l|}{\textit{TD + YV}} & 42.5 & 31.6  & 28.4 \\
\multicolumn{1}{l|}{Ours, refinement} &  & \checkmark & \multicolumn{1}{c|}{\checkmark} & \multicolumn{2}{l|}{\textit{TD + YV}} & \textbf{61.3} & \textbf{54.9}  & \textbf{32.5} \\ \hline
\multicolumn{9}{c}{\textbf{Datasets:} Pascal (P), ImageNet (I), TigDog (TD), YVIS (YV)}\\ \hline     
\end{tabular}
}
\end{minipage}
\begin{minipage}{.42\linewidth}
\vspace{-2.2cm}
\hspace{0.4cm}
\resizebox{.95\textwidth}{!}{%
\begin{tabular}{cccllc}
\multicolumn{1}{c|}{Method} & \multicolumn{2}{c|}{Supervision} & \multicolumn{2}{l|}{\begin{tabular}[c]{@{}l@{}}Training \\ Dataset\end{tabular}} &  Cow \\ \hline
\multicolumn{1}{c|}{} &  \multicolumn{1}{l}{Mask} & \multicolumn{1}{c|}{Motion} &  Pascal \\ \hline
\multicolumn{1}{l|}{CSM}  & \checkmark & \multicolumn{1}{c|}{} & \multicolumn{2}{l|}{\textit{P + I}} & 37.4  \\
\multicolumn{1}{l|}{ACSM}  & \checkmark & \multicolumn{1}{c|}{} & \multicolumn{2}{l|}{\textit{P + I}} & 43.8  \\
\multicolumn{1}{l|}{Ours, inference} & \checkmark & \multicolumn{1}{c|}{\checkmark} & \multicolumn{2}{l|}{\textit{TD + YV}}  & 44.6  \\
\multicolumn{1}{l|}{Ours, refinement} & \checkmark & \multicolumn{1}{c|}{\checkmark} & \multicolumn{2}{l|}{\textit{TD + YV}} & \textbf{53.9}  \\ \hline
\end{tabular}
}
\end{minipage}
\vspace{-0.25cm}
\caption{\textbf{Keypoint Reprojection Accuracy} We report  PCK accuracy (higher is better)  achieved by recent methods~\cite{kulkarni2019csm, kulkarni2020articulation} for articulate object categories. We indicate datasets used to train each method alongside with supervision method; the CSM/ACSM models trained on P+I  do not contain tiger models, while for cows we cannot provide keypoint-supervised results due to the lack of keypoints on videos. Both when training with keypoints and without keypoints we observe substantial improvements over models that were trained without exploiting motion.
\label{tbl:pck_scores} \vspace{-0.25cm}}
\end{table*}

\subsection{Data}
We report quantitative reconstruction results for objects with keypoint-annotated datasets, i.e birds, horses, tigers and cows. We have collected a dataset for a wide set of objects, mainly building on available video datasets~\cite{delpero15cvpr,yang2019video}. All of the videos in our datasets have been filtered manually for occluded or heavily truncated clips that are removed from the dataset. Indicative video samples are provided in the supplemental material; we will make our datasets publicly available to further foster research in this direction. 

\textbf{Birds} We  use  the CUB~\cite{WahCUB_200_2011} dataset for training and testing on birds which contains 6000 images. The train/val/test split we use for training and report is that of~\cite{cmrKanazawa18}. While this dataset is single-frame, we use it to compare our deformation module with prior works on similar grounds.

\textbf{Quadrupeds (Horses, Tigers)} We  use the TigDog Dataset~\cite{delpero15cvpr} which contains keypoint-annotated videos of horses and tigers. The segmentation masks are approximate since they are extracted using MaskRCNN~\cite{he2017mask}. We also drop the neck keypoint for both categories since there is a left-right ambiguity in all annotations. For every class we keep 14 videos purely for evaluation purposes and train with the rest, i.e 53 videos for horses and 44 for tigers. For these classes, the number of handles is set to $K=16$.

\textbf{Quadrupeds (cows, giraffes, zebras and 3 others)} We  use  Youtube Video Instance Segmentation dataset (YVIS)~\cite{yang2019video} to  reconstruct more animal classes in 3D. The cow category is used for evaluation  since it is the only one for which keypoing ground-truth is available;  for the remaining 5 classes we only provide qualitative results in the supplementary material. 

For all categories we downloaded template shapes from the internet and downsampled to a fixed number of $N=642$ vertices. For evaluation we use identical template shape and keypoint annotations to those of \cite{kulkarni2020articulation} for all classes.

\subsection{Results\vspace{-0.1cm}}

\subsubsection{Handle-based deformation evaluation\vspace{-0.1cm}}
\begin{figure*}[h]
\captionsetup[subfigure]{labelformat=empty, position=top}
\begin{centering}
\setcounter{subfigure}{0}
\subfloat{\includegraphics[width=.105\textwidth]{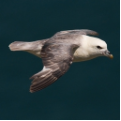}}~
\subfloat[ACSM]{\includegraphics[width=.105\textwidth]{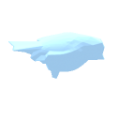}}~
\subfloat[CMR]{\includegraphics[width=.105\textwidth]{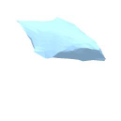}}~
\subfloat[Ours]{\includegraphics[width=.105\textwidth]{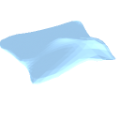}}~
\subfloat{\includegraphics[width=.105\textwidth]{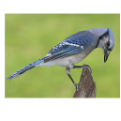}}~
\subfloat[ACSM]{\includegraphics[width=.105\textwidth]{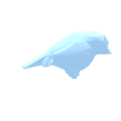}}~
\subfloat[CMR]{\includegraphics[width=.105\textwidth]{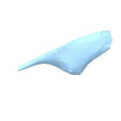}}~
\subfloat[Ours]{\includegraphics[width=.105\textwidth]{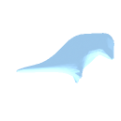}}~
\vspace{-0.2cm}
\caption{\textbf{Bird reconstructions} For each input image we provide the results of CMR~\cite{cmrKanazawa18} and ACSM~\cite{kulkarni2020articulation}
alongside with our method. We observe that we better capture wing and beak deformation. \vspace{-0.5cm}
\label{fig:cub}
}
\end{centering}
\end{figure*}

We start with  the CUB~\cite{WahCUB_200_2011} dataset  where we use the exact supervision of A-CSM~\cite{kulkarni2019csm}. We outperform the state-of-the-art system on reconstruction~\cite{cmrKanazawa18} by a significant margin in both mean Intersection over Union (mIoU) and  keypoint reprojection accuracy (PCK), while following their evaluation conventions. 
We ablate in particular the effect of the number of handles on the achieved 3D reconstruction in Table~\ref{tbl_cub_pck}. We observe that our results are outperforming previous methods even with a very small number of handles, however increasing the number of handles allows for improved performance. 
We also provide qualitative results in Figure~\ref{fig:cub} where we show that our method is capable of correctly deforming the template mesh to produce highly flexible wings, while the alternative methods barely capture open wing variation. These results clearly indicate the merit of our handle-based deformation layer.

\begin{table}[]
\centering
\resizebox{\textwidth}{!}{%
\begin{tabular}{l|cccc}
\hline
Horses                    & \multicolumn{2}{l}{w/ $L_{Motion}$} & \multicolumn{2}{l}{w/o $L_{Motion}$} \\ \hline
                          & mIoU            & PCK            & mIoU             & PCK            \\
Inference                 & 0.536           & 74.7           & 0.519            & 71.5           \\
Mask refinement           & 0.691           & 79.5           & 0.691            & 79.5           \\
Mask and motion refinement & 0.631           & 83.1           & 0.675            & 72.5           \\ \hline
\end{tabular}
}
\vspace{-0.25cm}
\caption{\textbf{Ablation} of horse reconstruction using motion-based supervision (left vs. right) and optimization-based reconstruction based on masks and motion (rows 1-3).  \vspace{-0.3cm}\label{motionoptim_horse}}
\end{table}

\vspace{-0.2cm}
\subsubsection{Motion- and Optimization- based  evaluation}
In Table~\ref{motionoptim_horse} we  ablate the impact of our motion-based supervision and optimization-based reconstruction for the category of horses.
We consider firstly the impact that motion-based supervision has as a source of training (left versus right columns). We observe that motion supervision systematically improves accuracy across all configurations and evaluation measures.

\begin{figure}[t]
  \centering
  \includegraphics[width=0.8\linewidth]{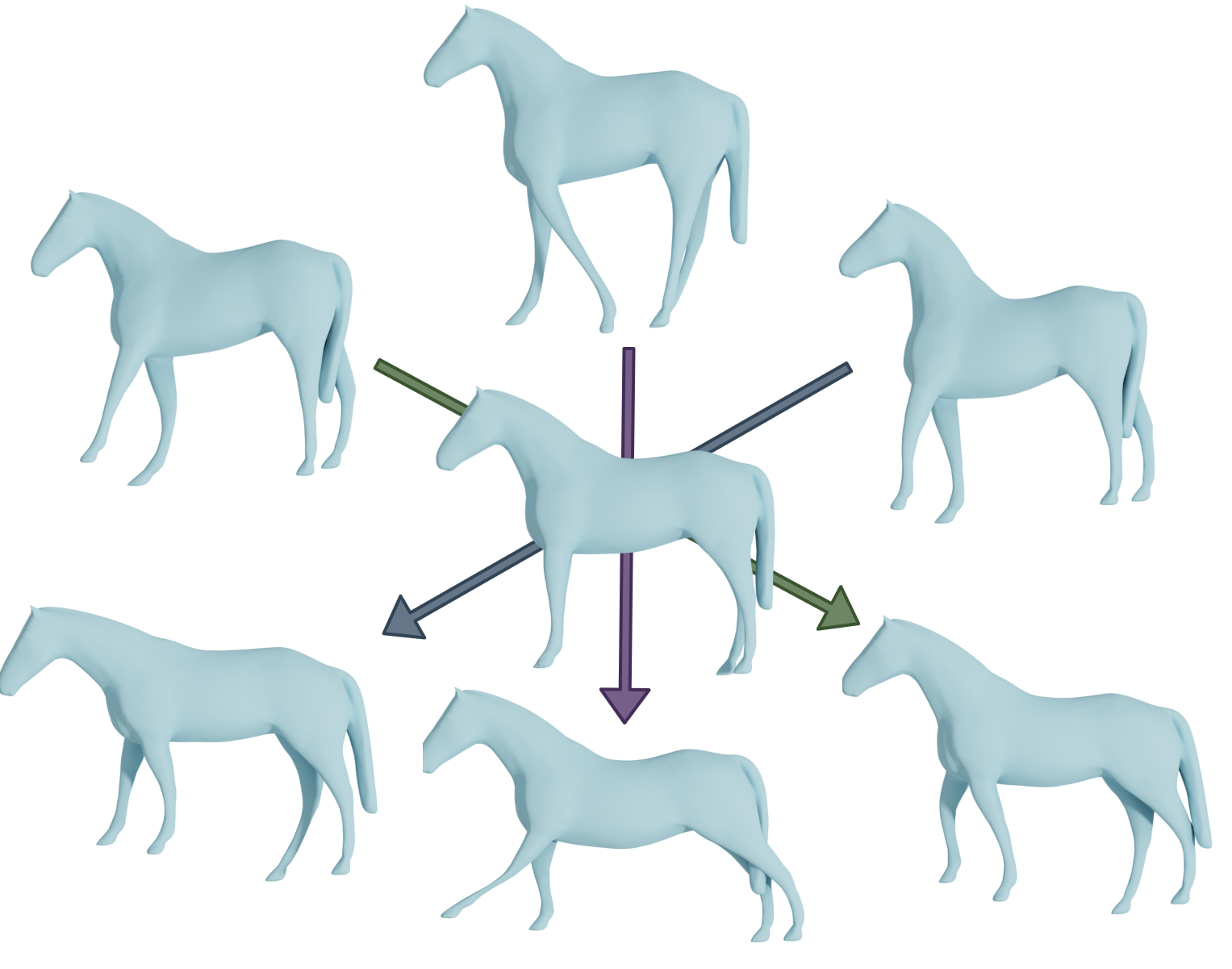}
  \vspace{-0.5cm}
  \caption{\textbf{Learned Deformations} Visualization of the predicted deformations by depicting the mean shape in the center and the first 3 modes obtained by PCA on the handle estimates obtained across the dataset.  \vspace{-0.3cm}\label{fig:pca}}
\end{figure}

When optimizing at test time as post-processing we observe how the terms that drive the optimization influence the final results: when using only masks we have a marked increase in mIoU, and a smaller increase in PCK, while when taking motion-based terms into account as well the increase in mIoU is not as big but we attain the highest improvement in PCK. We visualize in Figure~\ref{fig:pca} the mean shape of the horse along with the first 3 common deformation modes. 

\begin{figure*}[t]
\captionsetup[subfigure]{labelformat=empty, position=top}
\begin{centering}
\setcounter{subfigure}{0}

\subfloat{\includegraphics[width=.12\textwidth]{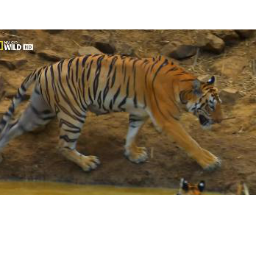}}~
\subfloat{\includegraphics[width=.12\textwidth]{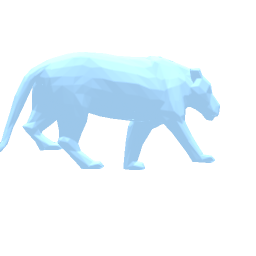}}~
\subfloat{\includegraphics[width=.12\textwidth]{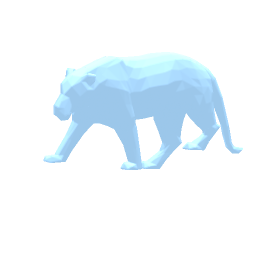}}~
\subfloat{\includegraphics[width=.12\textwidth]{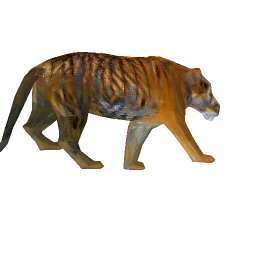}}~
\subfloat{\includegraphics[width=.12\textwidth]{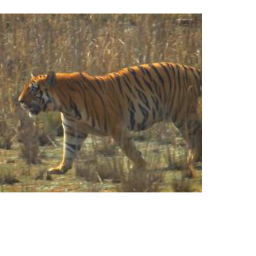}}~
\subfloat{\includegraphics[width=.12\textwidth]{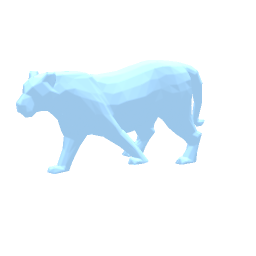}}~
\subfloat{\includegraphics[width=.12\textwidth]{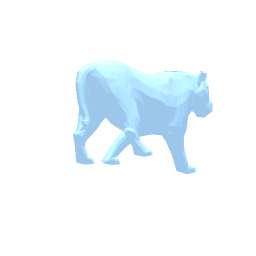}}~
\subfloat{\includegraphics[width=.12\textwidth]{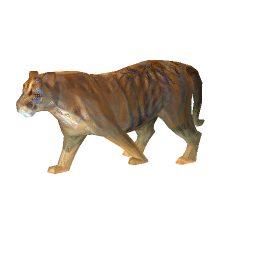}}~\\ 
\vspace{-0.88cm}
\subfloat{\includegraphics[width=.12\textwidth]{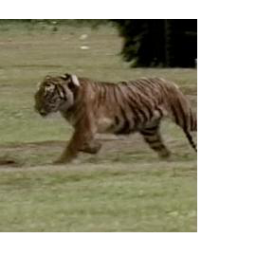}}~
\subfloat{\includegraphics[width=.12\textwidth]{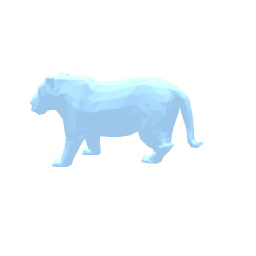}}~
\subfloat{\includegraphics[width=.12\textwidth]{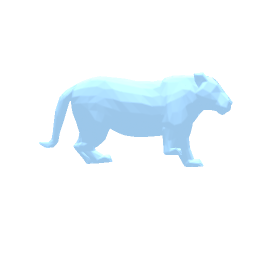}}~
\subfloat{\includegraphics[width=.12\textwidth]{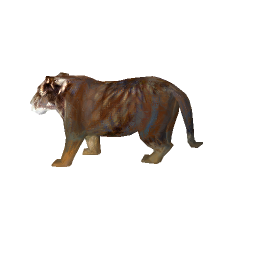}}~
\subfloat{\includegraphics[width=.12\textwidth]{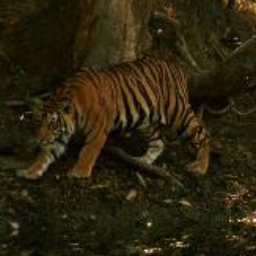}}~
\subfloat{\includegraphics[width=.12\textwidth]{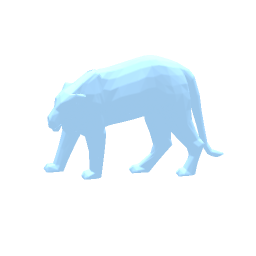}}~
\subfloat{\includegraphics[width=.12\textwidth]{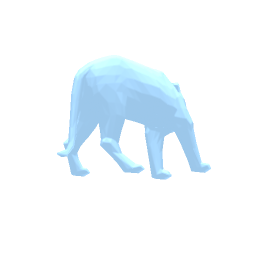}}~
\subfloat{\includegraphics[width=.12\textwidth]{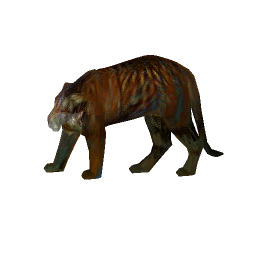}}~\\
\vspace{-0.55cm}
\subfloat{\includegraphics[width=.12\textwidth]{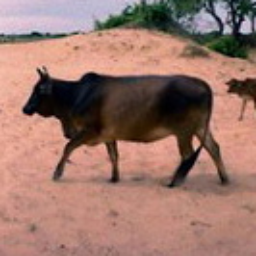}}~
\subfloat{\includegraphics[width=.12\textwidth]{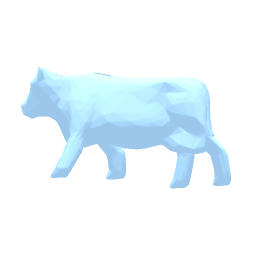}}~
\subfloat{\includegraphics[width=.12\textwidth]{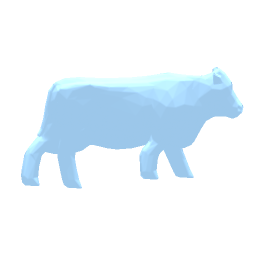}}~
\subfloat{\includegraphics[width=.12\textwidth]{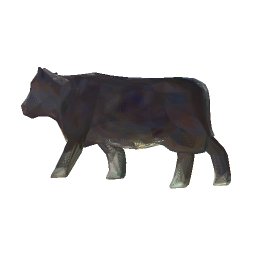}}~
\subfloat{\includegraphics[width=.12\textwidth]{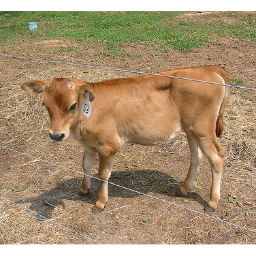}}~
\subfloat{\includegraphics[width=.12\textwidth]{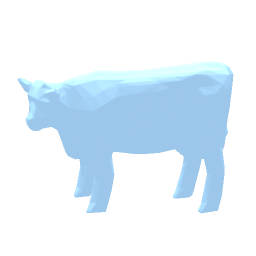}}~
\subfloat{\includegraphics[width=.12\textwidth]{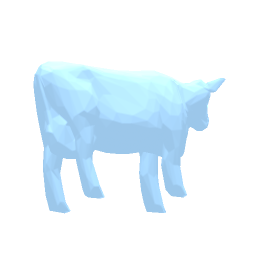}}~
\subfloat{\includegraphics[width=.12\textwidth]{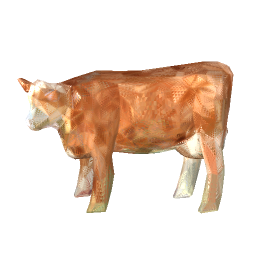}}~\\ 
\vspace{-0.37cm}
\subfloat{\includegraphics[width=.12\textwidth]{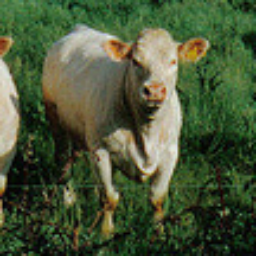}}~
\subfloat{\includegraphics[width=.12\textwidth]{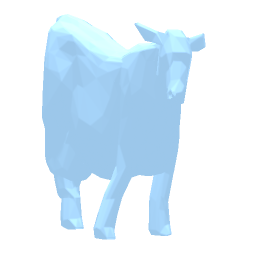}}~
\subfloat{\includegraphics[width=.12\textwidth]{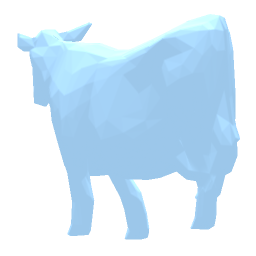}}~
\subfloat{\includegraphics[width=.12\textwidth]{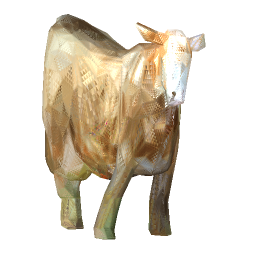}}~
\subfloat{\includegraphics[width=.12\textwidth]{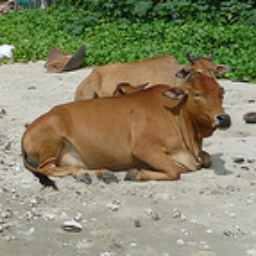}}~
\subfloat{\includegraphics[width=.12\textwidth]{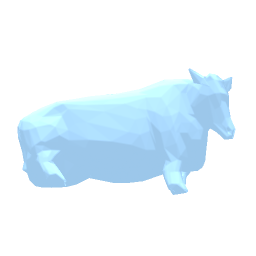}}~
\subfloat{\includegraphics[width=.12\textwidth]{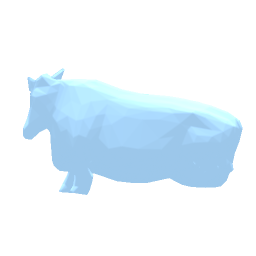}}~
\subfloat{\includegraphics[width=.12\textwidth]{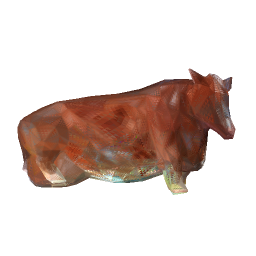}}~\\
\vspace{-0.35cm}
\subfloat{\includegraphics[width=.12\textwidth]{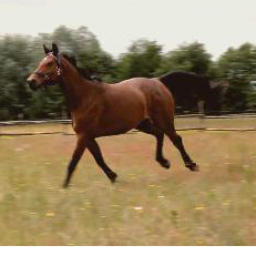}}~
\subfloat{\includegraphics[width=.12\textwidth]{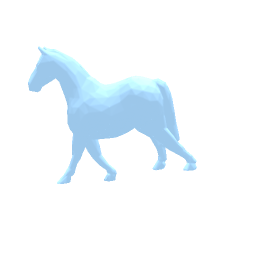}}~
\subfloat{\includegraphics[width=.12\textwidth]{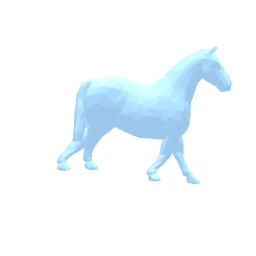}}~
\subfloat{\includegraphics[width=.12\textwidth]{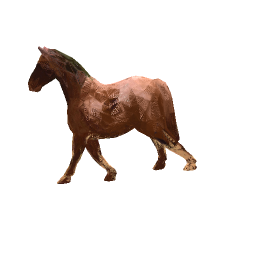}}~
\subfloat{\includegraphics[width=.12\textwidth]{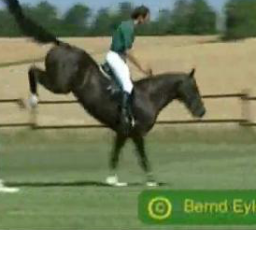}}~
\subfloat{\includegraphics[width=.12\textwidth]{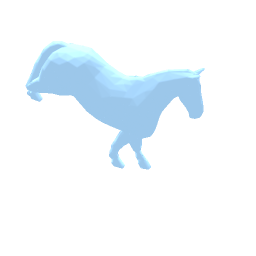}}~
\subfloat{\includegraphics[width=.12\textwidth]{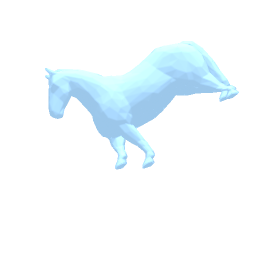}}~
\subfloat{\includegraphics[width=.12\textwidth]{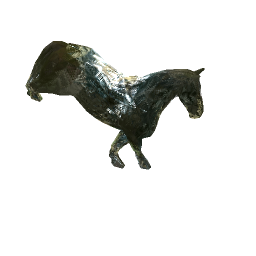}}~\\ 
\vspace{-0.12cm}
\caption{\textbf{Quadruped reconstructions} of our  method. We provide renderings of the 3D reconstruction using the estimated camera pose, a different viewpoint and the texture reconstruction.
We observe that our method successfully captures large articulated deformations as well as viewpoint variability. For videos of side-by-side comparisons to ~\cite{kulkarni2020articulation} please visit \url{https://fkokkinos.github.io/video_3d_reconstruction/}.
\vspace{-0.16cm}
}\label{fig:quad}
\end{centering}
\end{figure*}

\subsubsection{Comparisons on more categories}
In Table~\ref{tbl:pck_scores} we report results on more categories where we have been able to compare to the currently leading approaches to monocular 3D reconstruction ~\cite{cmrKanazawa18, kulkarni2019csm, kulkarni2020articulation}. We use a small number of videos (53  for horses, 44 for tigers, 24 for cows) compared to the thousands of images available in Imagenet and Pascal used by the existing approaches. 

Starting with the comparison on horses for the case where keypoints are available, we observe that our inference-only method has a clear lead when testing on the TigDog dataset (the other methods have not been trained on TigDog), while optimization results in a further boost. When tested on Pascal, our inference-only results are comparable to the best, while optimization gives us a clear edge. For cows we did not have videos with cow keypoints, as such we did not train our approach on it. Furthermore, we trained ACSM for horses and tigers on the TigDog dataset in order to have fair comparison to our method. The TigDog-trained ACSM got a significant boost on TigDog test set but the performance was still not on par with our result.

Turning to results where we do not use keypoints, we observe that when used in tandem with post-processing optimization our method outperforms both CSM and ACSM,
while when compared to ACSM trained on the same data we have a substantial boost on TigDog-Horse and TigDog-Tiger. 
Overall we observe a larger drop in accuracy compared to the results obtained when keypoint supervision is available. As we show in the supplemental material, this may be due to the large flexibility of our deformable model, which manages to  ``overfit'' to the mask rather than performing the appropriate global, rigid transforms. For the case of cows we observe that even though our model was never trained on Pascal data, it outperforms the mask-supervised variants of ACSM.

A pattern that is common for both sets of results is that post-processing optimization yields a substantial improvement in accuracy. As our qualitative results indicate in Figure~\ref{fig:quad} and the Supplemental, this is reflected also in the large amount of limb articulation achievable by our model.
Failure cases, provided in the supplementary material are predominantly due to wrong global camera parameters such as scale, which we attribute to the small diversity of appearance in our limited set of videos. We anticipate further improvements in the future by combining diverse images from static and strong,  motion-based supervision from dynamic datasets. Finally, in some cases our model fails to predict good textures commonly for moving parts of quadrupeds like the legs.

\section{Conclusion} 
We have presented a motion- and geometry-based deep learning  
framework for monocular reconstruction that
combines ideas from deep learning and geometry  
for the unsupervised reconstruction of highly articulated objects; we anticipate that the interpretable and controllable nature of our approach will help handle multiple animate object classes in augmented reality and graphics.

\begin{figure*}[t]
  \includegraphics[width=.9\linewidth]{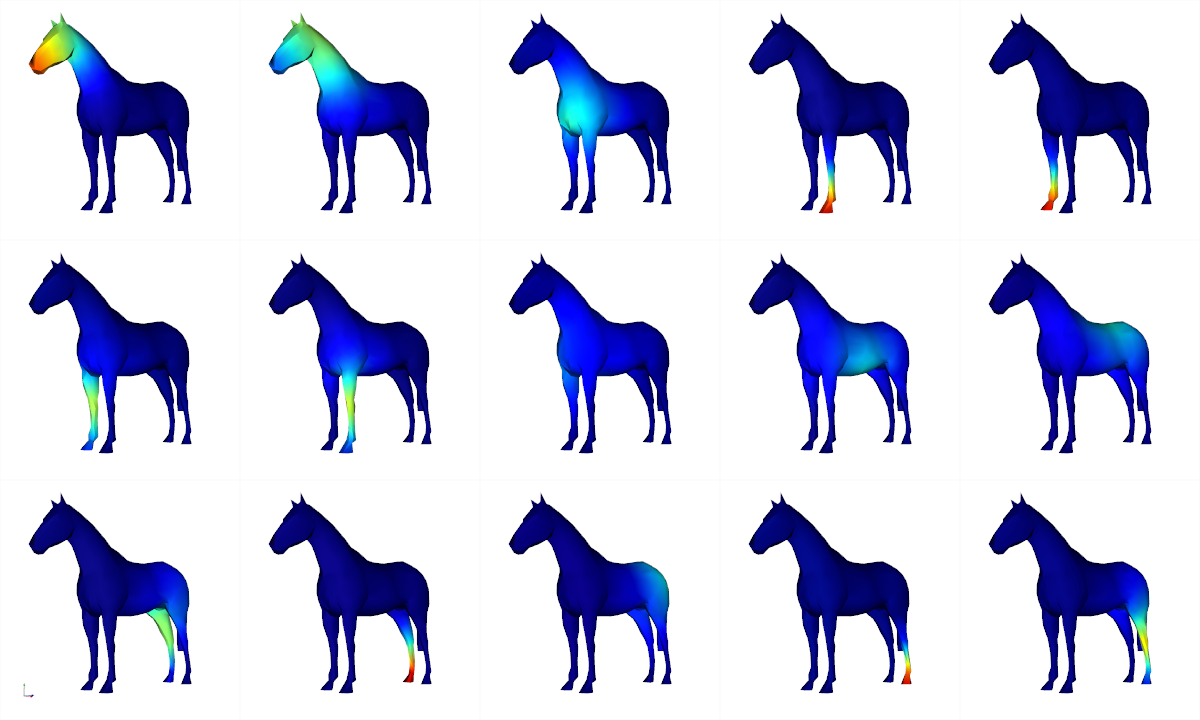}

  \caption{\textbf{Handle Influence:} We visualize each row of the matrix $\mathbf{D}$ as presented in Section~\ref{sec:articulation} for the horse class.  We observe that the associations between the learned handles and the 3D points are localized to areas of intuitive interest such as legs, tail and the head which allow for accurate deformation of the template.}
\end{figure*}

{\small
\bibliographystyle{ieee_fullname}
\bibliography{egbib}
}

\pagebreak

   \newpage
   \null
   \vskip .375in
   \begin{center}
      {\Large \bf \ Supplementary: Learning  monocular 3D reconstruction of articulated categories from motion \par}
      \vspace*{24pt}
      {
      \large
      \lineskip .5em
      \par
      }

      \vspace*{12pt}
   \end{center}

\appendix

\input{supplementary.tex}

\end{document}

%% file: supplementary.tex
\section{Overview}
We provide additional results including visualizations of the learned deformations, comparisons to CMR~\cite{cmrKanazawa18} and ACSM~\cite{kulkarni2020articulation}, qualitative results on a collection of different objects and technical details about the training procedure. Please note that the supplementary material contains a collection of video results. 

\section{Training Procedure}

\subsection{Per-sample optimization training framework}

We build on the camera multiplex training procedure proposed by Goel~\etal~\cite{ucmrGoel20} to train the proposed method. We provide a small review of the method for coherency, however we refer the reader to~\cite{ucmrGoel20} for a thorough explanation and technical details.

The authors propose using a learnable set of possible camera hypotheses  for each training instance that is learned simultaneously with the rest of the 3D reconstruction CNN. In detail, each training instance $i$ has $C_i= \{\pi_1, \dots, \pi_{N_c}\}$ associated camera hypotheses, which are modelled as weak perspective cameras $(s\in\mathbf{R}, \m t\in\mathbf{R}^2, \m q\in\mathbf{R}^4)$, that are retrieved from a 'camera database' during training using unique indices for each training sample. Each camera $\pi_i$ is optimized to minimize the reconstruction losses for the silhouette $L_{sil,i}$ and the texture $L_{tex,i}$ like those used in our proposed framework. For updating the parameters of the method, the resulting losses $L_i = L_{sil,i} + L_{tex,i}$ of the camera set are used as a distribution over the most likely camera pose. This is encoded as a probability $p_{i}=\frac{e^{-L_{i}}}{\sum_{j} e^{-L_{j}}}$ for $\pi_i$ to be the most likely camera. Using the computed distribution the final loss is formulated as 

$$ L_{\text {total }}=\sum_{i} p_{i}\left(L_{\text {sil}, i}+L_{\text {tex}, i}\right). $$

The final step of this training procedure is to train a camera predictor using the most probable camera of each training image conditioned on the image features extracted from the backbone CNN that is driving the whole reconstruction process.

Building on this optimization driven paradigm, we extend the training protocol with three distinct modifications. First of all, alongside the silhouette and textures losses we incorporate also the motion re-projection loss that is described in detail in the main paper. Furthermore, we extend the camera set with per image deformations $D$ which is only one per image unlike the multiple cameras. In our pipeline, each training image $i$ has a camera set $C_i$ and a single handle deformation vector $D_i \in \mathbf{R} ^ {K \times 3} $ (with $K$ being the number of handles) that are both used to express a multi-hypotheses distribution similar to~\cite{ucmrGoel20}. Thirdly, unlike the aforementioned work, we simultaneously train our deformation and camera prediction branches using the most probable explanation for both the camera and deformation in accordance to the resulting silhouette and texture losses. This is achieved by adding two extra losses in the total loss function that minimize the $\ell_2$ norm of the difference between the predicted quantity and the optimized one retrieved from the 'database' for each of the cameras and deformations.

In the main paper we presented experiments that relied on mask, motion, texture losses and in some cases also semantic keypoint re-projection loss. In the case of keypoint trained networks, we set $N_c=1$ and initialize the camera with a rigid SfM camera in accordance to CMR~\cite{cmrKanazawa18}. For all other experiments, we use $N_c=8$  and initialize the camera set $C$ for every image in the training set with camera hypotheses whose azimuth is uniformly spaced on the viewing sphere. The handle deformations $D$ are initialized with zeros which corresponds to the template shape. As a first step, each camera is optimized using the silhouette loss $L_{sil}$ and motion loss $L_{motion}$ using the template shape before training the rest of the method. We implement a drop hypotheses procedure~\cite{ucmrGoel20} to reduce the computational complexity where the most improbable hypotheses are discarded from the camera set. In detail, after 20 epochs we keep the four most probable cameras and after 100 epochs we keep only the 2 most probable cameras.  Any training augmentations that scale and translate the training image $i$ are directly encoded as affine transformations on the respective camera set $C_i$ while the deformation $D_i$ remains unchanged since the depicted deformation of the object remains identical. 

\subsection{Architecture Details}
We use the same encoder-decoder architecture that is presented in~\cite{cmrKanazawa18, ucmrGoel20}. Every image is encoded using an ImageNet pre-trained ResNet18 to a latent feature map $z \in \mathbf{R} ^{4 \times 4 \times 256}$. A flattened version of $z$ is processed with two MLP linear layers with output channels equal to 200 and the final result is given to the handle deformation  predictor and camera predictor branches. The handle deformation branch provides the handle offsets $\Delta_H \in \mathbf{R} ^{K \times 3}$ and the camera predictor predicts the scale, translation and rotation, which is encoded as quartenions, through 2 fully connected layers each with 200 channels. Finally, we use the same texture predictor architecture as~\cite{ucmrGoel20} which takes as input the encoded features $z$ and outputs the UV texture map $I^{uv} \in \mathbf{R} ^ {128 \times 256 \times3}$.

\section{Learnable Laplacian Solver}

Our algorithm builds on Laplacian surface editing techniques~\cite{sorkine2004laplacian} which allow us to control a
template mesh through handles while minimally distorting the template's shape. 
We represent the 3D shape of a category as a triangular mesh $M = (V,F)$ with vertices $\m V \in \mathbb{R}^{N \times 3}$ and fixed edges $F \in \mathbb{Z}^{N_f \times 3}$.
Our deformation approach relies on the cotangent-based discretization $\m L \in \mathbb{R}^{N\times N}$ of the continuous Laplace-Beltrami operator  used to calculate the curvature at each vertex of a mesh \cite{taubin1995signal}. 

We obtain our $K$  handles $H_{1,\dots, K}$ through a learnable dependency matrix $\m A \in \mathbb{R}_{+} ^{K \times N}$ that is right-stochastic, i.e. $\quad \sum_{v} \m A_{k, v}=1$, 
effectively forcing  every handle to lie in the convex hull of the mesh vertices by $\m H = \m A \m V$. 
The network's task is phrased as regressing the  handle positions, denoted as $\m \Delta_H$. Based on those handles, we obtain the deformed mesh $\m{V^*}$ as the minimum of the following quadratic loss:

\begin{equation}
\label{lpl_eq}
\m{\m V^*} = \argmin_{\m V} \frac{1}{2} \left\lVert \m{L}\m{V} - \m L \m \temp \right\rVert ^2 + \frac{1}{2} \left\lVert \m{A}\m{V} - \m{\tilde{H}}\right\rVert^2,
\end{equation}

where as in \cite{sorkine2004laplacian} the first term enforces the solution to respect the curvature of the template mesh, $\m L \m \temp$, while the second one penalizes the difference between the location of the handles according to $\m{V}$ and the target location, $\m{\tilde{H}} = \m H \m T + \m \Delta_H$.
The stationary point of \eqref{lpl_eq} can be found by solving the following linear system:

\begin{equation}
\label{grad_of_lpl_eq}
(\m{L}^\transp \m{L} + \m{A}^\transp \m{A}) \m{V} = \m{L}^\transp \boldsymbol{\m L \m T} + \m{A}^\transp \tilde{\m{H}} 
\end{equation}

The solution $\m{V ^*}$ of Eq.~\eqref{grad_of_lpl_eq} can be very efficiently computed with conjugate gradients or sparse solvers. In the forward case we obtain the solution using a sparse least square solver by concatenating the two matrices $\m L  $ and $\m A$. As it will be presented the backward operation requires a different treatment and as such in the backward operation we make use of a linear solver for PSD matrices.

 We want to compute the gradients with respect to the learnable dependency matrix $\m{A}$ and the handle offset $\tilde{\m H}$. We rewrite equation~\eqref{grad_of_lpl_eq} as $\m{W} \m{V} = \m{b}$ where $\m{W} = \m{L}^\transp \m{L} + \m{A}^\transp \m{A}$ and $\m{b}=\m{L}^\transp \m L \temp + \m{A}^\transp \m{H}$. The direct solution is $\m{V} = \m{W}^{-1} \m{b}$ and $\m{W}$ is a symmetric PSD matrix as the addition of two likewise matrices. Instead of resorting to matrix inversion, we compute $V$ using a linear solver for symmetric PSD matrices. To compute the necessary gradients for backpropagation of gradients through Equation~\eqref{lpl_eq}, we rely on matrix calculus and  use of three Kronecker product $\otimes$ properties~\cite{fackler2005notes}: 1) $\text{vec}(\m Q \m W \m E) = (\m E^\transp \otimes \m Q) \text{vec}(\m W)$, 2) $ T_{m,n} \text{vec}(Q) = \text{vec}(\m Q^\transp)$ and 3) $(\m Q \otimes \m W) (\m E \otimes \m R) = (\m Q \m E \otimes \m W \m R)$ .

The gradients of any linear solver for symmetric matrices are the following
\begin{equation}
 \begin{split}
     \frac{\partial g(\m{V})}{\partial \m{b}} & =  \frac{\partial \m{V}}{\partial \m{b}} \frac{\partial g(\m{V})}{\partial \m{V}} =   \frac{\partial \m{W}^{-1} \m{b}}{\partial \m{b}} \frac{\partial g(\m{V})}{\partial \m{V}} \\
     & = \m{W}^{-\transp} \frac{\partial g(\m{V})}{\partial \m{V}} = \m{W}^{-1} \frac{\partial g(\m{V})}{\partial \m{V}}
 \end{split}
 \label{eq:grad_over_b}
 \end{equation}

\begin{equation}
 \begin{split}
     \frac{\partial g(\m{V})}{\partial \vectorize{\m W}} & = \frac{\partial \m{V}}{\partial \vectorize{\m W}} \frac{\partial g(\m{V})}{\partial \m{V}} = \frac{\m W ^{-1}\m b}{\partial \vectorize{\m W}} \frac{\partial g(\m{V})}{\partial \m{V}} \\ 
     & = \frac{ \partial \vectorize{\m W^{-1}}}{\partial \vectorize{\m W}} \frac{\partial \m W ^{-1}\m b}{\partial \vectorize{\m W^{-1}}} \frac{\partial g(\m{V})}{\partial \m{V}} \\
     & = \frac{ \partial \vectorize{\m W^{-1}}}{\partial \vectorize{\m W}} \frac{\partial \m W ^{-1}\m b}{\partial \vectorize{\m W^{-1}}} \frac{\partial g(\m{V})}{\partial \m{V}} \\
     & = \frac{ \partial \vectorize{\m W^{-1}}}{\partial \vectorize{\m W}} \frac{\partial (\m b^\transp \otimes \m{I}) \vectorize{\m W ^ {-1}}}{\partial \vectorize{\m W^{-1}}} \frac{\partial g(\m{V})}{\partial \m{V}} \\
     & = (- \m W^{-1} \otimes \m W^{-1})^\transp (\m b^\transp \otimes \m I)^\transp \frac{\partial g(\m{V})}{\partial \m{V}} \\
     & = (- \m W^{-1} \otimes \m W^{-1}) (\m b \otimes \m I) \frac{\partial g(\m{V})}{\partial \m{V}} \\
     & = - ( \m V \otimes \m W^{-1}) \frac{\partial g(\m{V})}{\partial \m{V}} = - \m V \otimes \frac{\partial g(\m{V})}{\partial \m{b}}
 \end{split}
 \end{equation}
 
 As such,
 \begin{equation}
 \begin{split}
     \frac{\partial g(\m{V})}{\partial \m W} =  - \frac{\partial g(\m{V})}{\partial \m{b}} \m V^\transp
 \end{split}
 \end{equation}

After calculating the gradients of the linear solver, the final step is the computation of the gradients with respect to the handle position regression $\tilde{\m H}$ and learnable dependency matrix $\m A$. The gradient of the first quantity is straight forward to calculate using Equation~\eqref{eq:grad_over_b}.

\begin{equation}
\begin{split}
     \frac{\partial g(\m{V})}{\partial \m{\tilde{H}}} & = \frac{\partial \m{b}}{\partial \m{\tilde{H}}} \frac{\partial g(\m{V})}{\partial \m{b}} = \frac{\partial (\m{L}^{\transp} \m L \temp + \m{A}^{\transp} \m{\tilde{H}})}{\partial \m{\tilde{H}}} \frac{\partial g(\m{V})}{\partial \m{b}}  \\
     & = \m{A} \frac{\partial g(\m{V})}{\partial \m{b}} 
\end{split}
\end{equation}
 
Lastly we provide the gradient with respect to the learnable dependency matrix $\m A$ 
\begin{equation}
     \frac{\partial g(\m{V})}{\partial \vectorize{\m A}} = \frac{\partial \m{b}}{\partial \vectorize{\m A}} \frac{\partial g(\m{V})}{\partial \m{b}} + \frac{\partial \vectorize{\m{W}}}{\partial \vectorize{\m A}} \frac{\partial g(\m{V})}{\partial \vectorize{\m{W}}} 
 \end{equation}

\begin{equation}
 \begin{split}
     \frac{\partial  \m b}{\partial \vectorize{\m A}} & = \frac{\partial (\m L^\transp \m L \temp+\m A^\transp \tilde{\m H})}{\partial \vectorize{\m A}} 
     = \frac{\partial \vectorize{\m A^\transp \tilde{\m H}}}{\partial \vectorize{\m A}} \\
     & = \frac{\partial \vectorize{\m A^\transp}}{\partial \vectorize{\m A}} \frac{\partial (\tilde{\m H} \otimes \m I)\vectorize{\m A^\transp }}{\partial \vectorize{\m A^\transp}} \\
     & = \frac{\partial \m T_{K,N}\vectorize{\m A}}{\partial \vectorize{\m A}} (\tilde{\m H}^\transp \otimes \m I)^\transp = \m T_{N,K}(\tilde{\m H} \otimes \m I) 
 \end{split}
 \end{equation}

\begin{equation}
 \begin{split}
     \frac{\partial  \vectorize{\m W}}{\partial \vectorize{\m A}} & = \frac{\partial (\m L^\transp \m L+\m A^\transp \m A^)}{\partial \vectorize{\m A}} 
     = \frac{\partial \vectorize{\m A^\transp \m A}}{\partial \vectorize{\m A}} \\
     & = \m I_{N} \otimes \m A^\transp + (\m A^T \otimes \m I_{N}) \m T_{K,N}
 \end{split}
 \end{equation}
 
\begin{figure*}[t]
\captionsetup[subfigure]{position=bottom}
\begin{centering}
\subfloat{\includegraphics[width=\textwidth]{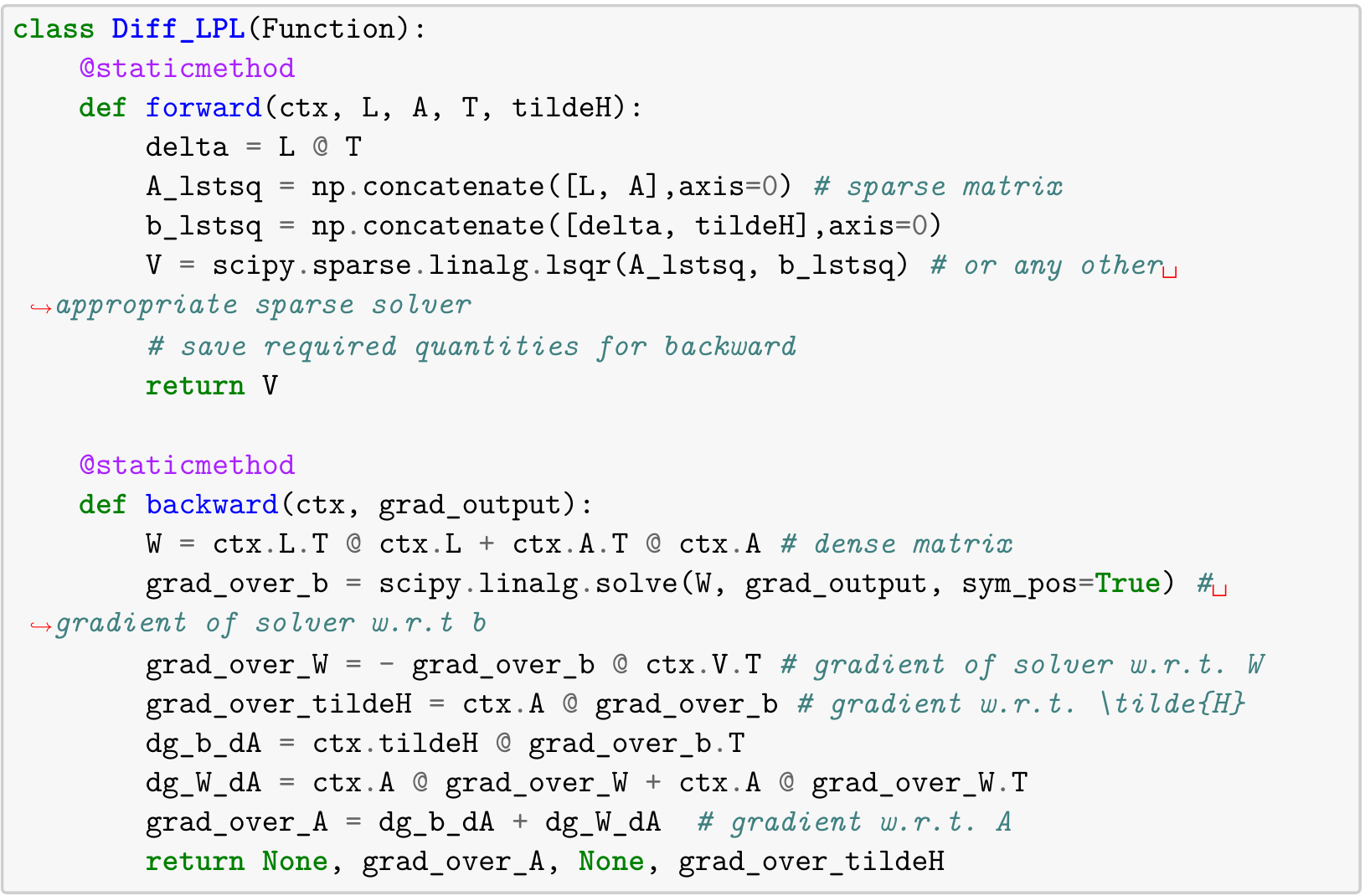}}~
\caption{PyTorch implementation of the proposed Differentiable Laplacian Solver.\label{fig:code}}
\end{centering}
\end{figure*}

In Figure~\ref{fig:code} we provide a PyTorch implementation of the differentiable Laplacian deformation module which has been thoroughly gradient checked. We omit some PyTorch related boilerplate code for clarity. 

\section{PCA on deformations}

\begin{figure*}[t]
\captionsetup[subfigure]{position=bottom}
\begin{centering}
\subfloat[Fox]{\includegraphics[width=.3\textwidth]{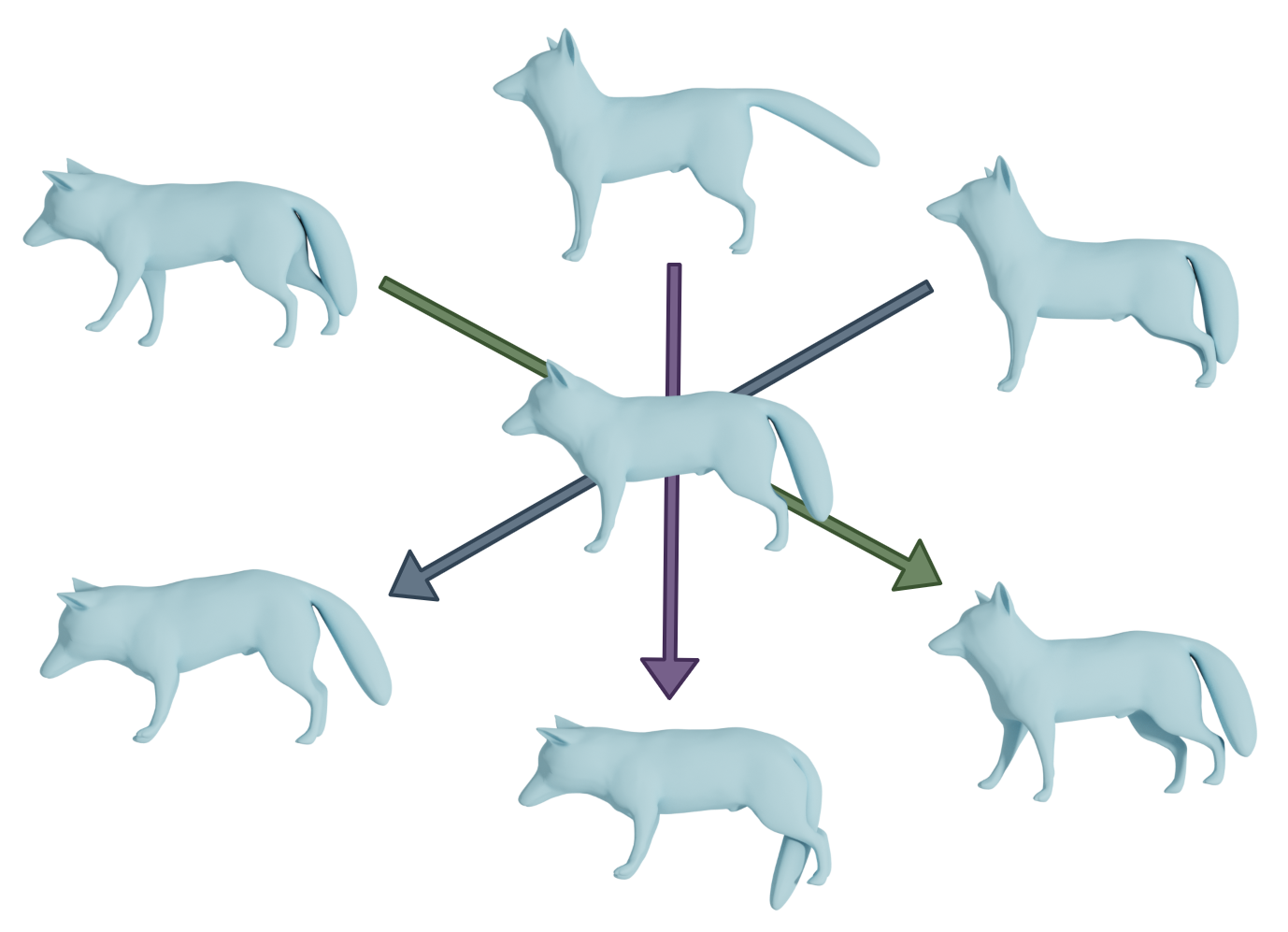}}~
\subfloat[Cow]{\includegraphics[width=.3\textwidth]{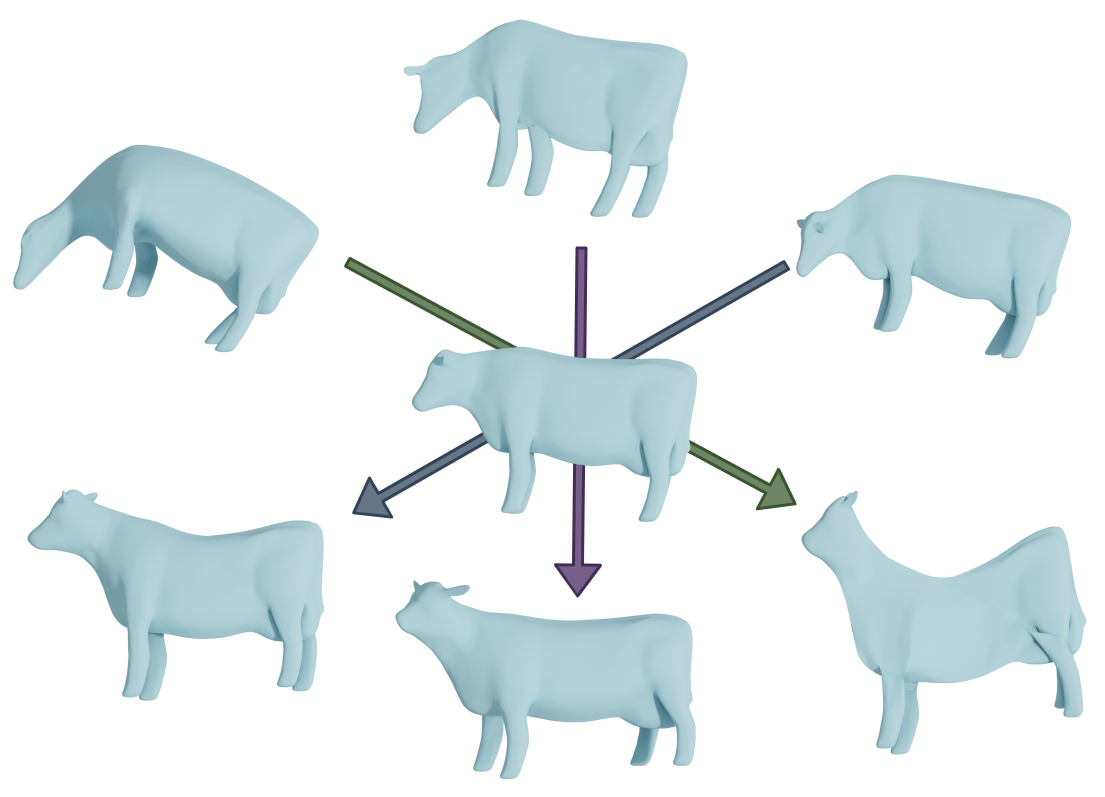}}~
\subfloat[Horse]{\includegraphics[width=.3\textwidth]{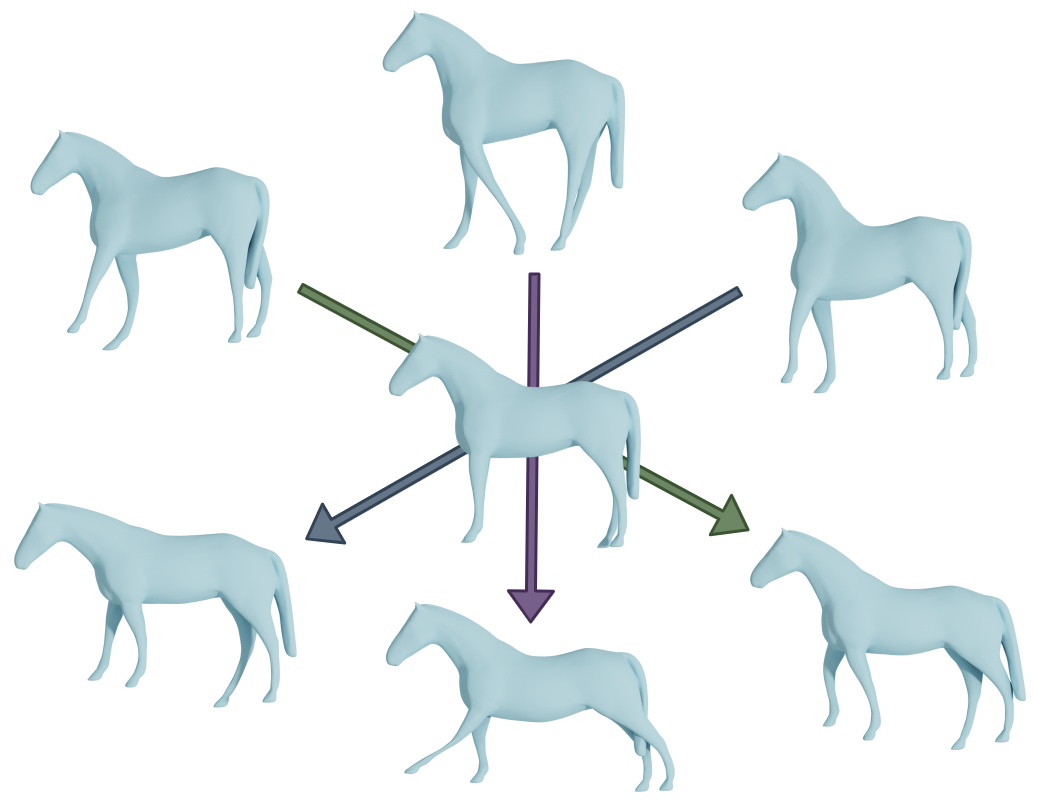}}~\\ 
\vspace{-0.35cm}
\subfloat[Tiger]{\includegraphics[width=.3\textwidth]{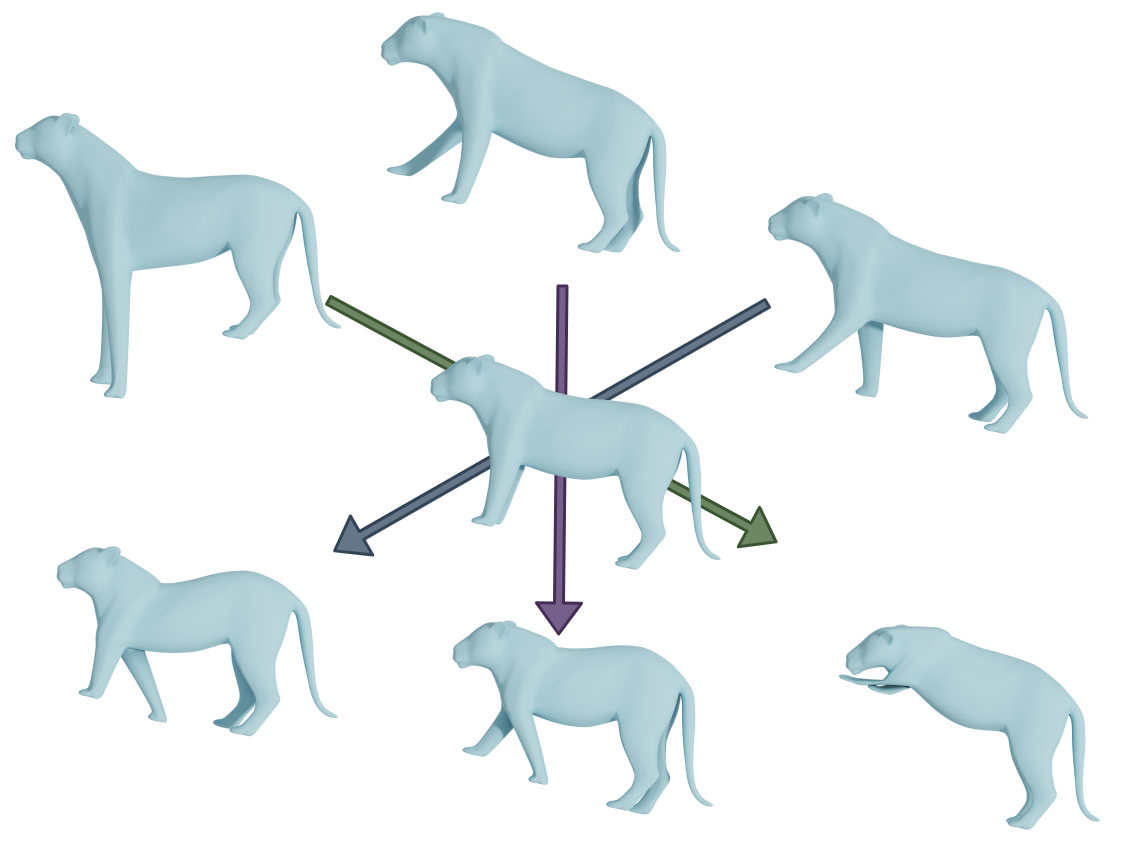}}~
\subfloat[Zebra]{\includegraphics[width=.3\textwidth]{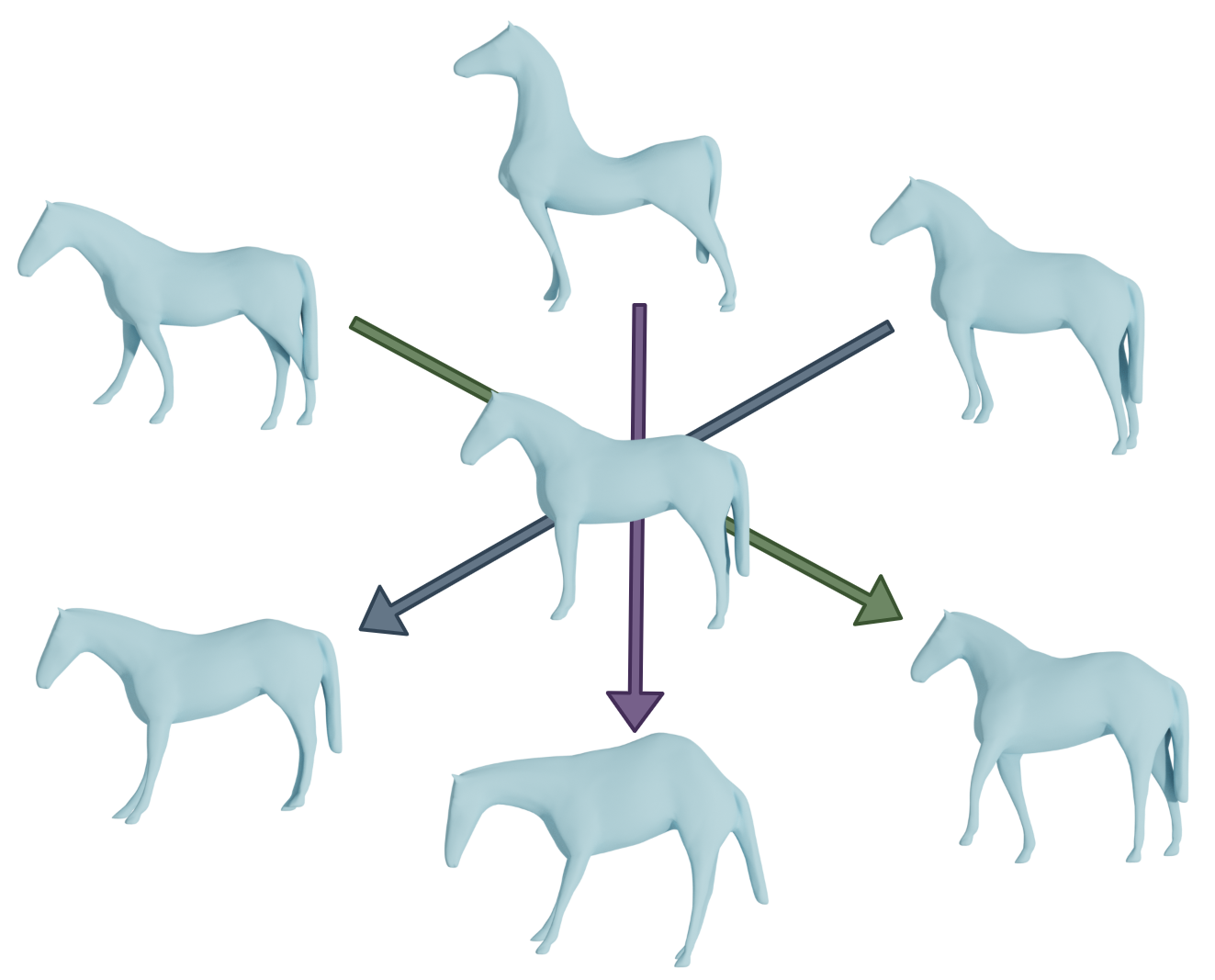}}~
\subfloat[Deer]{\includegraphics[width=.3\textwidth]{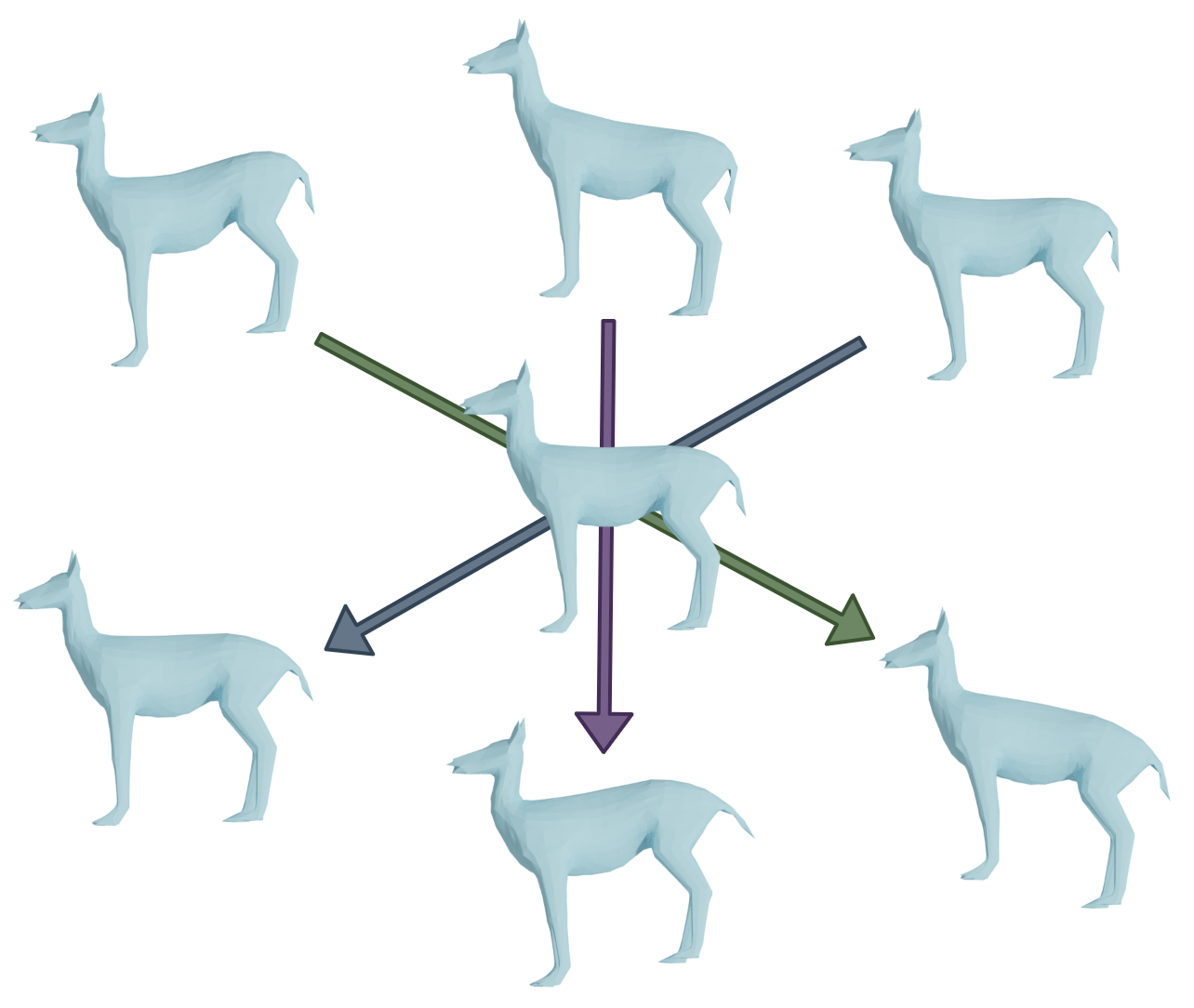}}~\\ 
\vspace{-0.35cm}
\subfloat[Birds]{\includegraphics[width=.3\textwidth]{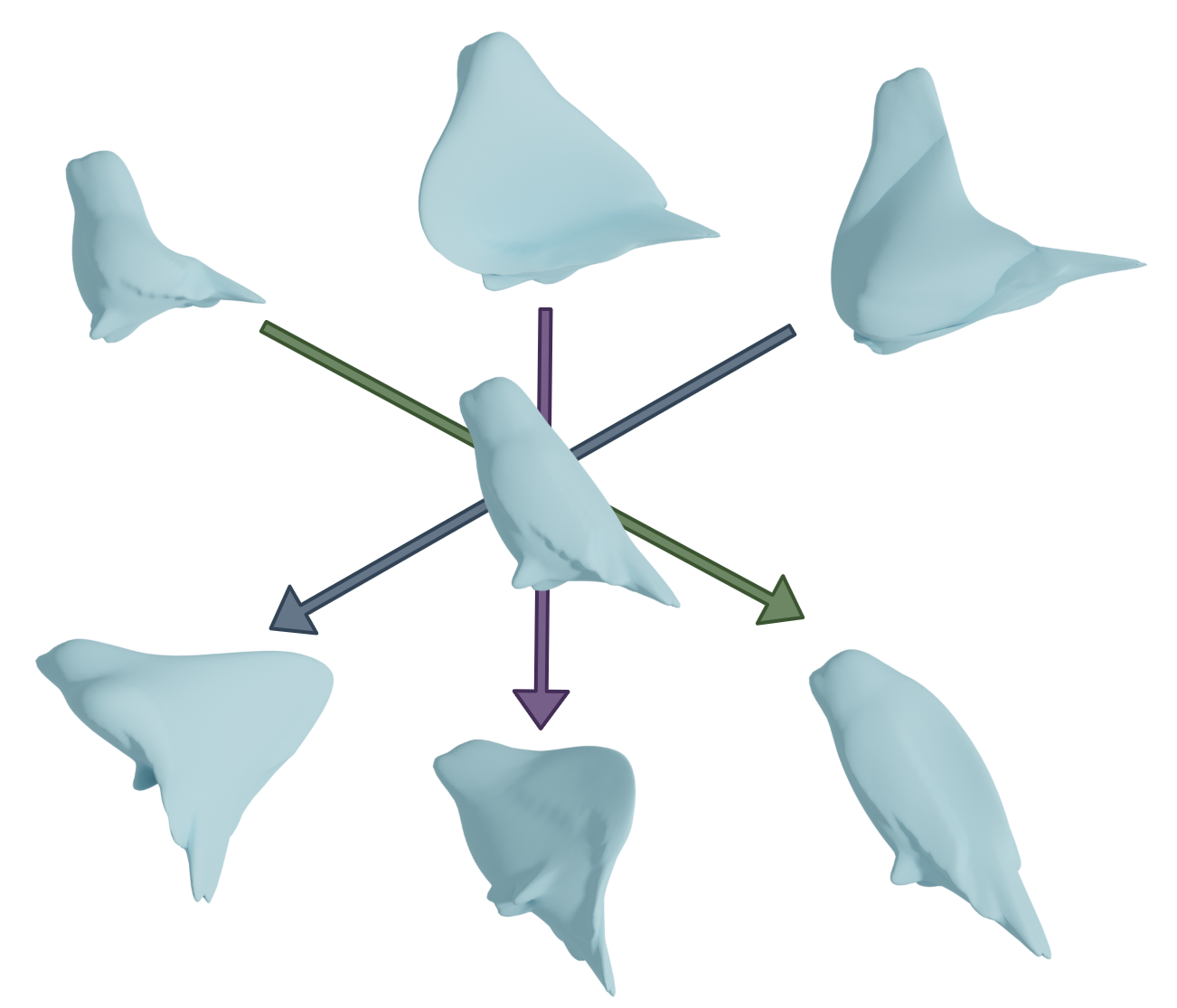}}~
\subfloat[Bear]{\includegraphics[width=.3\textwidth]{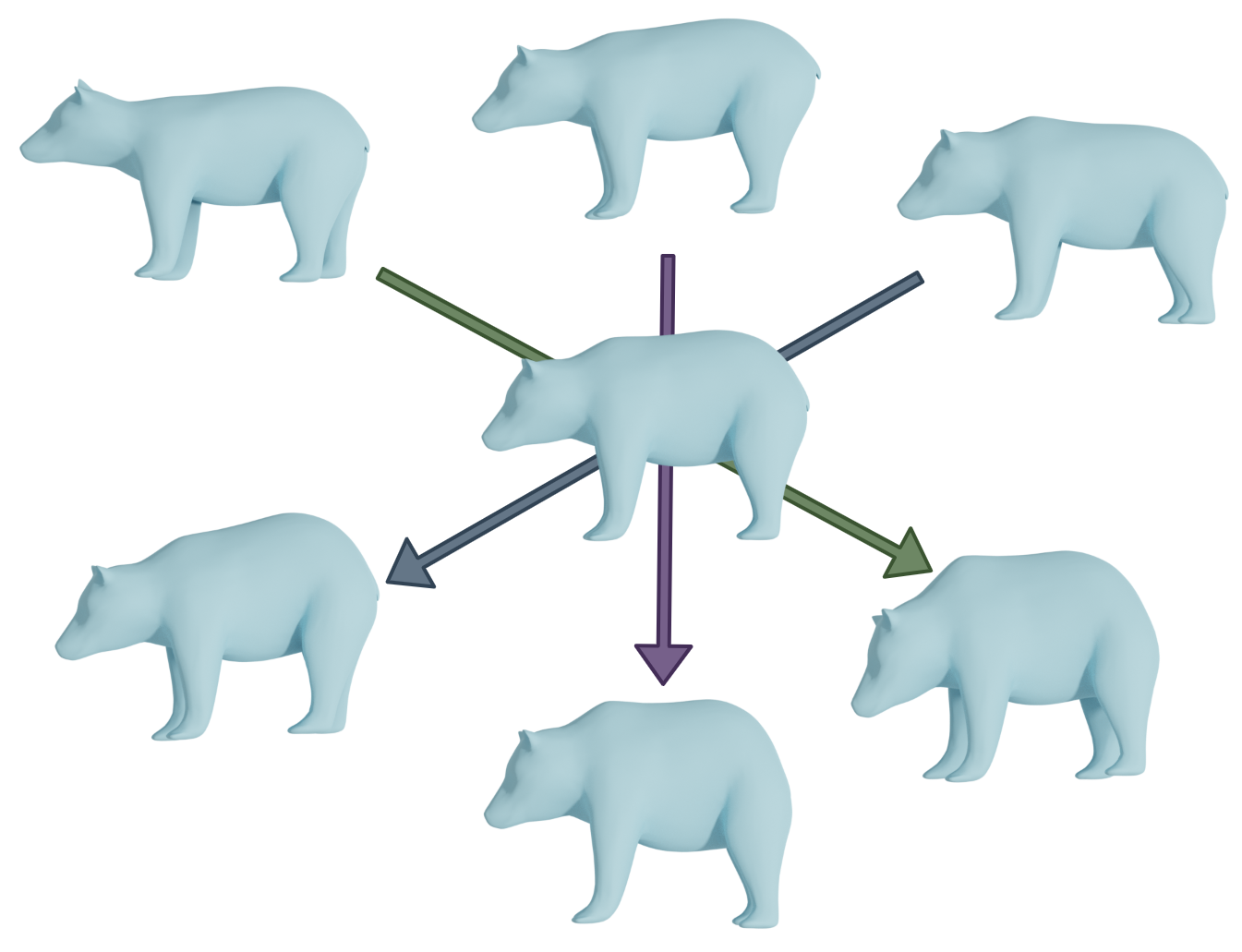}}~
\subfloat[Giraffe]{\includegraphics[width=.3\textwidth]{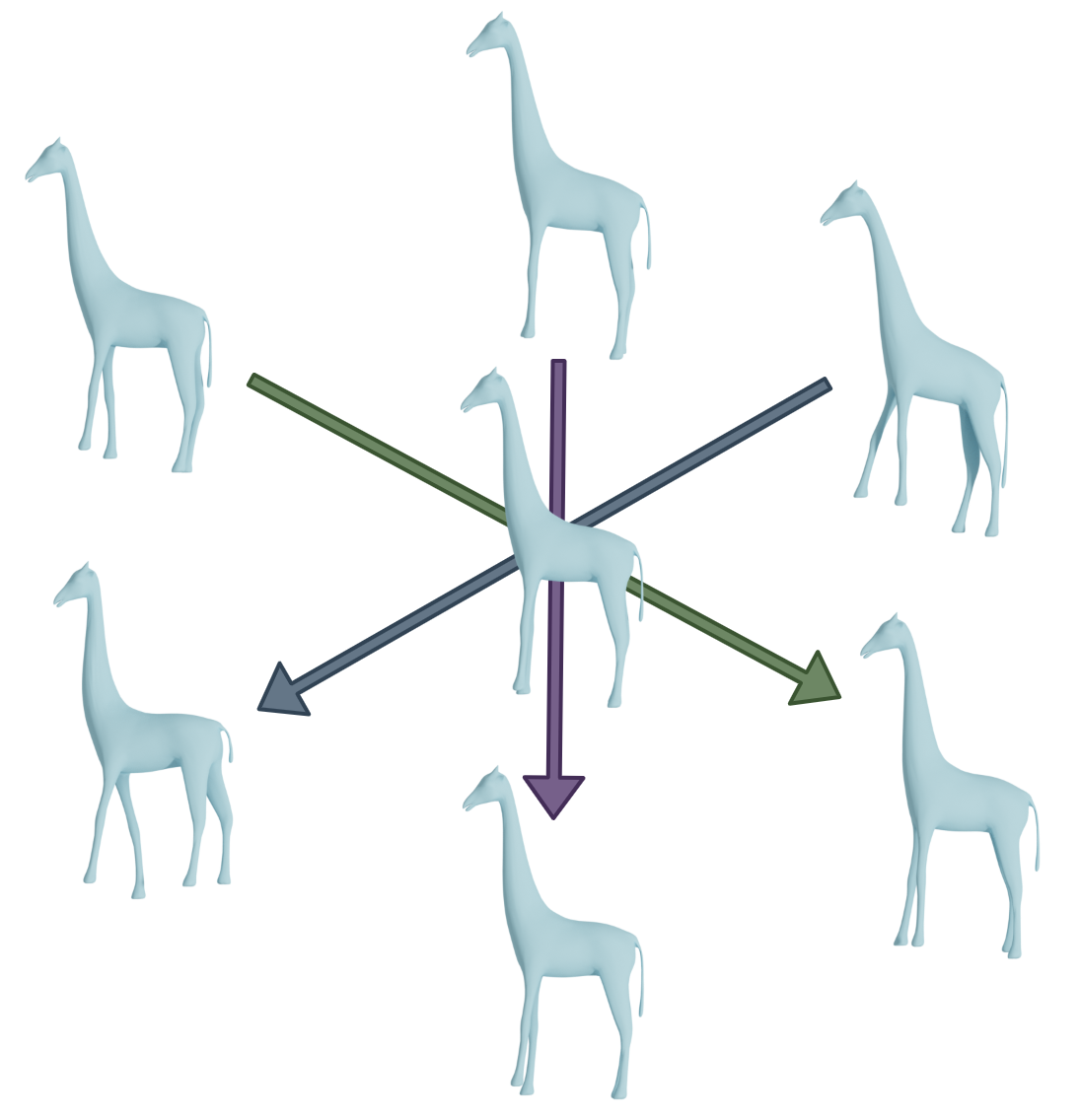}}~\\ 
\vspace{-0.3cm}
\caption{Visualization of the predicted deformations for several objects by depicting the mean shape in the center and the first 3 modes obtained by PCA on the handle estimates obtained across the dataset.\vspace{-0.3cm}\label{fig:pcas}}
\end{centering}
\end{figure*}

In Figure~\ref{fig:pcas} we visualize the learned deformations for a wide range of articulated objects. The visualization is created by running PCA for each object on all 3D reconstructions obtained on training and test dataset. The mean shape is depicted in the center while the first three PCA axis are visualized alongside the mean shape. The visualizations shows some interesting deformations across objects. In all cases the deformations capture clearly movements of legs, the tail and head.

\section{More Results}

\subsection{Comparisons on Horses}
\begin{figure*}[t]
\captionsetup[subfigure]{position=bottom}
\begin{centering}
\subfloat{\includegraphics[width=.9\textwidth]{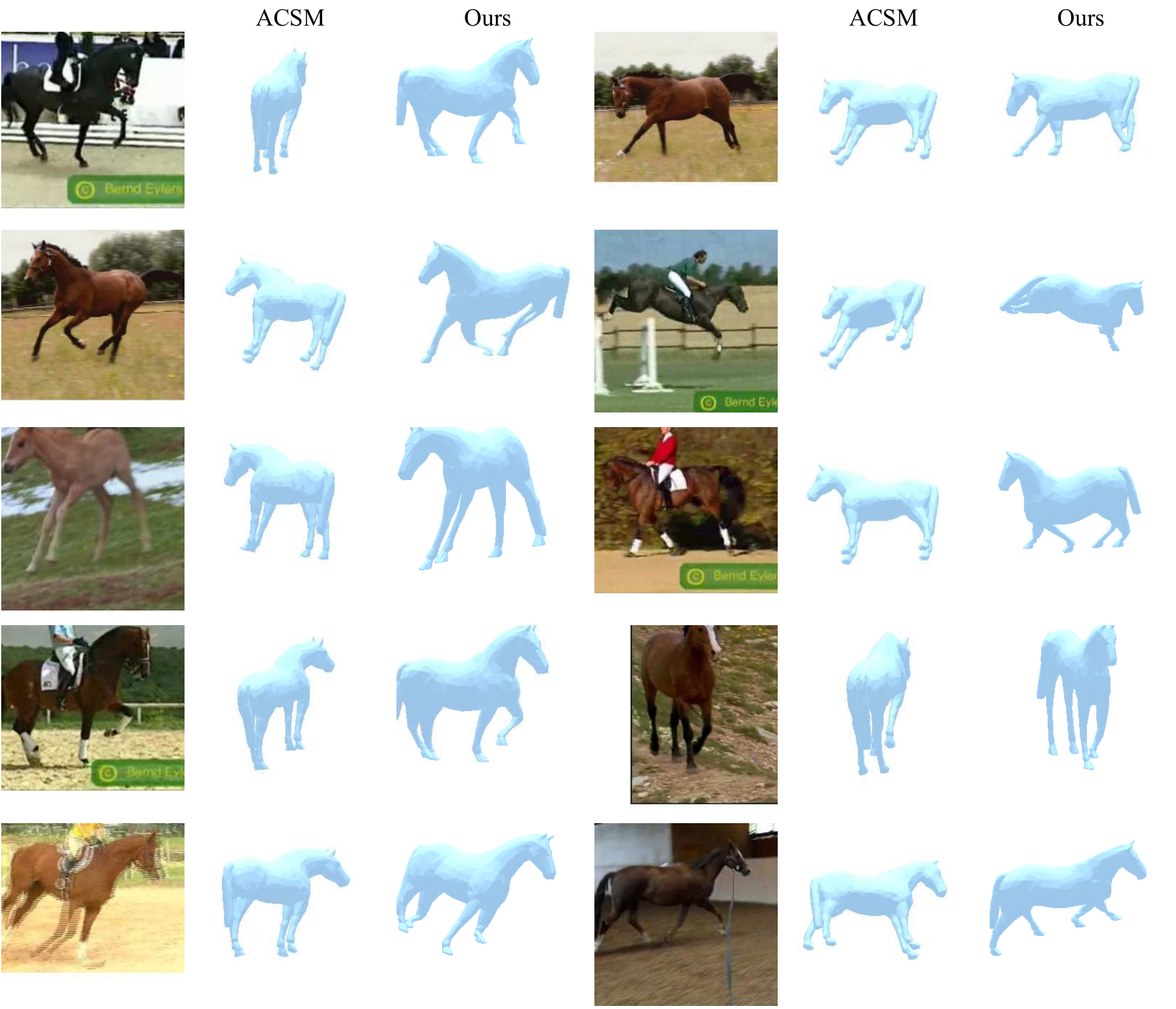}}~
\vspace{-0.35cm}
\caption{Reconstruction comparisons of our method and ACSM~\cite{kulkarni2020articulation} for the object horse.\label{fig:horses}}
\end{centering}
\end{figure*}

In Figure~\ref{fig:horses} we provide comparisons against prior articulated reconstruction work~\cite{kulkarni2020articulation}. Note that both methods used identical template mesh. The proposed method is capable of achieving realisting deformations without the requirement of manual part segmentation as a result of the proposed differentiable deformation module. We also include several videos of horses for extra comparisons. The videos demonstrate that our method provides camera predictions and deformations that are robust across frames and match closely the movements of the depicted object. 

\subsection{Comparisons on CUB}
\begin{figure*}[t]
\captionsetup[subfigure]{position=bottom}
\begin{centering}
\subfloat{\includegraphics[width=\textwidth]{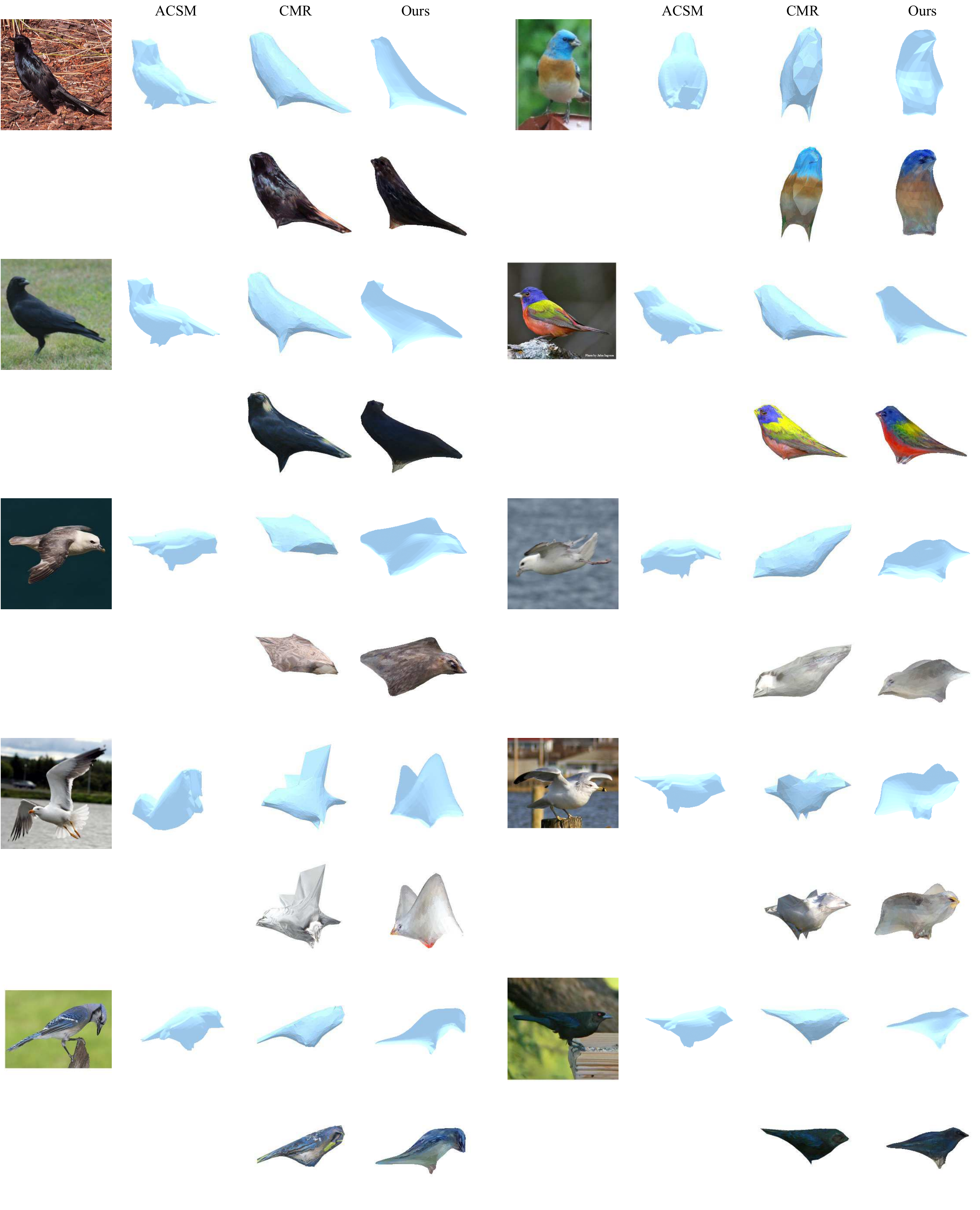}}~
\caption{\textbf{Bird Reconstructions:} Qualitative comparisons between our method, CMR~\cite{cmrKanazawa18} and ACSM~\cite{kulkarni2020articulation} with images from the CUB test set. \label{fig:birds}}
\end{centering}
\end{figure*}

In Figure~\ref{fig:birds} we provide comparisons against prior work on common training supervision. It is apparent that our method is capable of correctly deforming the template mesh to produce highly flexible wings or bending and turning the body and head of the birds. In nearly all results, ACSM lacks the necessary 3D shape deformation which is caused by the segmentation of the template shape in 3 parts (head, body and tail). CSM is providing results with better deformations to ACSM, however as it can be seen almost all open wing results are obtained with the offset of a small set of points which causes uneven surfaces (for example Row 4 left). Our method due to the Laplacian based Deformation produces always meshes with well allocated vertices. Furthermore, our results exhibit interesting features like rotation of the head (Row 2-left, Row 1-right), open wing formations and bending of the beak (Row 5-left), i.e. pecking.

\subsection{More Results}
\begin{figure*}[t]
 \captionsetup[subfigure]{labelformat=empty, position=top}
 \begin{centering}
 \setcounter{subfigure}{0}

 \subfloat{\includegraphics[width=.12\textwidth]{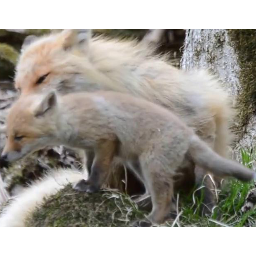}}~
 \subfloat{\includegraphics[width=.12\textwidth]{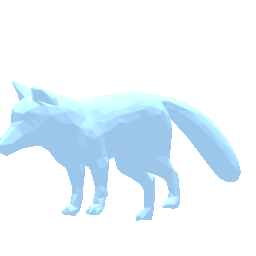}}~
 \subfloat{\includegraphics[width=.12\textwidth]{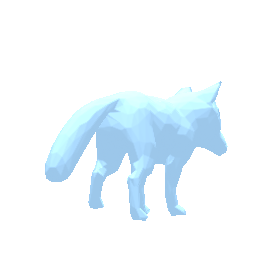}}~
 \subfloat{\includegraphics[width=.12\textwidth]{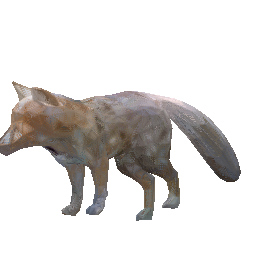}}~
 \subfloat{\includegraphics[width=.12\textwidth]{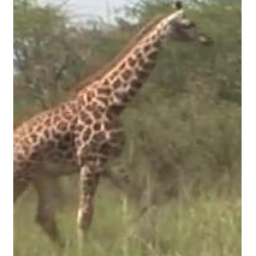}}~
 \subfloat{\includegraphics[width=.12\textwidth]{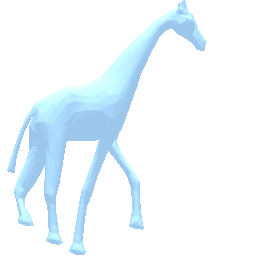}}~
 \subfloat{\includegraphics[width=.12\textwidth]{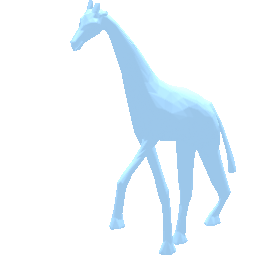}}~
 \subfloat{\includegraphics[width=.12\textwidth]{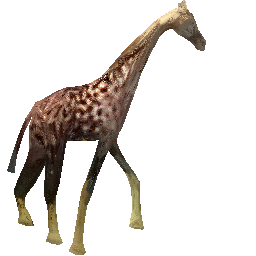}}~\\ 
 \vspace{-0.35cm}
 \subfloat{\includegraphics[width=.12\textwidth]{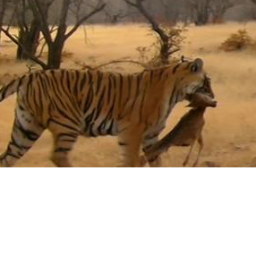}}~
 \subfloat{\includegraphics[width=.12\textwidth]{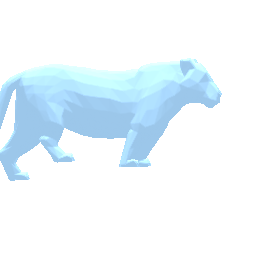}}~
 \subfloat{\includegraphics[width=.12\textwidth]{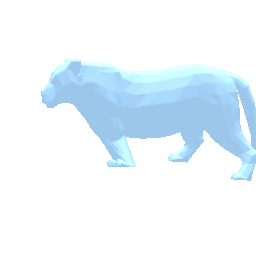}}~
 \subfloat{\includegraphics[width=.12\textwidth]{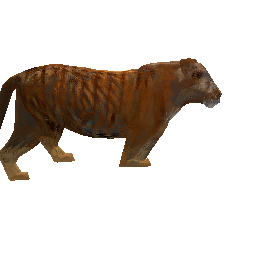}}~
 \subfloat{\includegraphics[width=.12\textwidth]{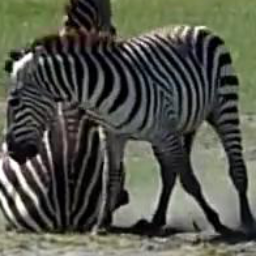}}~
 \subfloat{\includegraphics[width=.12\textwidth]{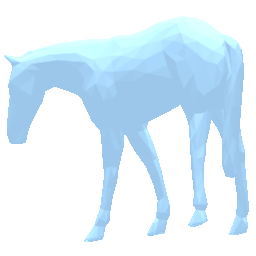}}~
 \subfloat{\includegraphics[width=.12\textwidth]{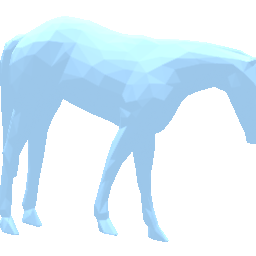}}~
 \subfloat{\includegraphics[width=.12\textwidth]{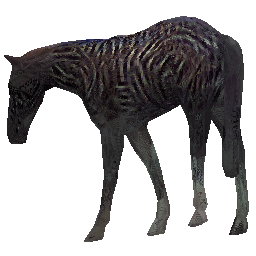}}~\\ 
 \vspace{-0.35cm}
 \subfloat{\includegraphics[width=.12\textwidth]{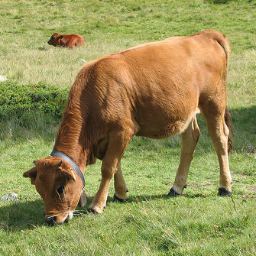}}~
 \subfloat{\includegraphics[width=.12\textwidth]{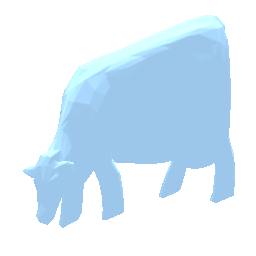}}~
 \subfloat{\includegraphics[width=.12\textwidth]{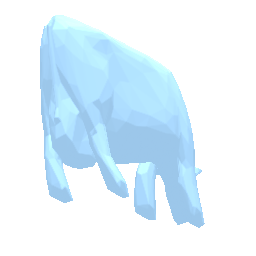}}~
 \subfloat{\includegraphics[width=.12\textwidth]{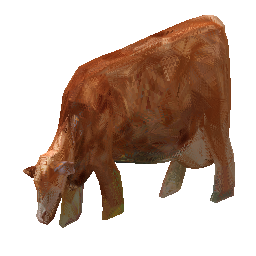}}~
 \subfloat{\includegraphics[width=.12\textwidth]{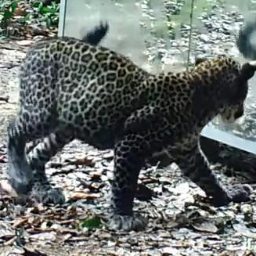}}~
 \subfloat{\includegraphics[width=.12\textwidth]{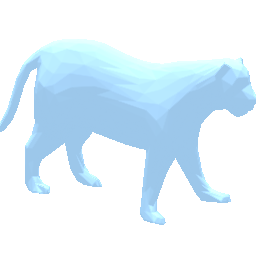}}~
 \subfloat{\includegraphics[width=.12\textwidth]{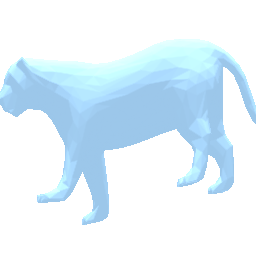}}~
 \subfloat{\includegraphics[width=.12\textwidth]{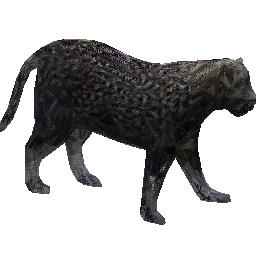}}~\\
 \vspace{-0.35cm}
 \subfloat{\includegraphics[width=.12\textwidth]{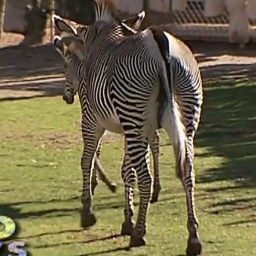}}~
 \subfloat{\includegraphics[width=.12\textwidth]{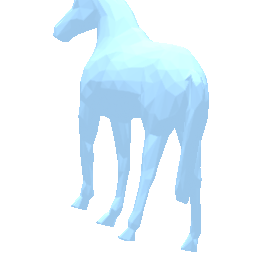}}~
 \subfloat{\includegraphics[width=.12\textwidth]{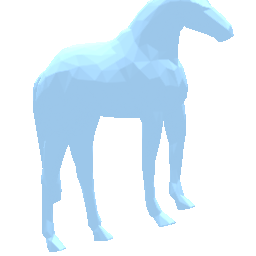}}~
 \subfloat{\includegraphics[width=.12\textwidth]{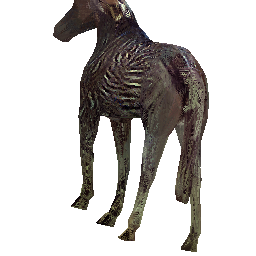}}~
 \subfloat{\includegraphics[width=.12\textwidth]{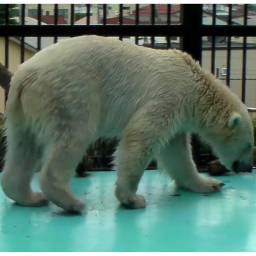}}~
 \subfloat{\includegraphics[width=.12\textwidth]{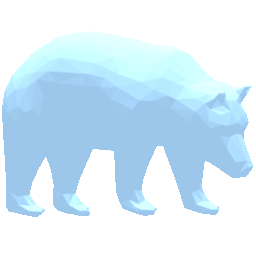}}~
 \subfloat{\includegraphics[width=.12\textwidth]{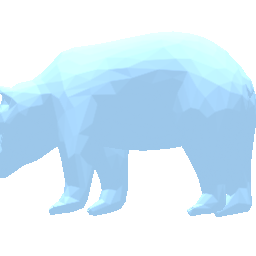}}~
 \subfloat{\includegraphics[width=.12\textwidth]{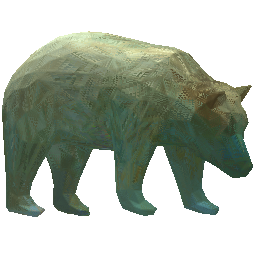}}~\\ 
 \vspace{-0.35cm}
 \subfloat{\includegraphics[width=.12\textwidth]{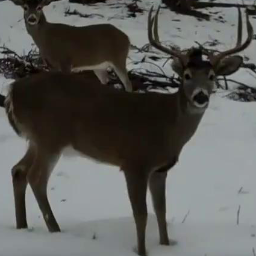}}~
 \subfloat{\includegraphics[width=.12\textwidth]{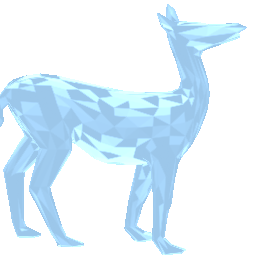}}~
 \subfloat{\includegraphics[width=.12\textwidth]{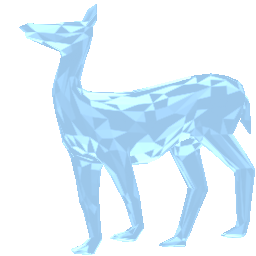}}~
 \subfloat{\includegraphics[width=.12\textwidth]{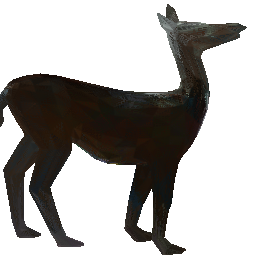}}~
 \subfloat{\includegraphics[width=.12\textwidth]{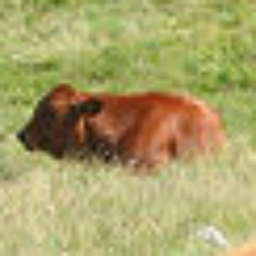}}~
 \subfloat{\includegraphics[width=.12\textwidth]{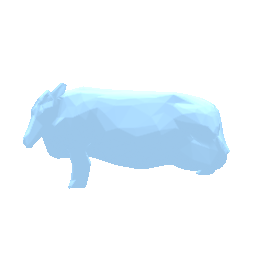}}~
 \subfloat{\includegraphics[width=.12\textwidth]{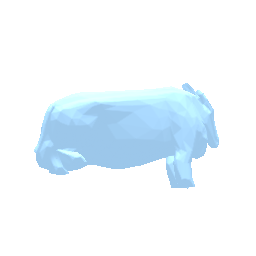}}~
 \subfloat{\includegraphics[width=.12\textwidth]{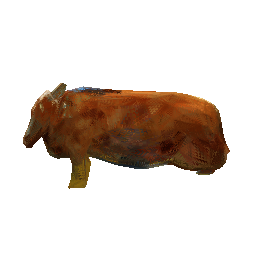}}~\\
 \vspace{-0.35cm}
 \subfloat{\includegraphics[width=.12\textwidth]{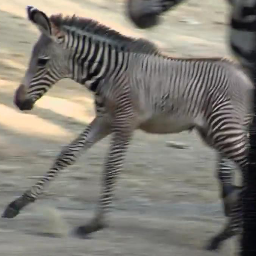}}~
 \subfloat{\includegraphics[width=.12\textwidth]{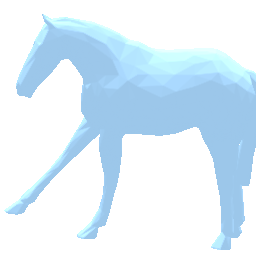}}~
 \subfloat{\includegraphics[width=.12\textwidth]{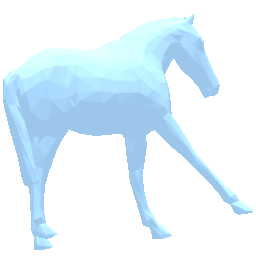}}~
 \subfloat{\includegraphics[width=.12\textwidth]{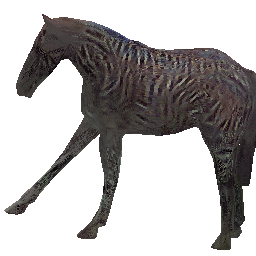}}~
 \subfloat{\includegraphics[width=.12\textwidth]{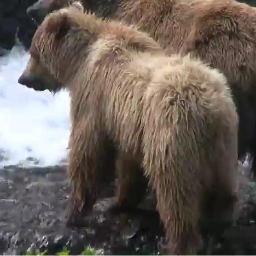}}~
 \subfloat{\includegraphics[width=.12\textwidth]{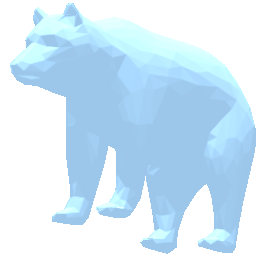}}~
 \subfloat{\includegraphics[width=.12\textwidth]{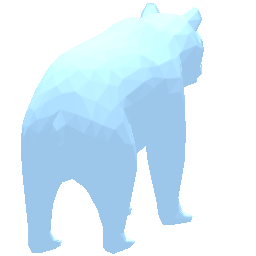}}~
 \subfloat{\includegraphics[width=.12\textwidth]{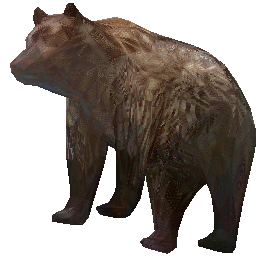}}~\\ 
 \vspace{-0.35cm}
 \subfloat{\includegraphics[width=.12\textwidth]{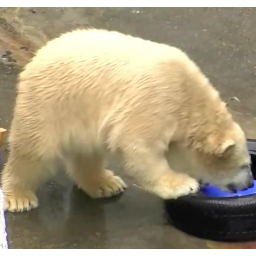}}~
 \subfloat{\includegraphics[width=.12\textwidth]{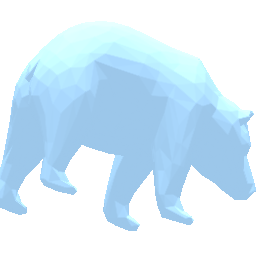}}~
 \subfloat{\includegraphics[width=.12\textwidth]{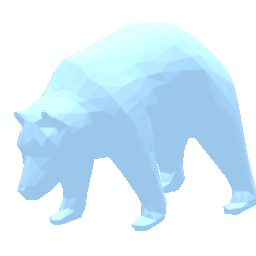}}~
 \subfloat{\includegraphics[width=.12\textwidth]{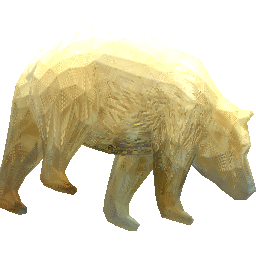}}~
 \subfloat{\includegraphics[width=.12\textwidth]{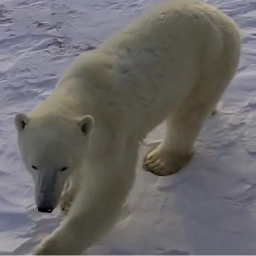}}~
 \subfloat{\includegraphics[width=.12\textwidth]{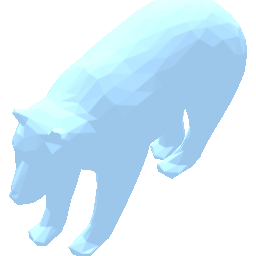}}~
 \subfloat{\includegraphics[width=.12\textwidth]{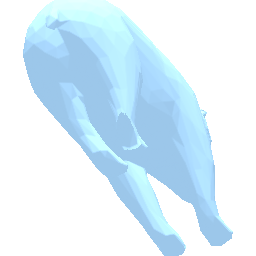}}~
 \subfloat{\includegraphics[width=.12\textwidth]{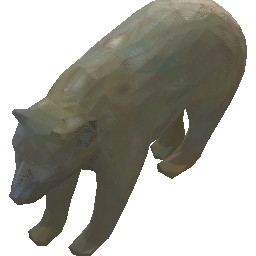}}~\\ 
 \vspace{-0.35cm}
 \subfloat{\includegraphics[width=.12\textwidth]{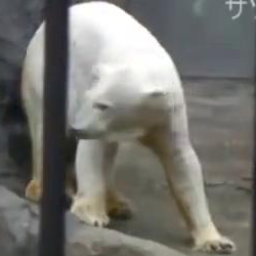}}~
 \subfloat{\includegraphics[width=.12\textwidth]{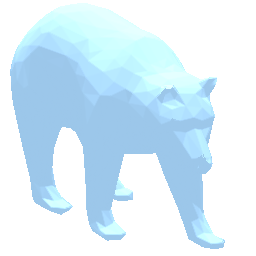}}~
 \subfloat{\includegraphics[width=.12\textwidth]{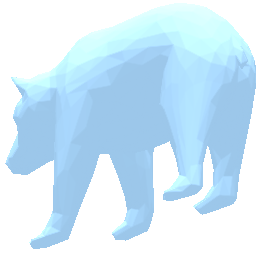}}~
 \subfloat{\includegraphics[width=.12\textwidth]{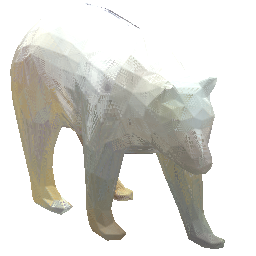}}~
 \subfloat{\includegraphics[width=.12\textwidth]{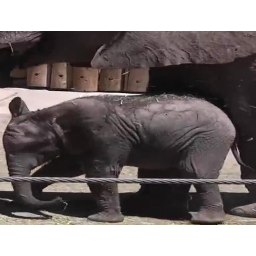}}~
 \subfloat{\includegraphics[width=.12\textwidth]{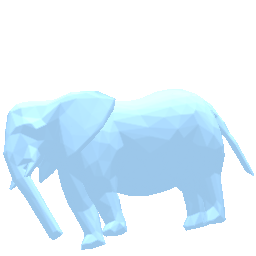}}~
 \subfloat{\includegraphics[width=.12\textwidth]{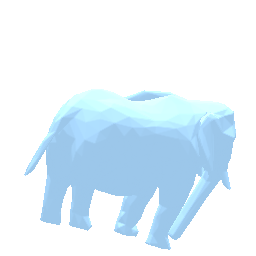}}~
 \subfloat{\includegraphics[width=.12\textwidth]{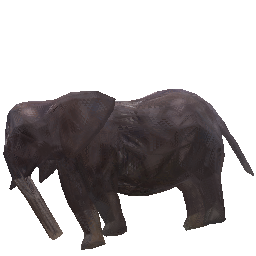}}~\\ 
 \vspace{-0.35cm}
 \subfloat{\includegraphics[width=.12\textwidth]{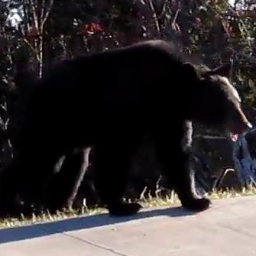}}~
 \subfloat{\includegraphics[width=.12\textwidth]{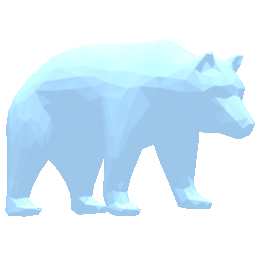}}~
 \subfloat{\includegraphics[width=.12\textwidth]{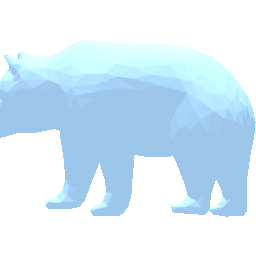}}~
 \subfloat{\includegraphics[width=.12\textwidth]{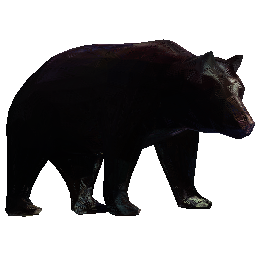}}~
 \subfloat{\includegraphics[width=.12\textwidth]{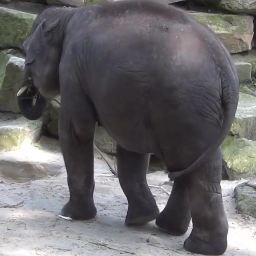}}~
 \subfloat{\includegraphics[width=.12\textwidth]{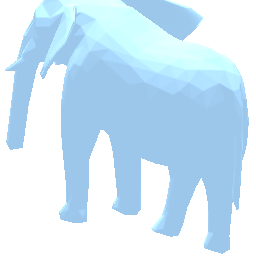}}~
 \subfloat{\includegraphics[width=.12\textwidth]{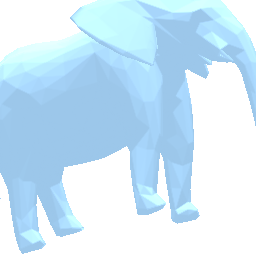}}~
 \subfloat{\includegraphics[width=.12\textwidth]{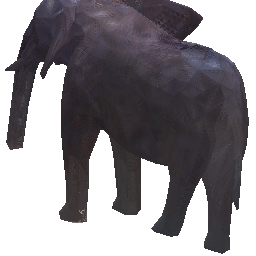}}~\\ 
 \vspace{-0.35cm}
 \subfloat{\includegraphics[width=.12\textwidth]{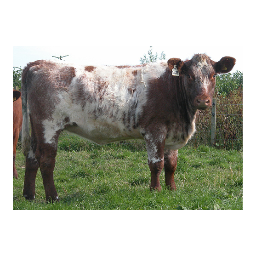}}~
 \subfloat{\includegraphics[width=.12\textwidth]{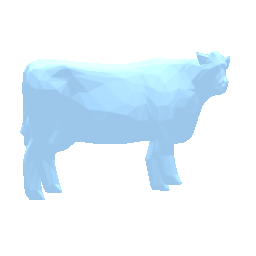}}~
 \subfloat{\includegraphics[width=.12\textwidth]{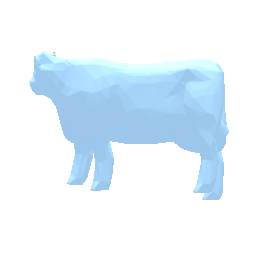}}~
 \subfloat{\includegraphics[width=.12\textwidth]{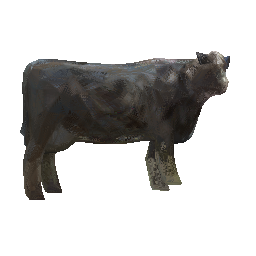}}~
 \subfloat{\includegraphics[width=.12\textwidth]{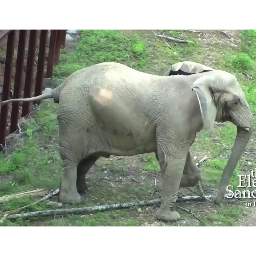}}~
 \subfloat{\includegraphics[width=.12\textwidth]{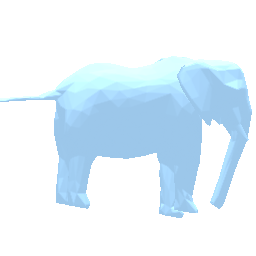}}~
 \subfloat{\includegraphics[width=.12\textwidth]{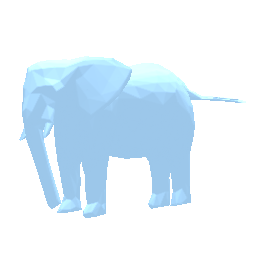}}~
 \subfloat{\includegraphics[width=.12\textwidth]{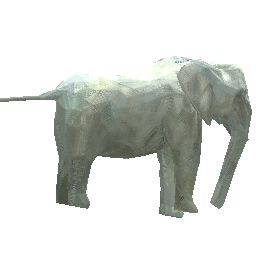}}~\\ 
 \vspace{-0.3cm}
 \caption{\textbf{3D reconstructions} We show the input image, the reconstructed shape from the predicted and a novel view and finally the predicted texture.\label{fig:2}}
 \end{centering}
 \end{figure*}

\begin{figure*}[t]
 \captionsetup[subfigure]{labelformat=empty, position=top}
 \begin{centering}
 \setcounter{subfigure}{0}

 \subfloat{\includegraphics[width=.12\textwidth]{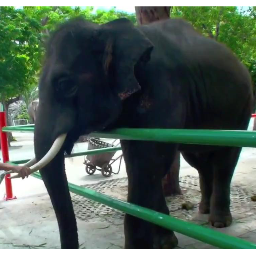}}~
 \subfloat{\includegraphics[width=.12\textwidth]{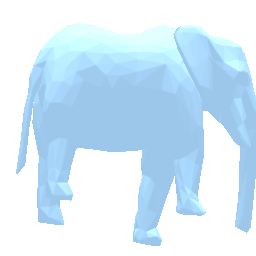}}~
 \subfloat{\includegraphics[width=.12\textwidth]{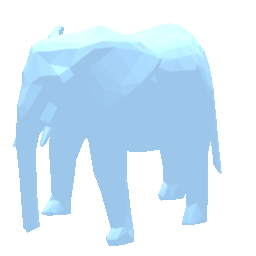}}~
 \subfloat{\includegraphics[width=.12\textwidth]{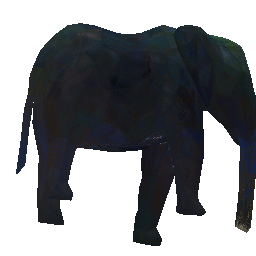}}~
 \subfloat{\includegraphics[width=.12\textwidth]{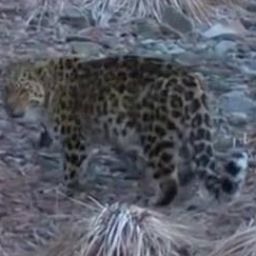}}~
 \subfloat{\includegraphics[width=.12\textwidth]{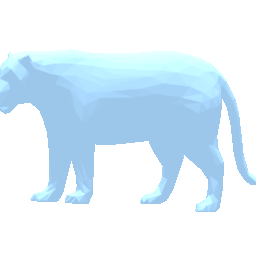}}~
 \subfloat{\includegraphics[width=.12\textwidth]{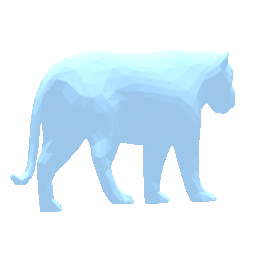}}~
 \subfloat{\includegraphics[width=.12\textwidth]{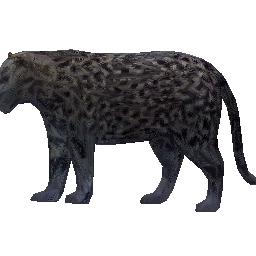}}~\\ 
 \vspace{-0.35cm}
 \subfloat{\includegraphics[width=.12\textwidth]{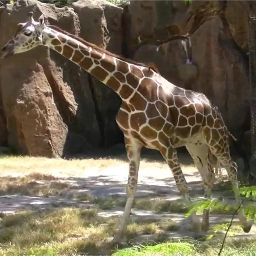}}~
 \subfloat{\includegraphics[width=.12\textwidth]{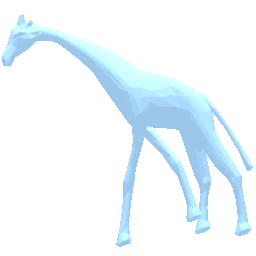}}~
 \subfloat{\includegraphics[width=.12\textwidth]{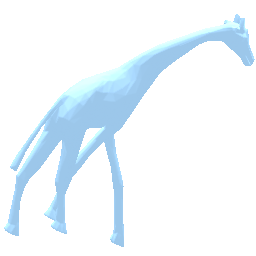}}~
 \subfloat{\includegraphics[width=.12\textwidth]{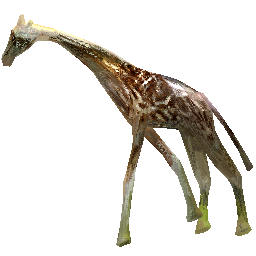}}~
 \subfloat{\includegraphics[width=.12\textwidth]{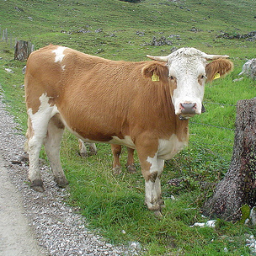}}~
 \subfloat{\includegraphics[width=.12\textwidth]{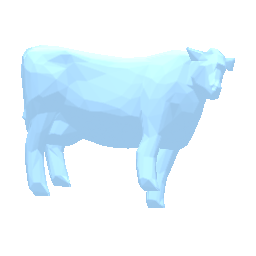}}~
 \subfloat{\includegraphics[width=.12\textwidth]{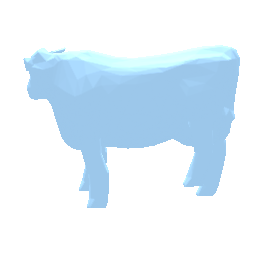}}~
 \subfloat{\includegraphics[width=.12\textwidth]{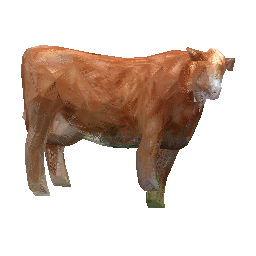}}~\\
 \vspace{-0.35cm}
 \subfloat{\includegraphics[width=.12\textwidth]{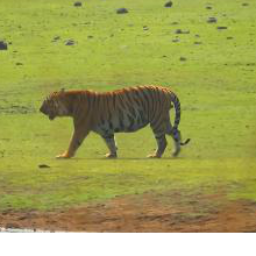}}~
 \subfloat{\includegraphics[width=.12\textwidth]{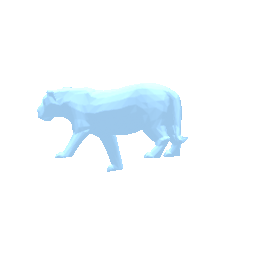}}~
 \subfloat{\includegraphics[width=.12\textwidth]{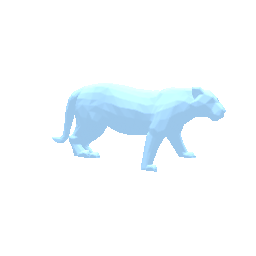}}~
 \subfloat{\includegraphics[width=.12\textwidth]{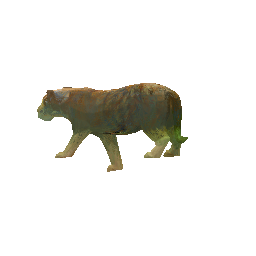}}~
 \subfloat{\includegraphics[width=.12\textwidth]{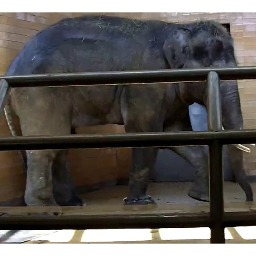}}~
 \subfloat{\includegraphics[width=.12\textwidth]{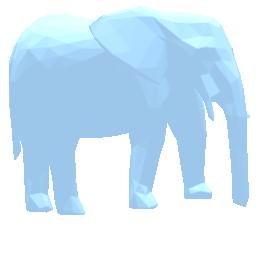}}~
 \subfloat{\includegraphics[width=.12\textwidth]{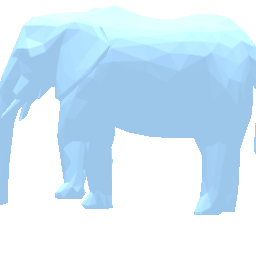}}~
 \subfloat{\includegraphics[width=.12\textwidth]{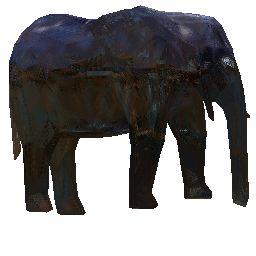}}~\\ 
 \vspace{-0.35cm}
 \subfloat{\includegraphics[width=.12\textwidth]{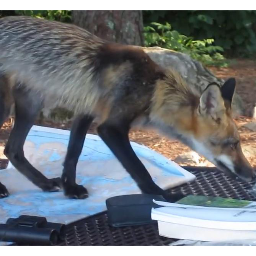}}~
 \subfloat{\includegraphics[width=.12\textwidth]{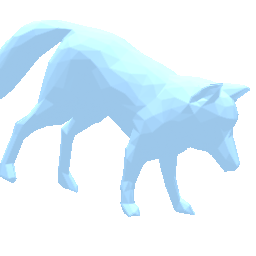}}~
 \subfloat{\includegraphics[width=.12\textwidth]{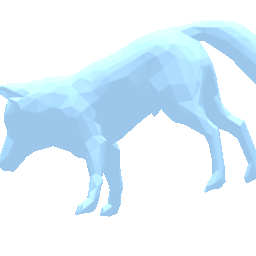}}~
 \subfloat{\includegraphics[width=.12\textwidth]{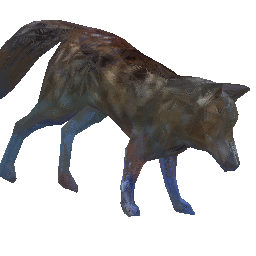}}~
 \subfloat{\includegraphics[width=.12\textwidth]{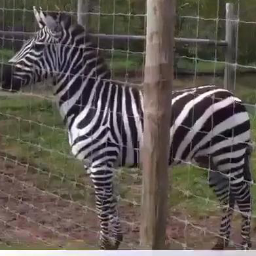}}~
 \subfloat{\includegraphics[width=.12\textwidth]{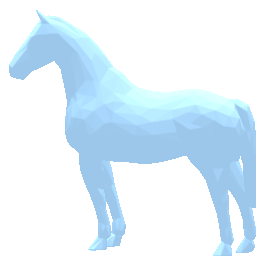}}~
 \subfloat{\includegraphics[width=.12\textwidth]{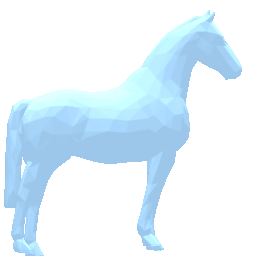}}~
 \subfloat{\includegraphics[width=.12\textwidth]{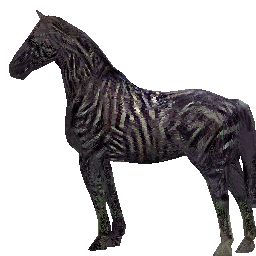}}~\\
 \vspace{-0.35cm}
 \subfloat{\includegraphics[width=.12\textwidth]{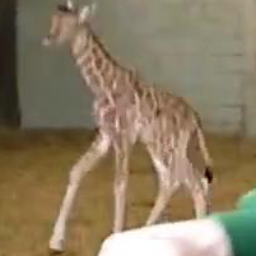}}~
 \subfloat{\includegraphics[width=.12\textwidth]{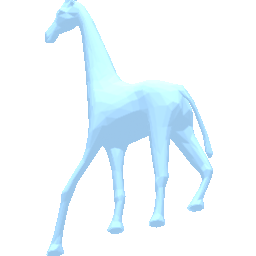}}~
 \subfloat{\includegraphics[width=.12\textwidth]{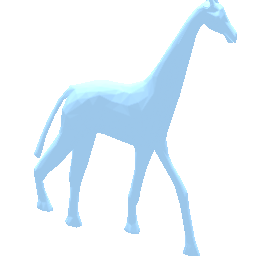}}~
 \subfloat{\includegraphics[width=.12\textwidth]{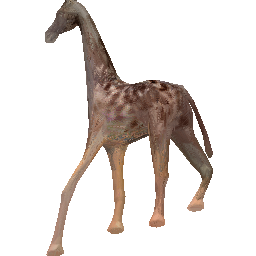}}~
 \subfloat{\includegraphics[width=.12\textwidth]{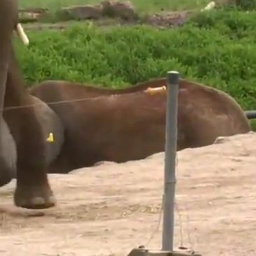}}~
 \subfloat{\includegraphics[width=.12\textwidth]{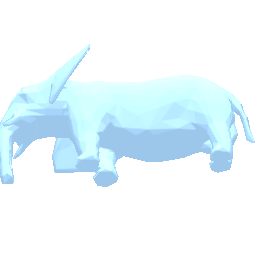}}~
 \subfloat{\includegraphics[width=.12\textwidth]{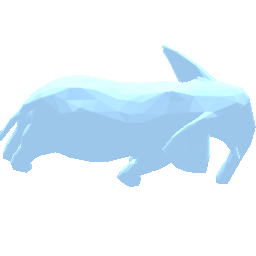}}~
 \subfloat{\includegraphics[width=.12\textwidth]{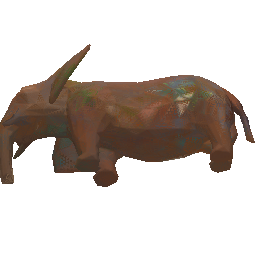}}~\\ 
 \vspace{-0.35cm}
 \subfloat{\includegraphics[width=.12\textwidth]{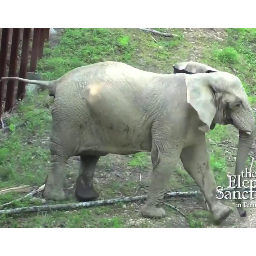}}~
 \subfloat{\includegraphics[width=.12\textwidth]{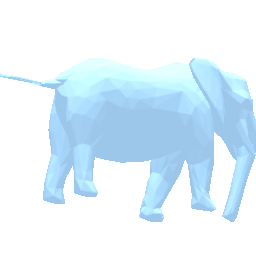}}~
 \subfloat{\includegraphics[width=.12\textwidth]{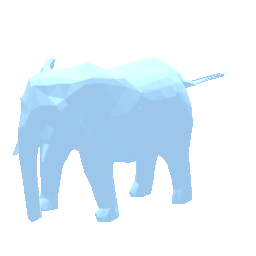}}~
 \subfloat{\includegraphics[width=.12\textwidth]{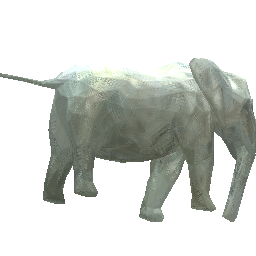}}~
 \subfloat{\includegraphics[width=.12\textwidth]{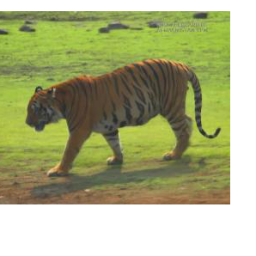}}~
 \subfloat{\includegraphics[width=.12\textwidth]{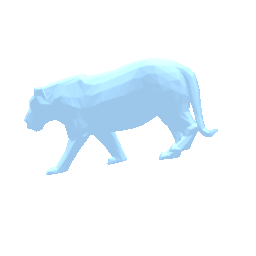}}~
 \subfloat{\includegraphics[width=.12\textwidth]{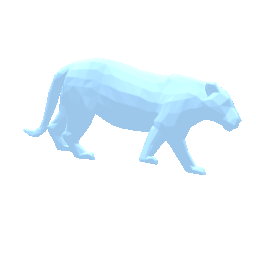}}~
 \subfloat{\includegraphics[width=.12\textwidth]{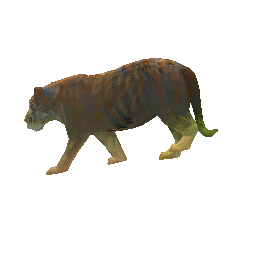}}~\\ 
 \vspace{-0.35cm}
 \subfloat{\includegraphics[width=.12\textwidth]{images/random_results/2/input_gt.png}}~
 \subfloat{\includegraphics[width=.12\textwidth]{images/random_results/2/mesh.png}}~
 \subfloat{\includegraphics[width=.12\textwidth]{images/random_results/2/mesh_vp.png}}~
 \subfloat{\includegraphics[width=.12\textwidth]{images/random_results/2/tex.png}}~
 \subfloat{\includegraphics[width=.12\textwidth]{images/random_results/8/input_gt.png}}~
 \subfloat{\includegraphics[width=.12\textwidth]{images/random_results/8/mesh.png}}~
 \subfloat{\includegraphics[width=.12\textwidth]{images/random_results/8/mesh_vp.png}}~
 \subfloat{\includegraphics[width=.12\textwidth]{images/random_results/8/tex.png}}~\\ 
 \vspace{-0.35cm}
 \subfloat{\includegraphics[width=.12\textwidth]{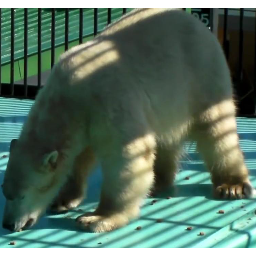}}~
 \subfloat{\includegraphics[width=.12\textwidth]{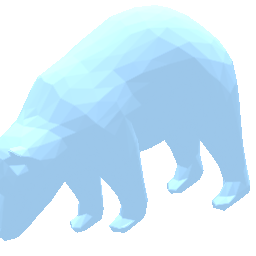}}~
 \subfloat{\includegraphics[width=.12\textwidth]{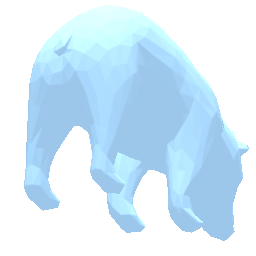}}~
 \subfloat{\includegraphics[width=.12\textwidth]{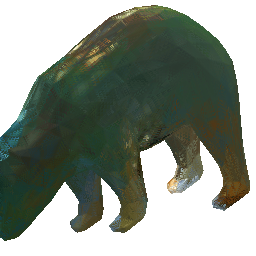}}~
 \subfloat{\includegraphics[width=.12\textwidth]{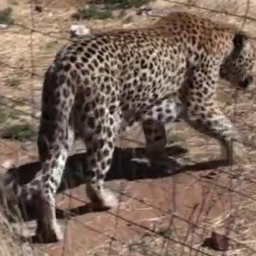}}~
 \subfloat{\includegraphics[width=.12\textwidth]{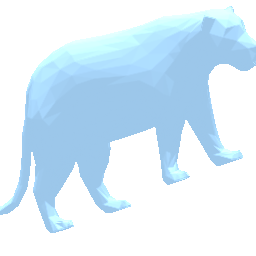}}~
 \subfloat{\includegraphics[width=.12\textwidth]{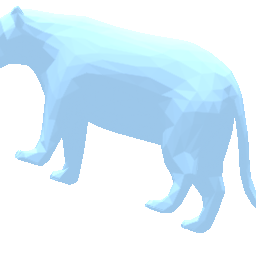}}~
 \subfloat{\includegraphics[width=.12\textwidth]{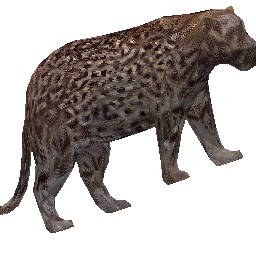}}~\\ 
 \vspace{-0.35cm}
 \subfloat{\includegraphics[width=.12\textwidth]{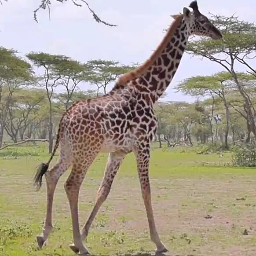}}~
 \subfloat{\includegraphics[width=.12\textwidth]{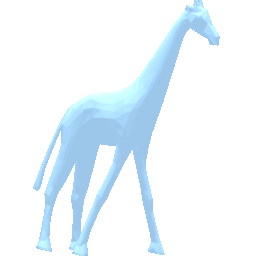}}~
 \subfloat{\includegraphics[width=.12\textwidth]{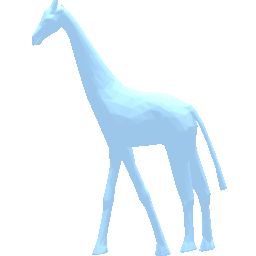}}~
 \subfloat{\includegraphics[width=.12\textwidth]{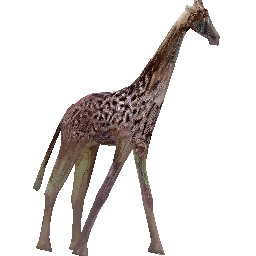}}~
 \subfloat{\includegraphics[width=.12\textwidth]{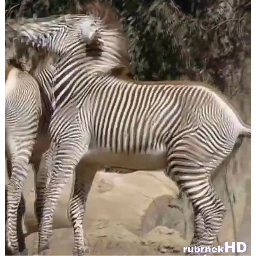}}~
 \subfloat{\includegraphics[width=.12\textwidth]{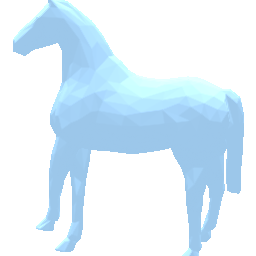}}~
 \subfloat{\includegraphics[width=.12\textwidth]{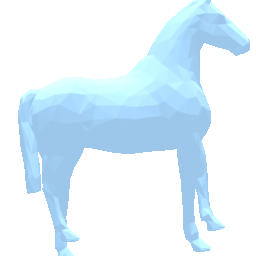}}~
 \subfloat{\includegraphics[width=.12\textwidth]{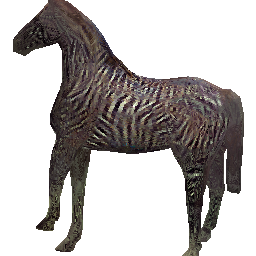}}~\\ 
 \vspace{-0.35cm}
 \subfloat{\includegraphics[width=.12\textwidth]{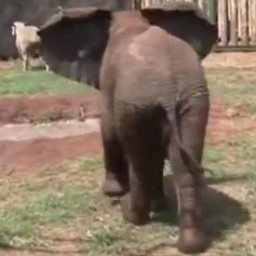}}~
 \subfloat{\includegraphics[width=.12\textwidth]{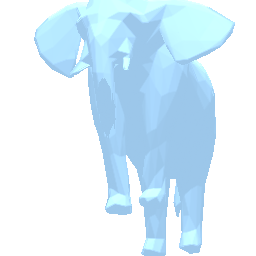}}~
 \subfloat{\includegraphics[width=.12\textwidth]{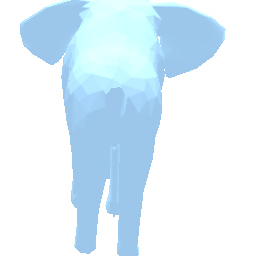}}~
 \subfloat{\includegraphics[width=.12\textwidth]{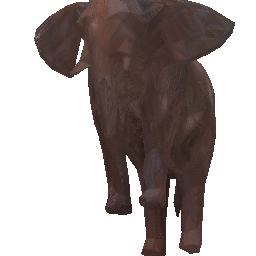}}~
 \subfloat{\includegraphics[width=.12\textwidth]{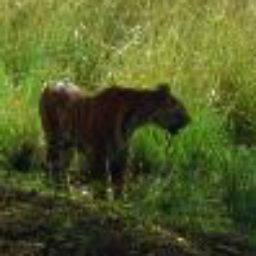}}~
 \subfloat{\includegraphics[width=.12\textwidth]{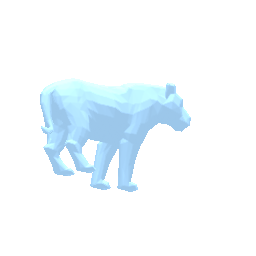}}~
 \subfloat{\includegraphics[width=.12\textwidth]{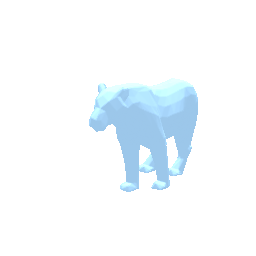}}~
 \subfloat{\includegraphics[width=.12\textwidth]{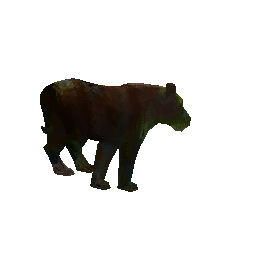}}~\\ 
 \vspace{-0.3cm}
 \caption{\textbf{3D reconstructions} We show the input image, the reconstructed shape from the predicted and a novel view and finally the predicted texture.\label{fig:3}}
 \end{centering}
 \end{figure*}

\begin{figure*}[t]
 \captionsetup[subfigure]{labelformat=empty, position=top}
 \begin{centering}
 \setcounter{subfigure}{0}

\subfloat{\includegraphics[width=.12\textwidth]{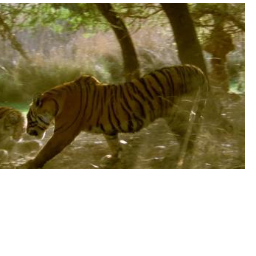}}~
 \subfloat{\includegraphics[width=.12\textwidth]{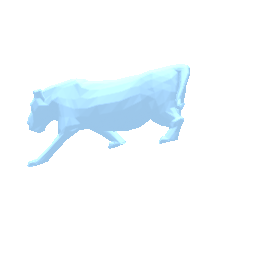}}~
 \subfloat{\includegraphics[width=.12\textwidth]{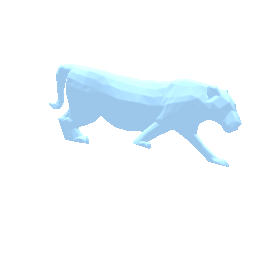}}~
 \subfloat{\includegraphics[width=.12\textwidth]{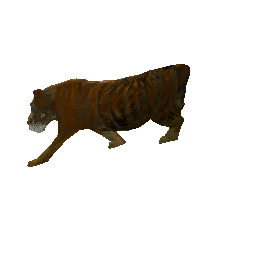}}~
 \subfloat{\includegraphics[width=.12\textwidth]{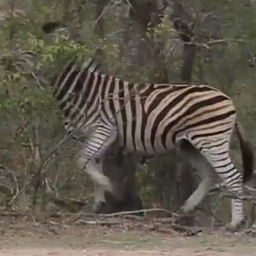}}~
 \subfloat{\includegraphics[width=.12\textwidth]{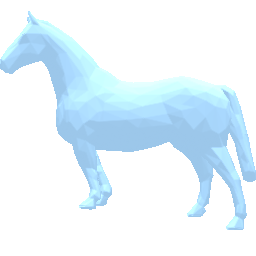}}~
 \subfloat{\includegraphics[width=.12\textwidth]{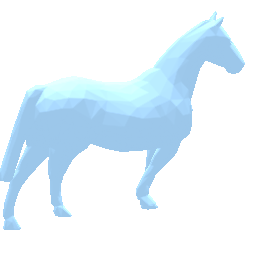}}~
 \subfloat{\includegraphics[width=.12\textwidth]{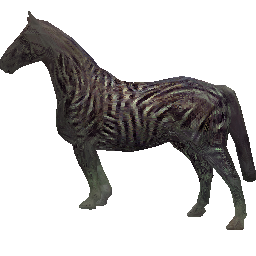}}~\\ 
 \vspace{-0.35cm}
 \subfloat{\includegraphics[width=.12\textwidth]{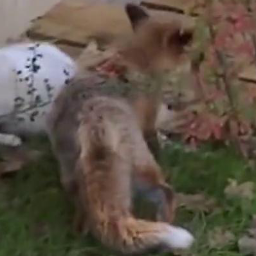}}~
 \subfloat{\includegraphics[width=.12\textwidth]{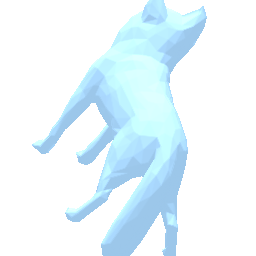}}~
 \subfloat{\includegraphics[width=.12\textwidth]{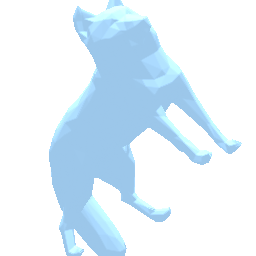}}~
 \subfloat{\includegraphics[width=.12\textwidth]{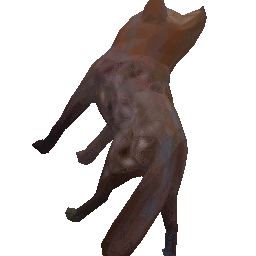}}~
 \subfloat{\includegraphics[width=.12\textwidth]{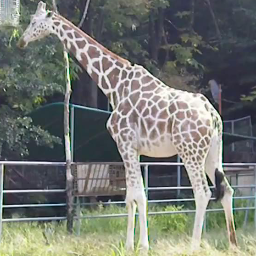}}~
 \subfloat{\includegraphics[width=.12\textwidth]{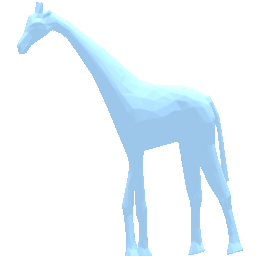}}~
 \subfloat{\includegraphics[width=.12\textwidth]{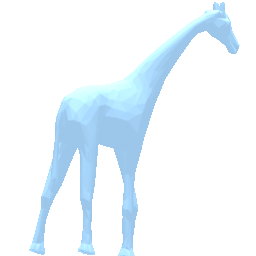}}~
 \subfloat{\includegraphics[width=.12\textwidth]{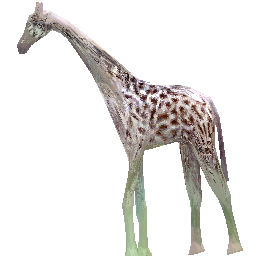}}~\\
 \vspace{-0.35cm}
 \subfloat{\includegraphics[width=.12\textwidth]{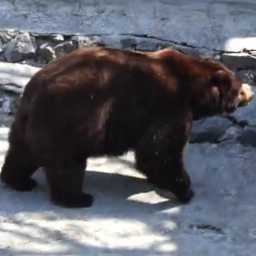}}~
 \subfloat{\includegraphics[width=.12\textwidth]{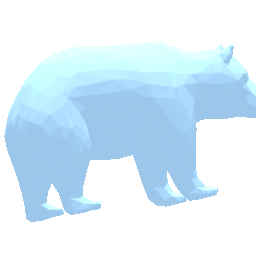}}~
 \subfloat{\includegraphics[width=.12\textwidth]{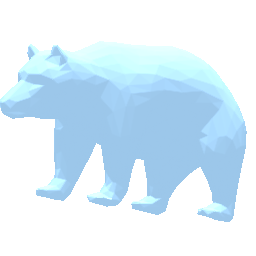}}~
 \subfloat{\includegraphics[width=.12\textwidth]{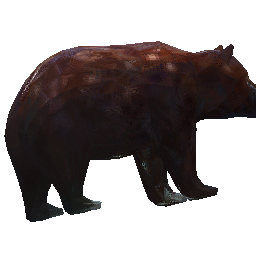}}~
 \subfloat{\includegraphics[width=.12\textwidth]{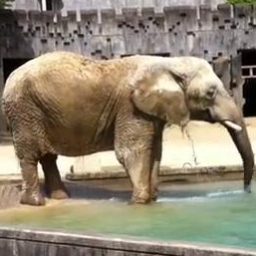}}~
 \subfloat{\includegraphics[width=.12\textwidth]{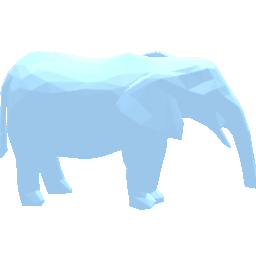}}~
 \subfloat{\includegraphics[width=.12\textwidth]{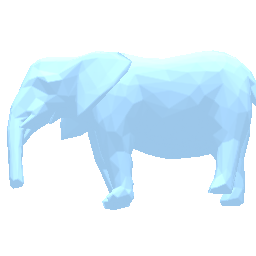}}~
 \subfloat{\includegraphics[width=.12\textwidth]{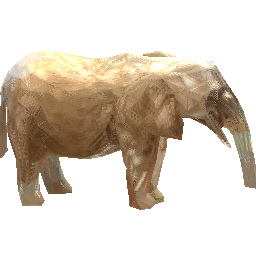}}~\\ 
 \vspace{-0.35cm}
 \subfloat{\includegraphics[width=.12\textwidth]{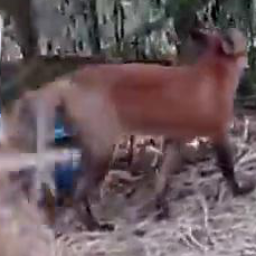}}~
 \subfloat{\includegraphics[width=.12\textwidth]{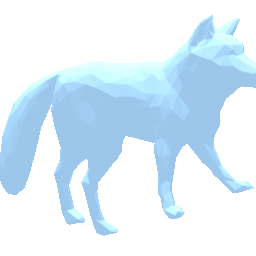}}~
 \subfloat{\includegraphics[width=.12\textwidth]{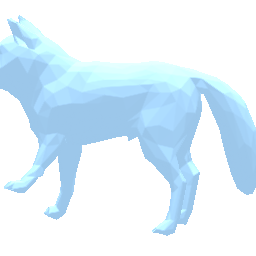}}~
 \subfloat{\includegraphics[width=.12\textwidth]{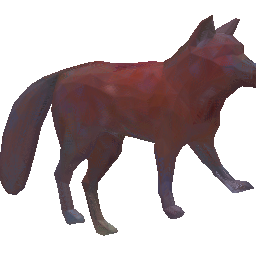}}~
 \subfloat{\includegraphics[width=.12\textwidth]{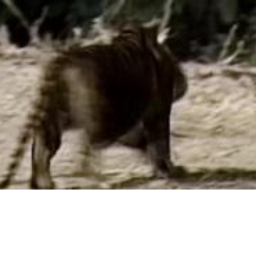}}~
 \subfloat{\includegraphics[width=.12\textwidth]{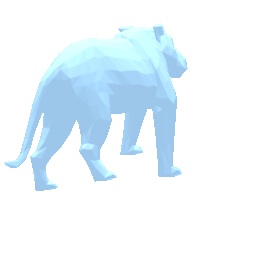}}~
 \subfloat{\includegraphics[width=.12\textwidth]{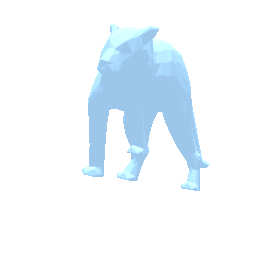}}~
 \subfloat{\includegraphics[width=.12\textwidth]{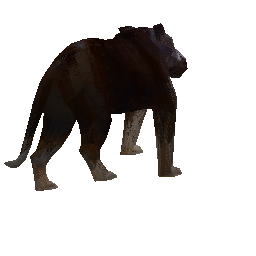}}~\\
 \vspace{-0.35cm}
 \subfloat{\includegraphics[width=.12\textwidth]{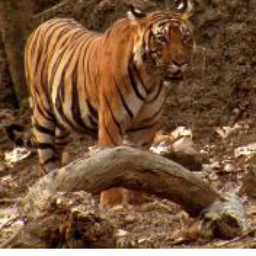}}~
 \subfloat{\includegraphics[width=.12\textwidth]{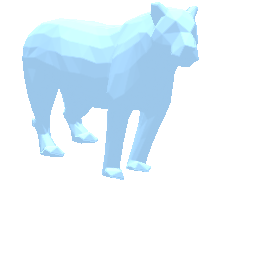}}~
 \subfloat{\includegraphics[width=.12\textwidth]{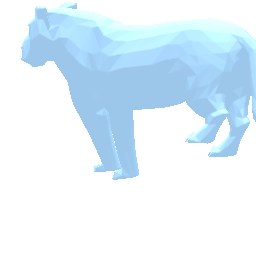}}~
 \subfloat{\includegraphics[width=.12\textwidth]{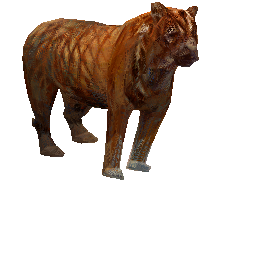}}~
 \subfloat{\includegraphics[width=.12\textwidth]{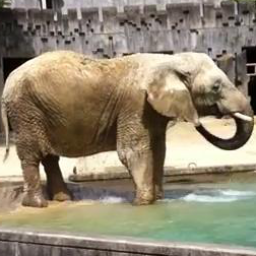}}~
 \subfloat{\includegraphics[width=.12\textwidth]{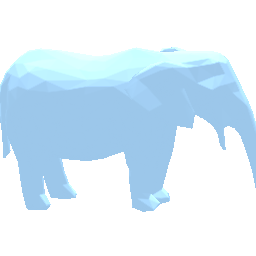}}~
 \subfloat{\includegraphics[width=.12\textwidth]{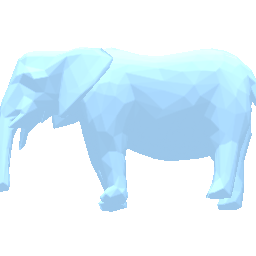}}~
 \subfloat{\includegraphics[width=.12\textwidth]{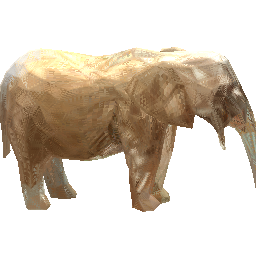}}~\\ 
 \vspace{-0.35cm}
 \subfloat{\includegraphics[width=.12\textwidth]{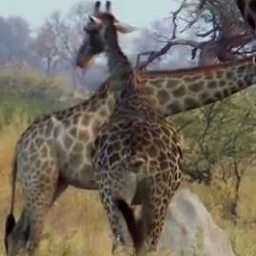}}~
 \subfloat{\includegraphics[width=.12\textwidth]{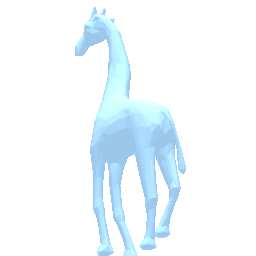}}~
 \subfloat{\includegraphics[width=.12\textwidth]{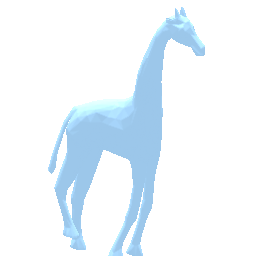}}~
 \subfloat{\includegraphics[width=.12\textwidth]{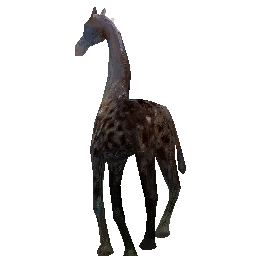}}~
 \subfloat{\includegraphics[width=.12\textwidth]{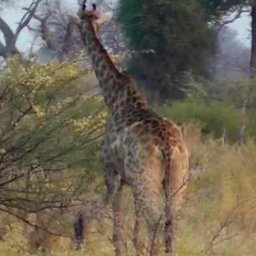}}~
 \subfloat{\includegraphics[width=.12\textwidth]{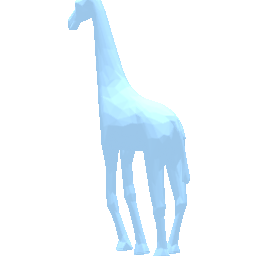}}~
 \subfloat{\includegraphics[width=.12\textwidth]{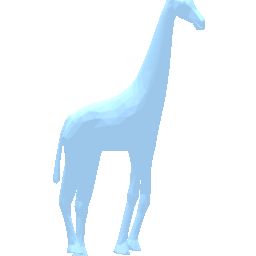}}~
 \subfloat{\includegraphics[width=.12\textwidth]{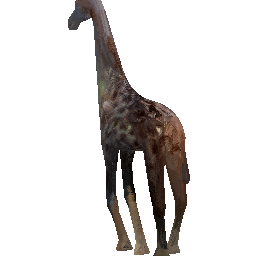}}~\\ 
 \vspace{-0.35cm}
 \subfloat{\includegraphics[width=.12\textwidth]{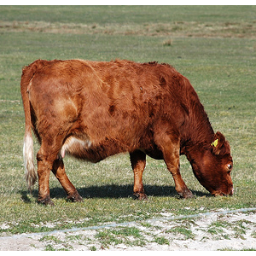}}~
 \subfloat{\includegraphics[width=.12\textwidth]{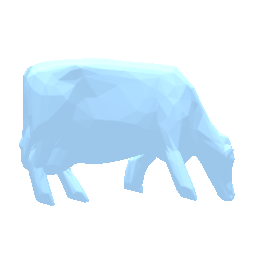}}~
 \subfloat{\includegraphics[width=.12\textwidth]{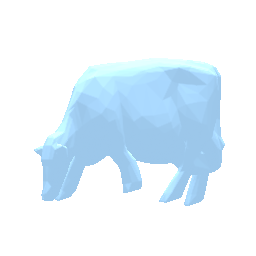}}~
 \subfloat{\includegraphics[width=.12\textwidth]{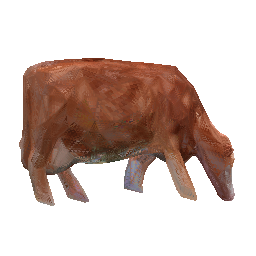}}~
 \subfloat{\includegraphics[width=.12\textwidth]{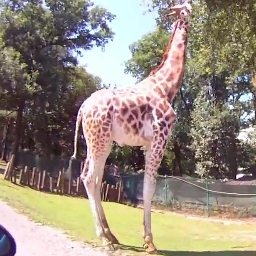}}~
 \subfloat{\includegraphics[width=.12\textwidth]{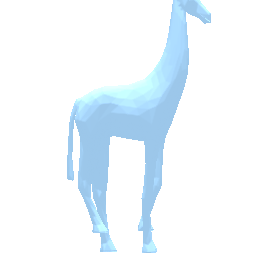}}~
 \subfloat{\includegraphics[width=.12\textwidth]{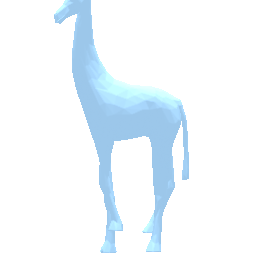}}~
 \subfloat{\includegraphics[width=.12\textwidth]{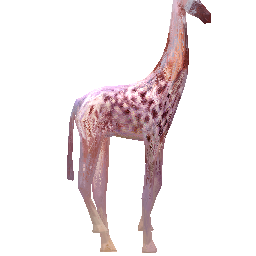}}~\\ 
 \vspace{-0.35cm}
 \subfloat{\includegraphics[width=.12\textwidth]{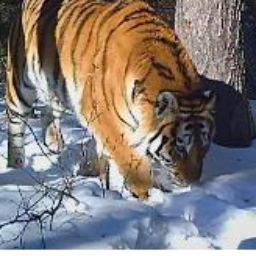}}~
 \subfloat{\includegraphics[width=.12\textwidth]{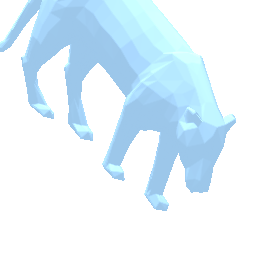}}~
 \subfloat{\includegraphics[width=.12\textwidth]{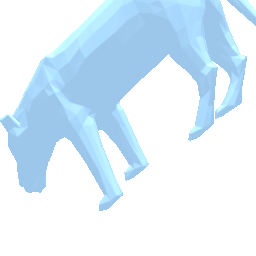}}~
 \subfloat{\includegraphics[width=.12\textwidth]{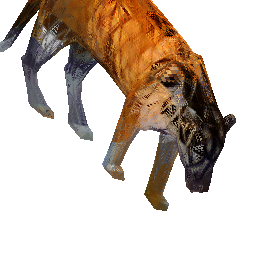}}~
 \subfloat{\includegraphics[width=.12\textwidth]{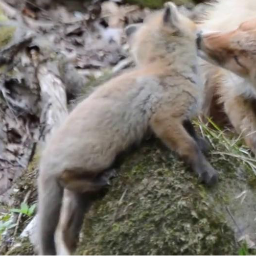}}~
 \subfloat{\includegraphics[width=.12\textwidth]{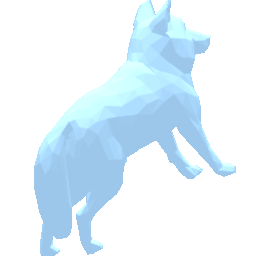}}~
 \subfloat{\includegraphics[width=.12\textwidth]{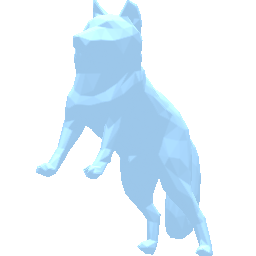}}~
 \subfloat{\includegraphics[width=.12\textwidth]{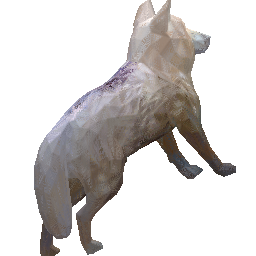}}~\\ 
 \vspace{-0.35cm}
 \subfloat{\includegraphics[width=.12\textwidth]{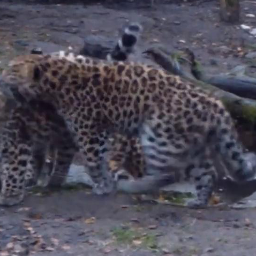}}~
 \subfloat{\includegraphics[width=.12\textwidth]{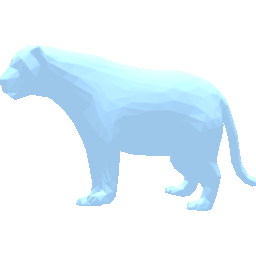}}~
 \subfloat{\includegraphics[width=.12\textwidth]{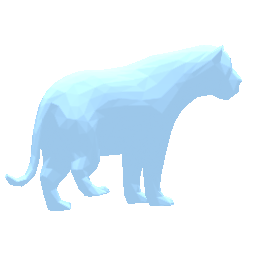}}~
 \subfloat{\includegraphics[width=.12\textwidth]{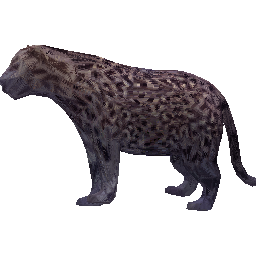}}~
 \subfloat{\includegraphics[width=.12\textwidth]{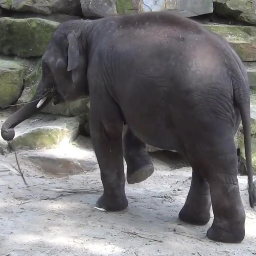}}~
 \subfloat{\includegraphics[width=.12\textwidth]{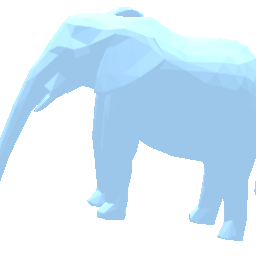}}~
 \subfloat{\includegraphics[width=.12\textwidth]{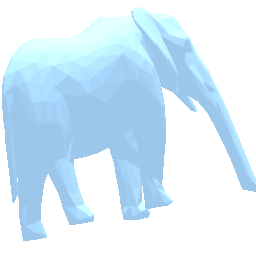}}~
 \subfloat{\includegraphics[width=.12\textwidth]{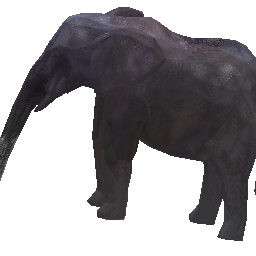}}~\\ 
 \vspace{-0.35cm}
 \subfloat{\includegraphics[width=.12\textwidth]{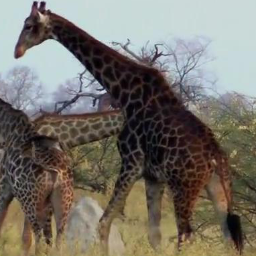}}~
 \subfloat{\includegraphics[width=.12\textwidth]{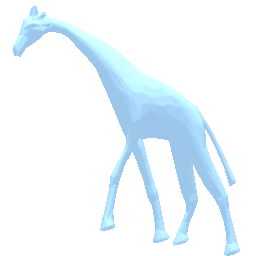}}~
 \subfloat{\includegraphics[width=.12\textwidth]{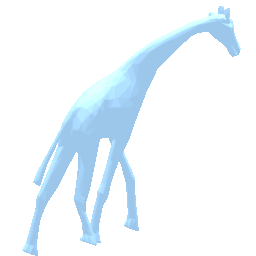}}~
 \subfloat{\includegraphics[width=.12\textwidth]{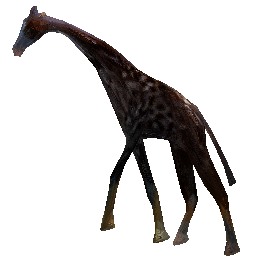}}~
 \subfloat{\includegraphics[width=.12\textwidth]{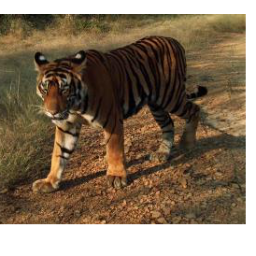}}~
 \subfloat{\includegraphics[width=.12\textwidth]{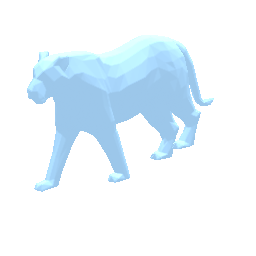}}~
 \subfloat{\includegraphics[width=.12\textwidth]{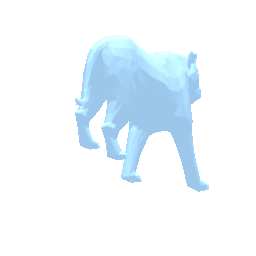}}~
 \subfloat{\includegraphics[width=.12\textwidth]{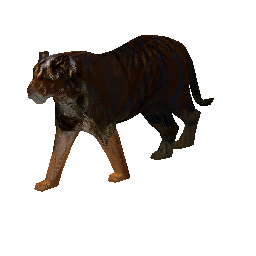}}~\\ 
 \vspace{-0.3cm}
 \caption{\textbf{3D reconstructions} We show the input image, the reconstructed shape from the predicted and a novel view and finally the predicted texture.\label{fig:4}}
 \end{centering}
 \end{figure*}

\begin{figure*}[t]
 \captionsetup[subfigure]{labelformat=empty, position=top}
 \begin{centering}
 \setcounter{subfigure}{0}

\subfloat{\includegraphics[width=.12\textwidth]{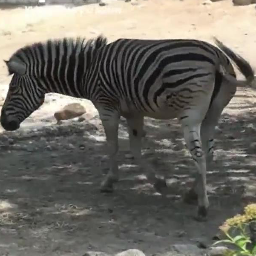}}~
 \subfloat{\includegraphics[width=.12\textwidth]{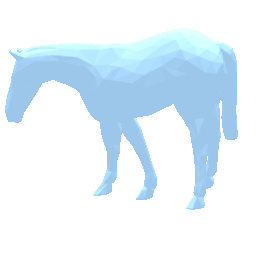}}~
 \subfloat{\includegraphics[width=.12\textwidth]{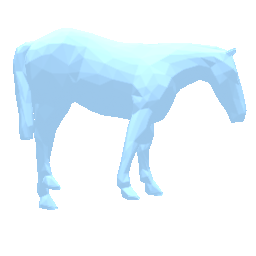}}~
 \subfloat{\includegraphics[width=.12\textwidth]{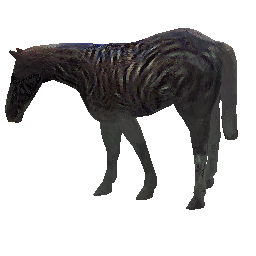}}~
 \subfloat{\includegraphics[width=.12\textwidth]{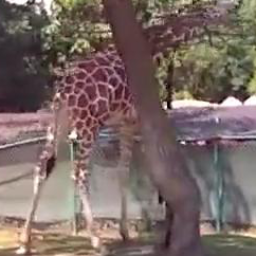}}~
 \subfloat{\includegraphics[width=.12\textwidth]{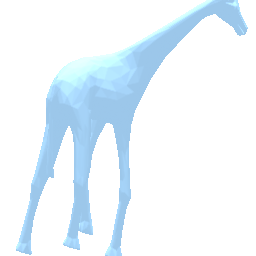}}~
 \subfloat{\includegraphics[width=.12\textwidth]{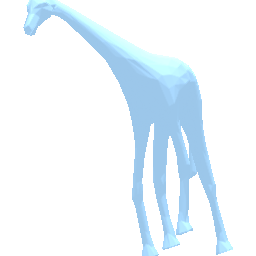}}~
 \subfloat{\includegraphics[width=.12\textwidth]{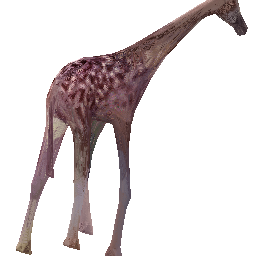}}~\\ 
 \vspace{-0.35cm}
 \subfloat{\includegraphics[width=.12\textwidth]{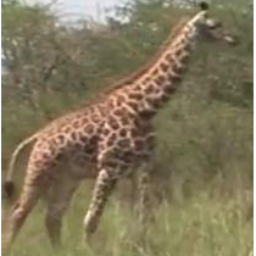}}~
 \subfloat{\includegraphics[width=.12\textwidth]{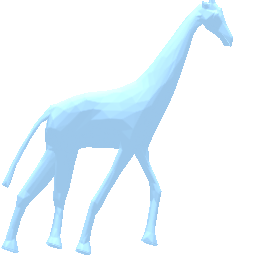}}~
 \subfloat{\includegraphics[width=.12\textwidth]{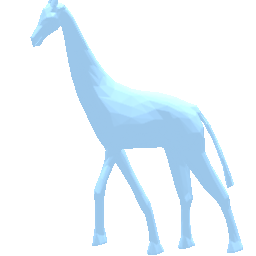}}~
 \subfloat{\includegraphics[width=.12\textwidth]{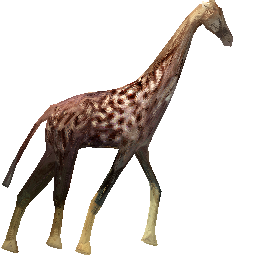}}~
 \subfloat{\includegraphics[width=.12\textwidth]{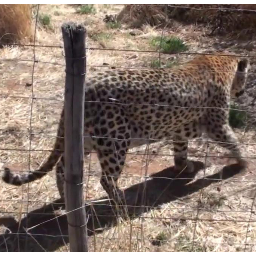}}~
 \subfloat{\includegraphics[width=.12\textwidth]{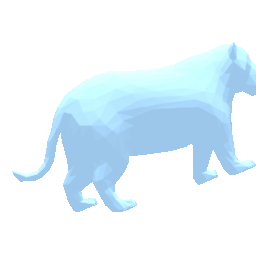}}~
 \subfloat{\includegraphics[width=.12\textwidth]{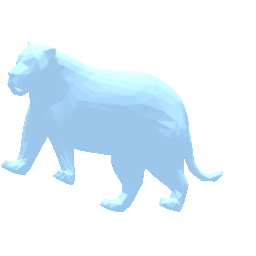}}~
 \subfloat{\includegraphics[width=.12\textwidth]{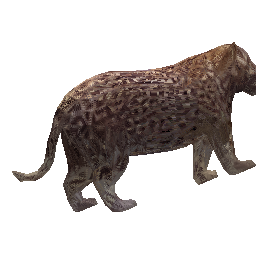}}~\\
 \vspace{-0.35cm}
 \subfloat{\includegraphics[width=.12\textwidth]{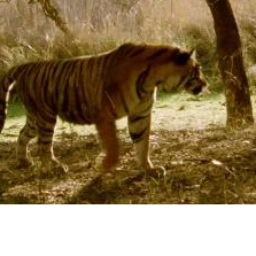}}~
 \subfloat{\includegraphics[width=.12\textwidth]{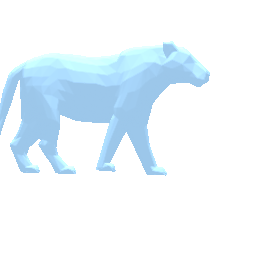}}~
 \subfloat{\includegraphics[width=.12\textwidth]{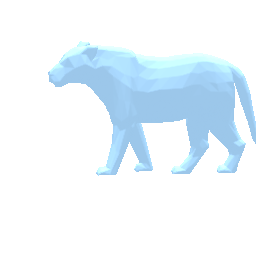}}~
 \subfloat{\includegraphics[width=.12\textwidth]{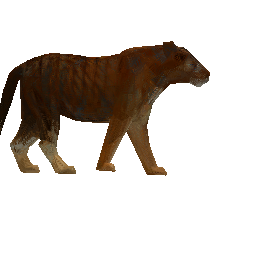}}~
 \subfloat{\includegraphics[width=.12\textwidth]{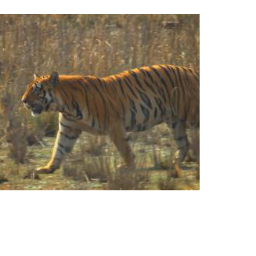}}~
 \subfloat{\includegraphics[width=.12\textwidth]{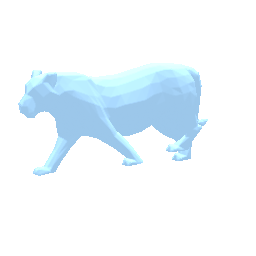}}~
 \subfloat{\includegraphics[width=.12\textwidth]{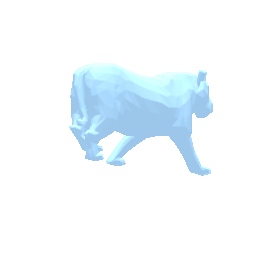}}~
 \subfloat{\includegraphics[width=.12\textwidth]{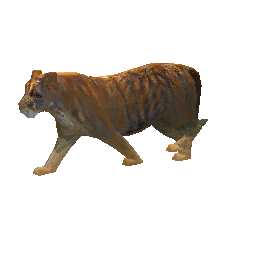}}~\\ 
 \vspace{-0.35cm}
 \subfloat{\includegraphics[width=.12\textwidth]{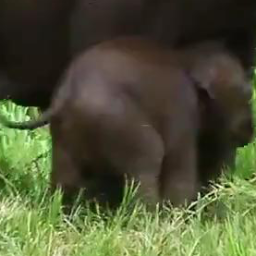}}~
 \subfloat{\includegraphics[width=.12\textwidth]{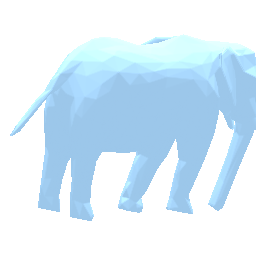}}~
 \subfloat{\includegraphics[width=.12\textwidth]{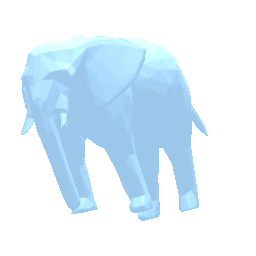}}~
 \subfloat{\includegraphics[width=.12\textwidth]{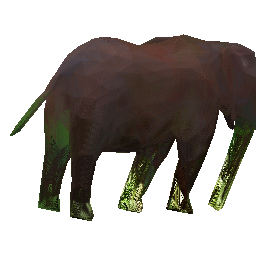}}~
 \subfloat{\includegraphics[width=.12\textwidth]{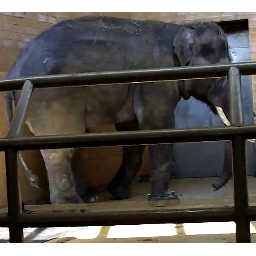}}~
 \subfloat{\includegraphics[width=.12\textwidth]{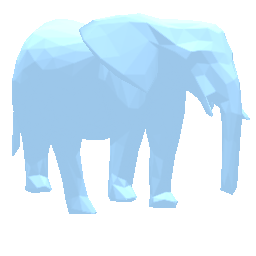}}~
 \subfloat{\includegraphics[width=.12\textwidth]{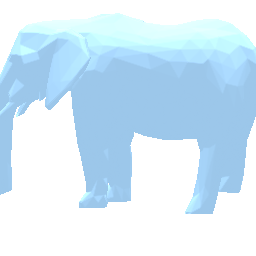}}~
 \subfloat{\includegraphics[width=.12\textwidth]{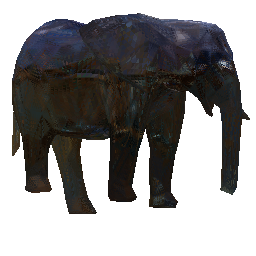}}~\\
 \vspace{-0.35cm}
 \subfloat{\includegraphics[width=.12\textwidth]{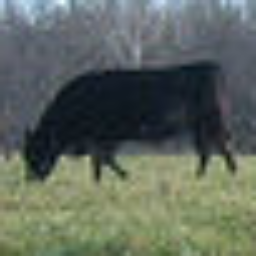}}~
 \subfloat{\includegraphics[width=.12\textwidth]{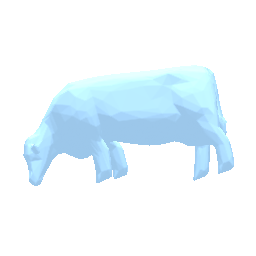}}~
 \subfloat{\includegraphics[width=.12\textwidth]{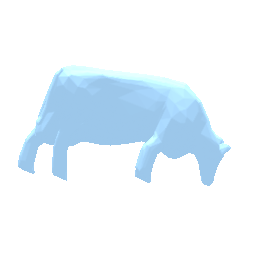}}~
 \subfloat{\includegraphics[width=.12\textwidth]{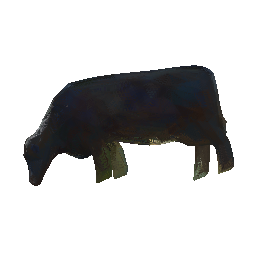}}~
 \subfloat{\includegraphics[width=.12\textwidth]{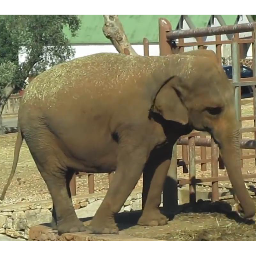}}~
 \subfloat{\includegraphics[width=.12\textwidth]{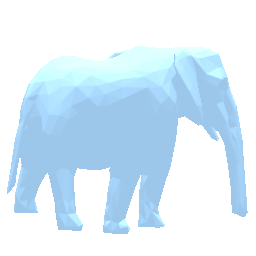}}~
 \subfloat{\includegraphics[width=.12\textwidth]{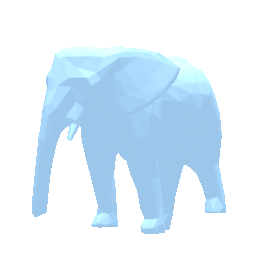}}~
 \subfloat{\includegraphics[width=.12\textwidth]{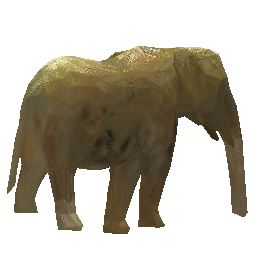}}~\\ 
 \vspace{-0.35cm}
 \subfloat{\includegraphics[width=.12\textwidth]{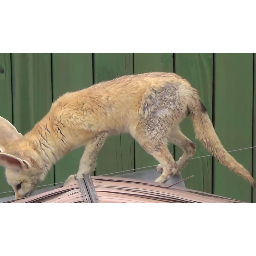}}~
 \subfloat{\includegraphics[width=.12\textwidth]{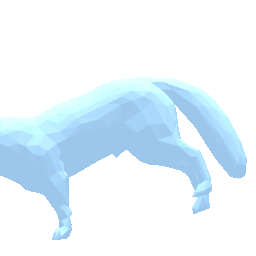}}~
 \subfloat{\includegraphics[width=.12\textwidth]{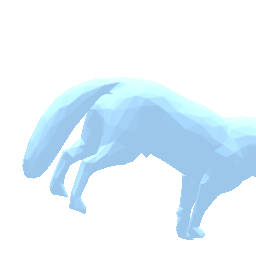}}~
 \subfloat{\includegraphics[width=.12\textwidth]{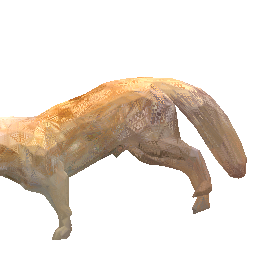}}~
 \subfloat{\includegraphics[width=.12\textwidth]{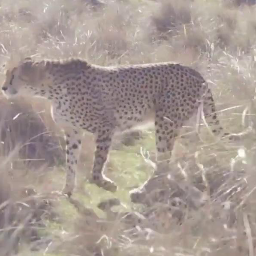}}~
 \subfloat{\includegraphics[width=.12\textwidth]{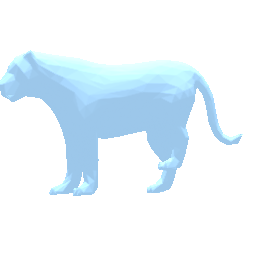}}~
 \subfloat{\includegraphics[width=.12\textwidth]{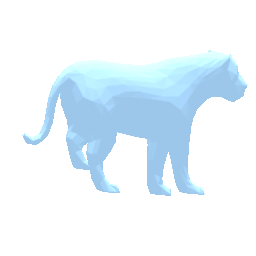}}~
 \subfloat{\includegraphics[width=.12\textwidth]{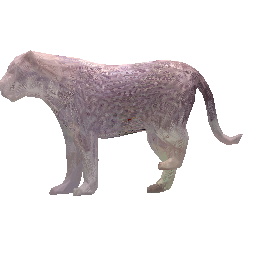}}~\\ 
 \vspace{-0.35cm}
 \subfloat{\includegraphics[width=.12\textwidth]{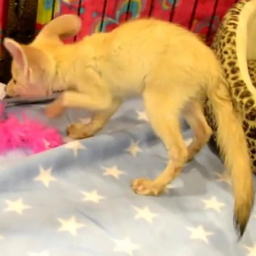}}~
 \subfloat{\includegraphics[width=.12\textwidth]{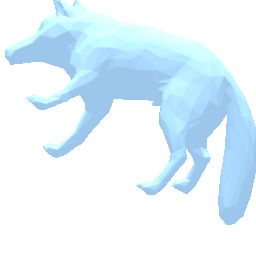}}~
 \subfloat{\includegraphics[width=.12\textwidth]{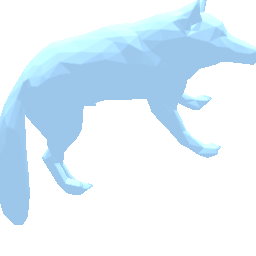}}~
 \subfloat{\includegraphics[width=.12\textwidth]{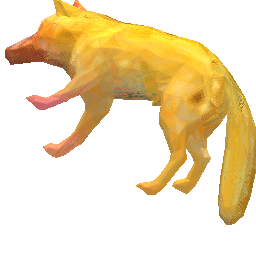}}~
 \subfloat{\includegraphics[width=.12\textwidth]{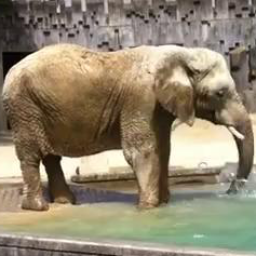}}~
 \subfloat{\includegraphics[width=.12\textwidth]{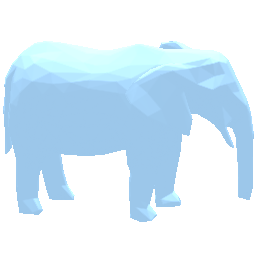}}~
 \subfloat{\includegraphics[width=.12\textwidth]{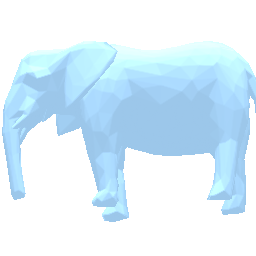}}~
 \subfloat{\includegraphics[width=.12\textwidth]{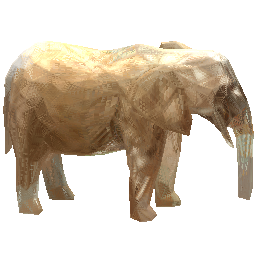}}~\\ 
 \vspace{-0.35cm}
 \subfloat{\includegraphics[width=.12\textwidth]{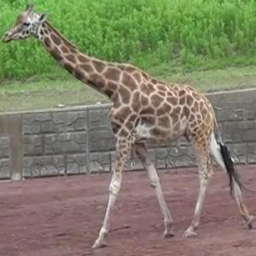}}~
 \subfloat{\includegraphics[width=.12\textwidth]{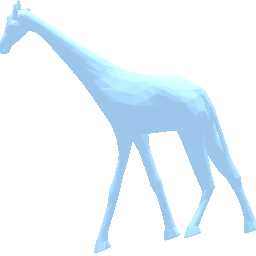}}~
 \subfloat{\includegraphics[width=.12\textwidth]{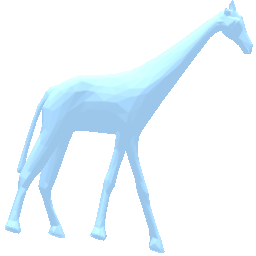}}~
 \subfloat{\includegraphics[width=.12\textwidth]{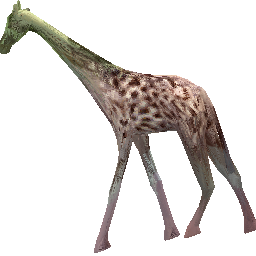}}~
 \subfloat{\includegraphics[width=.12\textwidth]{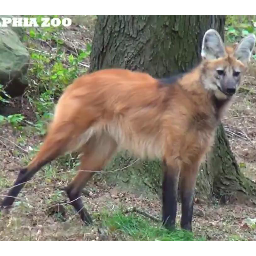}}~
 \subfloat{\includegraphics[width=.12\textwidth]{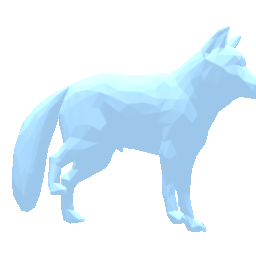}}~
 \subfloat{\includegraphics[width=.12\textwidth]{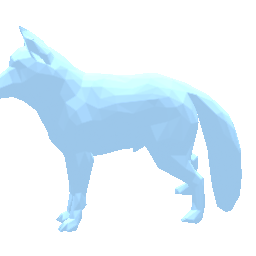}}~
 \subfloat{\includegraphics[width=.12\textwidth]{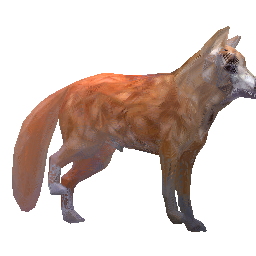}}~\\ 
 \vspace{-0.35cm}
 \subfloat{\includegraphics[width=.12\textwidth]{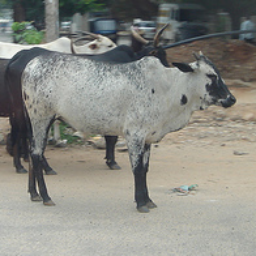}}~
 \subfloat{\includegraphics[width=.12\textwidth]{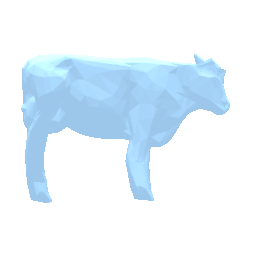}}~
 \subfloat{\includegraphics[width=.12\textwidth]{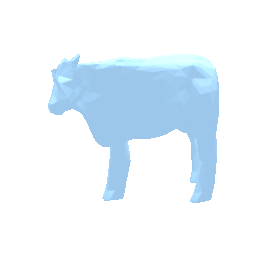}}~
 \subfloat{\includegraphics[width=.12\textwidth]{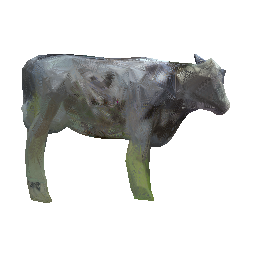}}~
 \subfloat{\includegraphics[width=.12\textwidth]{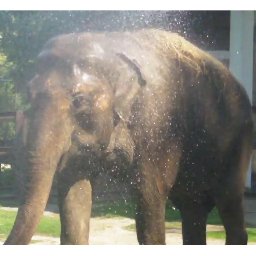}}~
 \subfloat{\includegraphics[width=.12\textwidth]{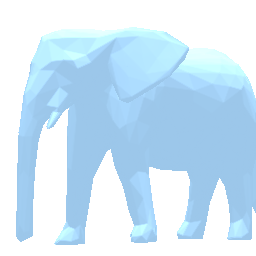}}~
 \subfloat{\includegraphics[width=.12\textwidth]{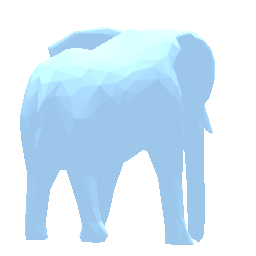}}~
 \subfloat{\includegraphics[width=.12\textwidth]{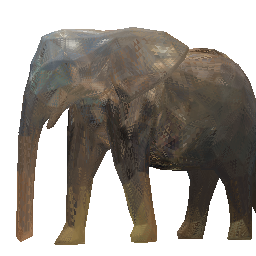}}~\\ 
 \vspace{-0.35cm}
 \subfloat{\includegraphics[width=.12\textwidth]{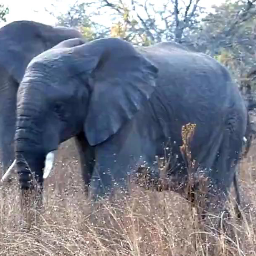}}~
 \subfloat{\includegraphics[width=.12\textwidth]{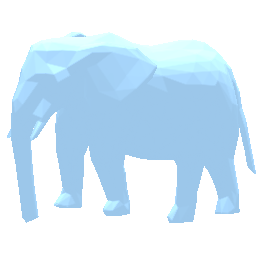}}~
 \subfloat{\includegraphics[width=.12\textwidth]{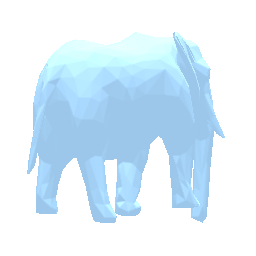}}~
 \subfloat{\includegraphics[width=.12\textwidth]{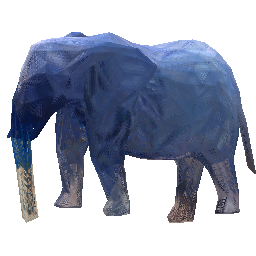}}~
 \subfloat{\includegraphics[width=.12\textwidth]{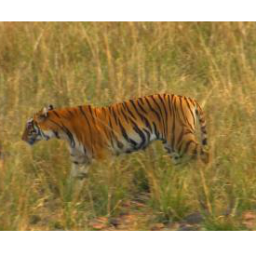}}~
 \subfloat{\includegraphics[width=.12\textwidth]{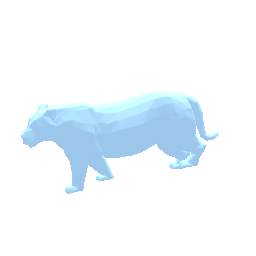}}~
 \subfloat{\includegraphics[width=.12\textwidth]{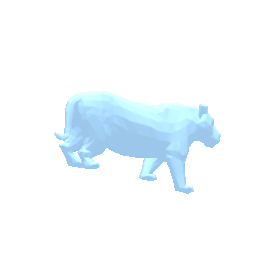}}~
 \subfloat{\includegraphics[width=.12\textwidth]{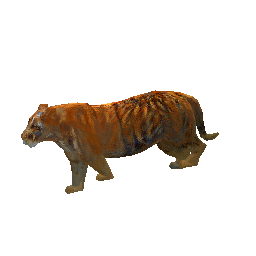}}~\\ 
 \vspace{-0.3cm}
 \caption{\textbf{3D reconstructions} We show the input image, the reconstructed shape from the predicted and a novel view and finally the predicted texture.\label{fig:5}}
 \end{centering}
 \end{figure*}

\begin{figure*}[t]
 \captionsetup[subfigure]{labelformat=empty, position=top}
 \begin{centering}
 \setcounter{subfigure}{0}

\subfloat{\includegraphics[width=.12\textwidth]{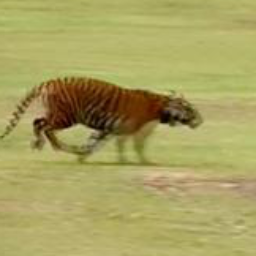}}~
 \subfloat{\includegraphics[width=.12\textwidth]{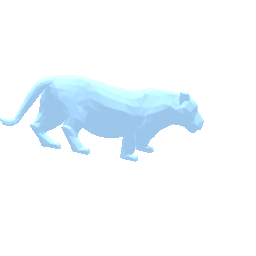}}~
 \subfloat{\includegraphics[width=.12\textwidth]{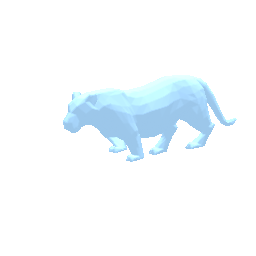}}~
 \subfloat{\includegraphics[width=.12\textwidth]{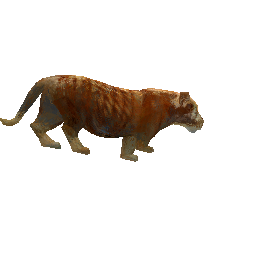}}~
 \subfloat{\includegraphics[width=.12\textwidth]{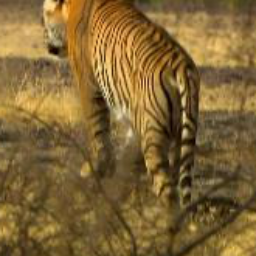}}~
 \subfloat{\includegraphics[width=.12\textwidth]{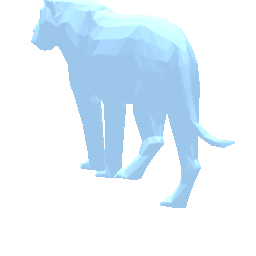}}~
 \subfloat{\includegraphics[width=.12\textwidth]{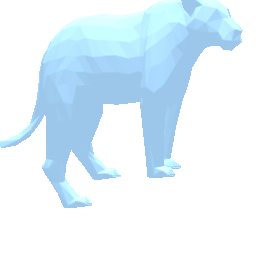}}~
 \subfloat{\includegraphics[width=.12\textwidth]{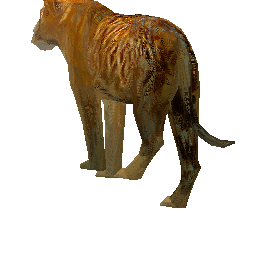}}~\\ 
 \vspace{-0.35cm}
 \subfloat{\includegraphics[width=.12\textwidth]{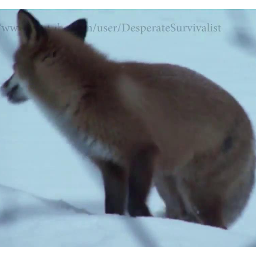}}~
 \subfloat{\includegraphics[width=.12\textwidth]{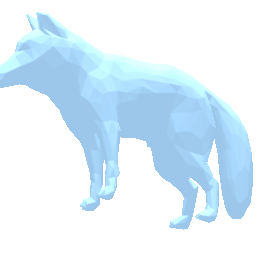}}~
 \subfloat{\includegraphics[width=.12\textwidth]{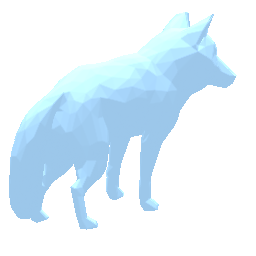}}~
 \subfloat{\includegraphics[width=.12\textwidth]{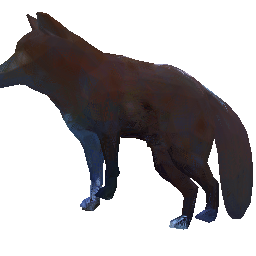}}~
 \subfloat{\includegraphics[width=.12\textwidth]{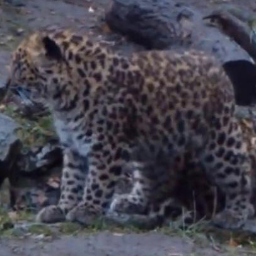}}~
 \subfloat{\includegraphics[width=.12\textwidth]{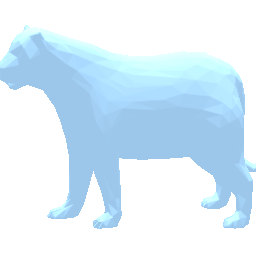}}~
 \subfloat{\includegraphics[width=.12\textwidth]{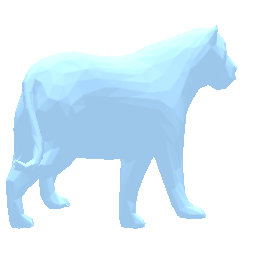}}~
 \subfloat{\includegraphics[width=.12\textwidth]{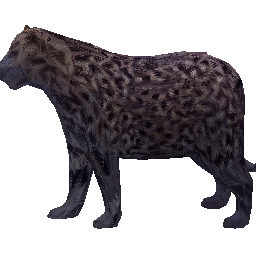}}~\\
 \vspace{-0.35cm}
 \subfloat{\includegraphics[width=.12\textwidth]{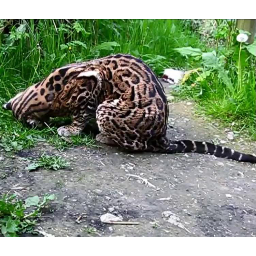}}~
 \subfloat{\includegraphics[width=.12\textwidth]{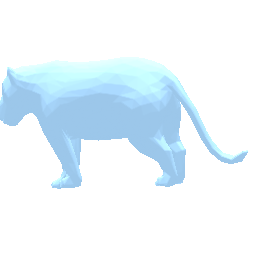}}~
 \subfloat{\includegraphics[width=.12\textwidth]{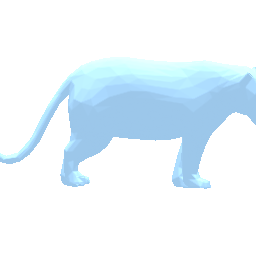}}~
 \subfloat{\includegraphics[width=.12\textwidth]{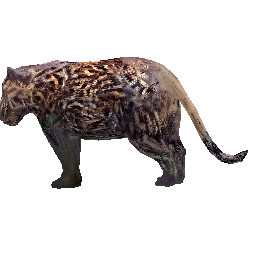}}~
 \subfloat{\includegraphics[width=.12\textwidth]{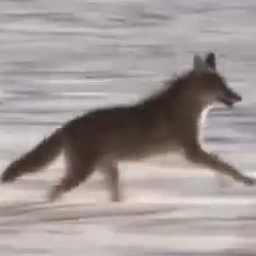}}~
 \subfloat{\includegraphics[width=.12\textwidth]{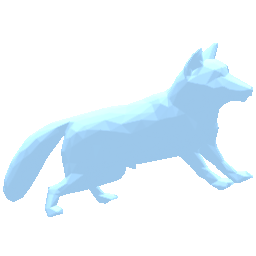}}~
 \subfloat{\includegraphics[width=.12\textwidth]{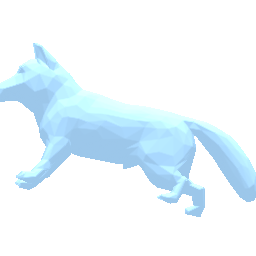}}~
 \subfloat{\includegraphics[width=.12\textwidth]{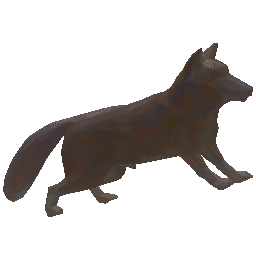}}~\\ 
 \vspace{-0.35cm}
 \subfloat{\includegraphics[width=.12\textwidth]{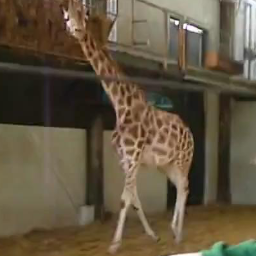}}~
 \subfloat{\includegraphics[width=.12\textwidth]{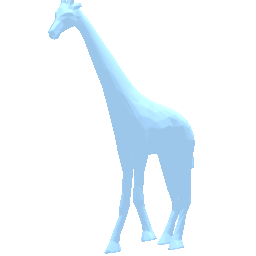}}~
 \subfloat{\includegraphics[width=.12\textwidth]{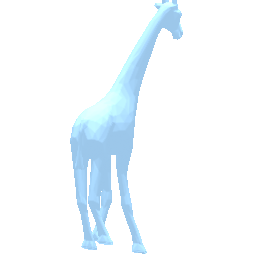}}~
 \subfloat{\includegraphics[width=.12\textwidth]{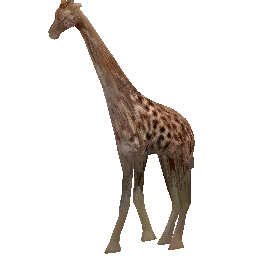}}~
 \subfloat{\includegraphics[width=.12\textwidth]{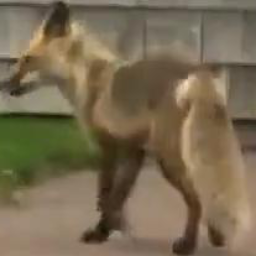}}~
 \subfloat{\includegraphics[width=.12\textwidth]{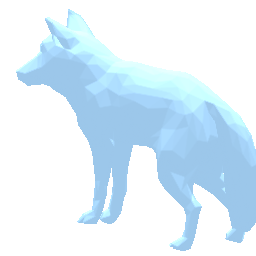}}~
 \subfloat{\includegraphics[width=.12\textwidth]{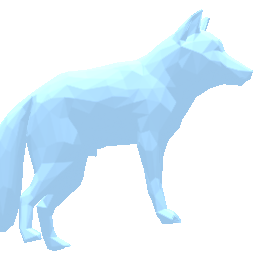}}~
 \subfloat{\includegraphics[width=.12\textwidth]{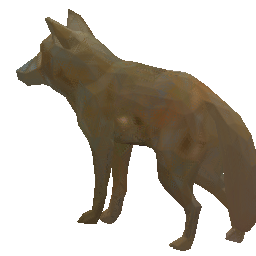}}~\\
 \vspace{-0.35cm}
 \subfloat{\includegraphics[width=.12\textwidth]{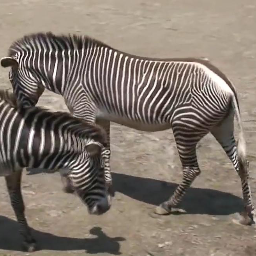}}~
 \subfloat{\includegraphics[width=.12\textwidth]{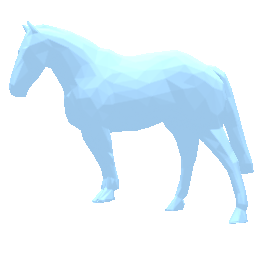}}~
 \subfloat{\includegraphics[width=.12\textwidth]{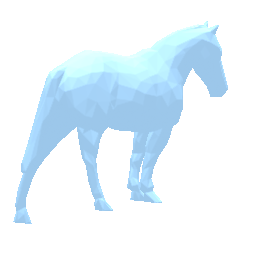}}~
 \subfloat{\includegraphics[width=.12\textwidth]{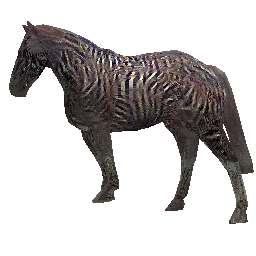}}~
 \subfloat{\includegraphics[width=.12\textwidth]{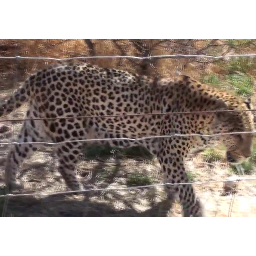}}~
 \subfloat{\includegraphics[width=.12\textwidth]{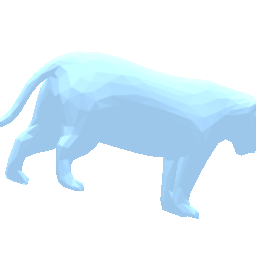}}~
 \subfloat{\includegraphics[width=.12\textwidth]{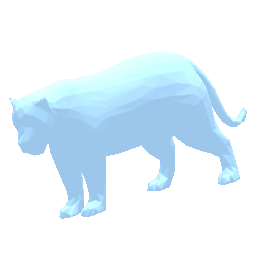}}~
 \subfloat{\includegraphics[width=.12\textwidth]{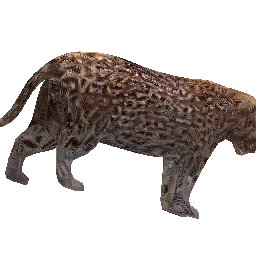}}~\\ 
 \vspace{-0.35cm}
 \subfloat{\includegraphics[width=.12\textwidth]{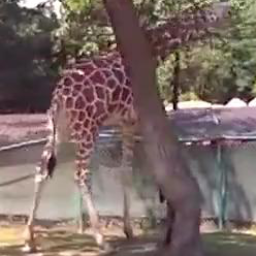}}~
 \subfloat{\includegraphics[width=.12\textwidth]{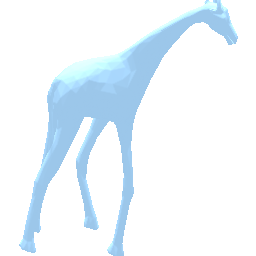}}~
 \subfloat{\includegraphics[width=.12\textwidth]{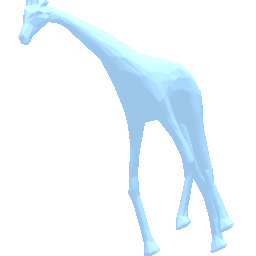}}~
 \subfloat{\includegraphics[width=.12\textwidth]{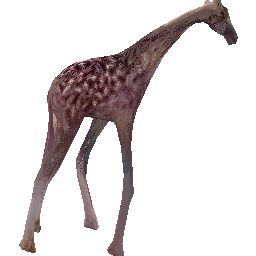}}~
 \subfloat{\includegraphics[width=.12\textwidth]{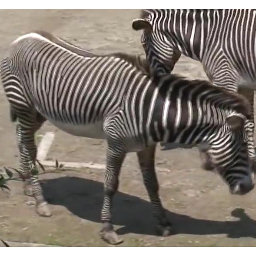}}~
 \subfloat{\includegraphics[width=.12\textwidth]{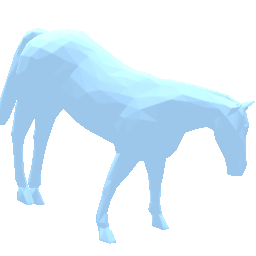}}~
 \subfloat{\includegraphics[width=.12\textwidth]{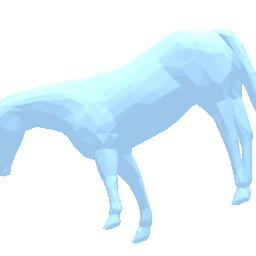}}~
 \subfloat{\includegraphics[width=.12\textwidth]{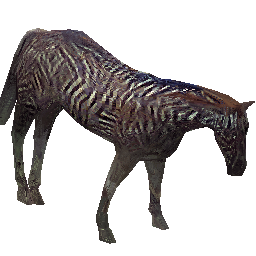}}~\\ 
 \vspace{-0.35cm}
 \subfloat{\includegraphics[width=.12\textwidth]{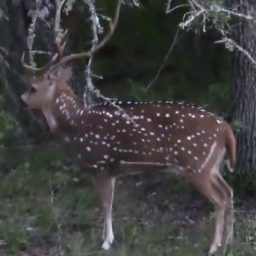}}~
 \subfloat{\includegraphics[width=.12\textwidth]{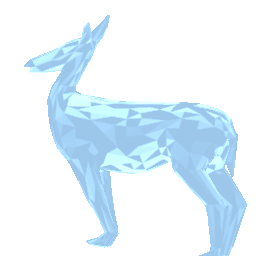}}~
 \subfloat{\includegraphics[width=.12\textwidth]{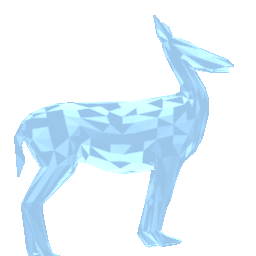}}~
 \subfloat{\includegraphics[width=.12\textwidth]{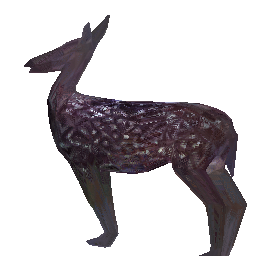}}~
 \subfloat{\includegraphics[width=.12\textwidth]{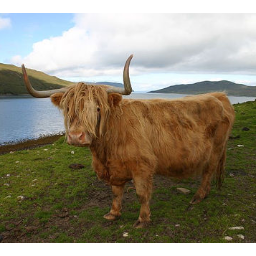}}~
 \subfloat{\includegraphics[width=.12\textwidth]{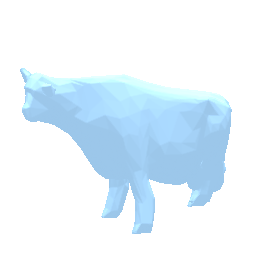}}~
 \subfloat{\includegraphics[width=.12\textwidth]{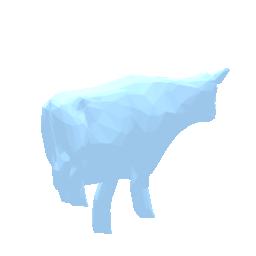}}~
 \subfloat{\includegraphics[width=.12\textwidth]{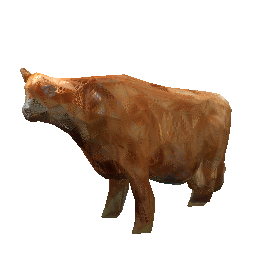}}~\\ 
 \vspace{-0.35cm}
 \subfloat{\includegraphics[width=.12\textwidth]{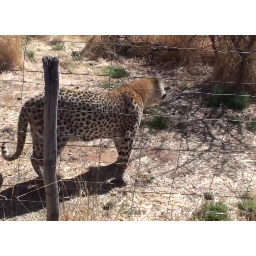}}~
 \subfloat{\includegraphics[width=.12\textwidth]{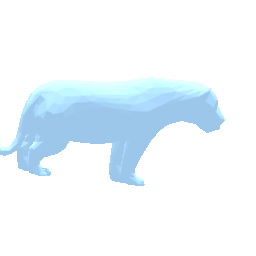}}~
 \subfloat{\includegraphics[width=.12\textwidth]{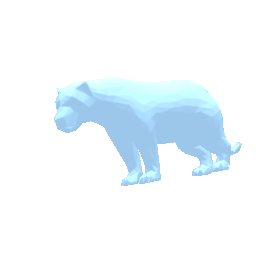}}~
 \subfloat{\includegraphics[width=.12\textwidth]{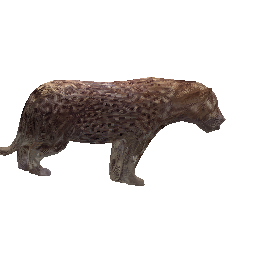}}~
 \subfloat{\includegraphics[width=.12\textwidth]{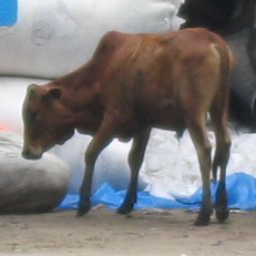}}~
 \subfloat{\includegraphics[width=.12\textwidth]{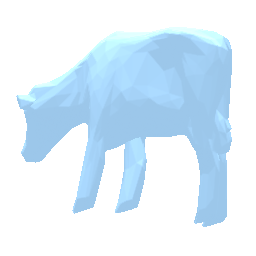}}~
 \subfloat{\includegraphics[width=.12\textwidth]{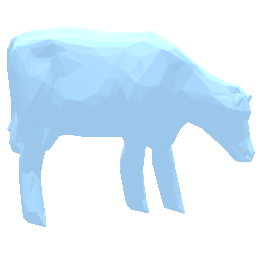}}~
 \subfloat{\includegraphics[width=.12\textwidth]{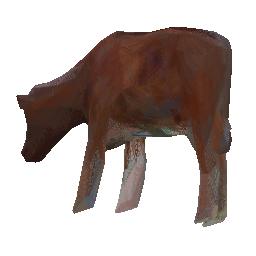}}~\\ 
 \vspace{-0.35cm}
 \subfloat{\includegraphics[width=.12\textwidth]{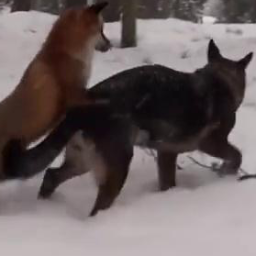}}~
 \subfloat{\includegraphics[width=.12\textwidth]{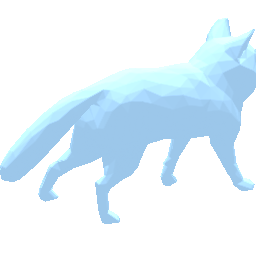}}~
 \subfloat{\includegraphics[width=.12\textwidth]{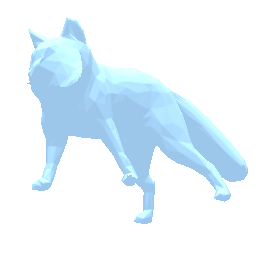}}~
 \subfloat{\includegraphics[width=.12\textwidth]{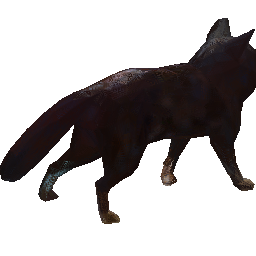}}~
 \subfloat{\includegraphics[width=.12\textwidth]{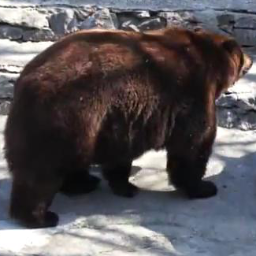}}~
 \subfloat{\includegraphics[width=.12\textwidth]{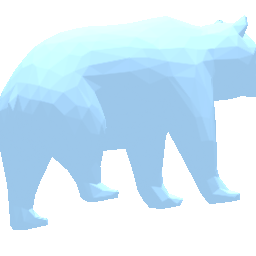}}~
 \subfloat{\includegraphics[width=.12\textwidth]{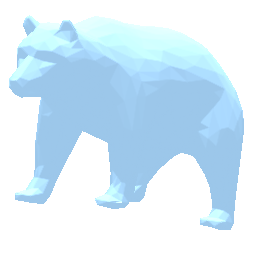}}~
 \subfloat{\includegraphics[width=.12\textwidth]{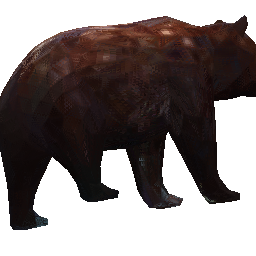}}~\\ 
 \vspace{-0.35cm}
 \subfloat{\includegraphics[width=.12\textwidth]{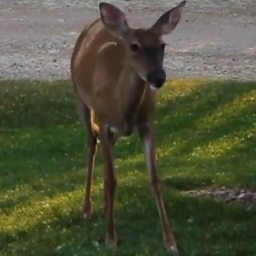}}~
 \subfloat{\includegraphics[width=.12\textwidth]{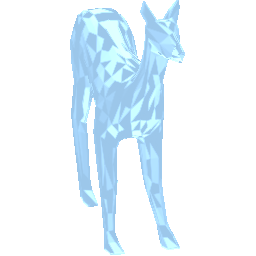}}~
 \subfloat{\includegraphics[width=.12\textwidth]{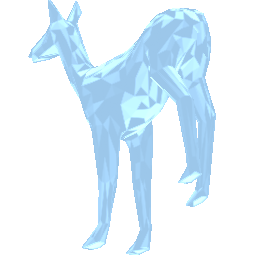}}~
 \subfloat{\includegraphics[width=.12\textwidth]{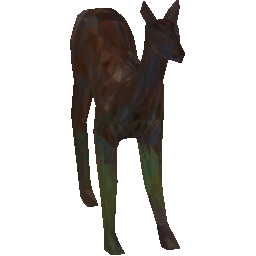}}~
 \subfloat{\includegraphics[width=.12\textwidth]{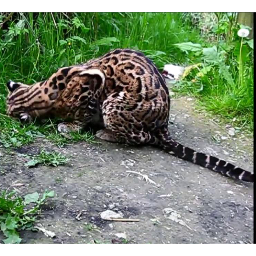}}~
 \subfloat{\includegraphics[width=.12\textwidth]{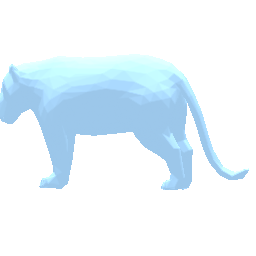}}~
 \subfloat{\includegraphics[width=.12\textwidth]{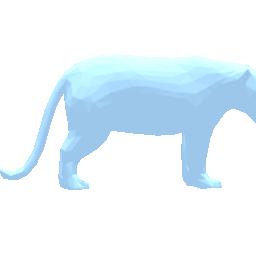}}~
 \subfloat{\includegraphics[width=.12\textwidth]{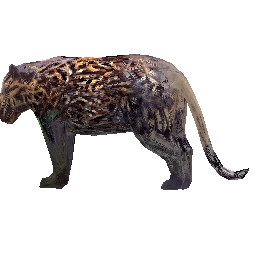}}~\\ 
 \vspace{-0.3cm}
 \caption{\textbf{3D reconstructions} We show the input image, the reconstructed shape from the predicted and a novel view and finally the predicted texture.\label{fig:6}}
 \end{centering}
 \end{figure*}

In Figures~\ref{fig:2}-\ref{fig:6} we provide a wide collection of 3D reconstructions for several highly articulated objects. We also include a collection of videos as part of the supplementary material with reconstructions of several classes. Our common visualization setup is input image, 3D reconstruction from the predicted viewpoint and a different one and finally the textured mesh reconstruction.

\subsection{Failure Cases}
\begin{figure*}[t]
 \captionsetup[subfigure]{labelformat=empty, position=top}
 \begin{centering}
 \setcounter{subfigure}{0}
 \subfloat{\includegraphics[width=.12\textwidth]{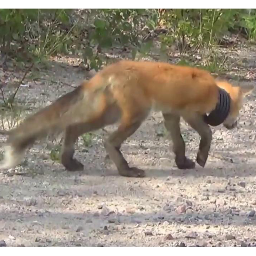}}~
 \subfloat{\includegraphics[width=.12\textwidth]{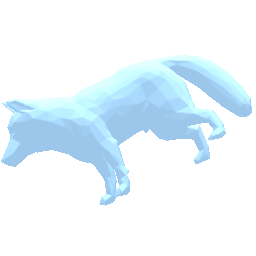}}~
 \subfloat{\includegraphics[width=.12\textwidth]{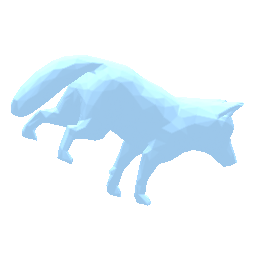}}~
 \subfloat{\includegraphics[width=.12\textwidth]{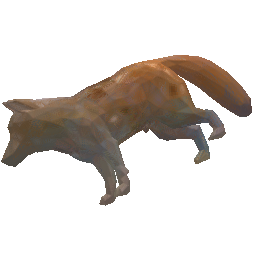}}~
 \subfloat{\includegraphics[width=.12\textwidth]{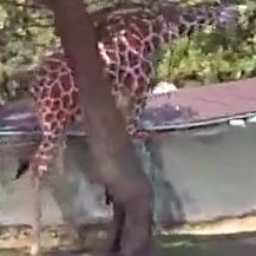}}~
 \subfloat{\includegraphics[width=.12\textwidth]{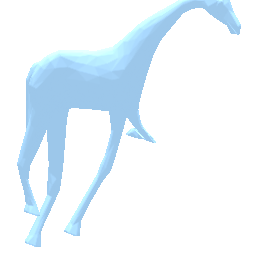}}~
 \subfloat{\includegraphics[width=.12\textwidth]{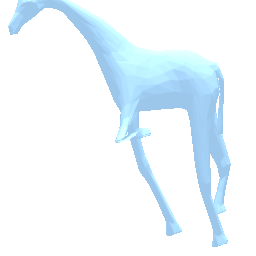}}~
 \subfloat{\includegraphics[width=.12\textwidth]{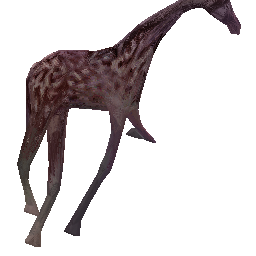}}~\\ 
 \vspace{-0.35cm}
 \subfloat{\includegraphics[width=.12\textwidth]{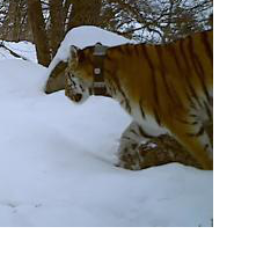}}~
 \subfloat{\includegraphics[width=.12\textwidth]{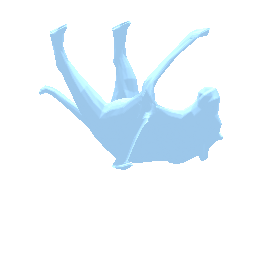}}~
 \subfloat{\includegraphics[width=.12\textwidth]{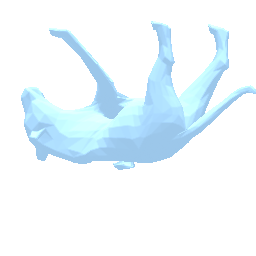}}~
 \subfloat{\includegraphics[width=.12\textwidth]{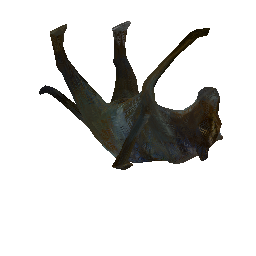}}~
 \subfloat{\includegraphics[width=.12\textwidth]{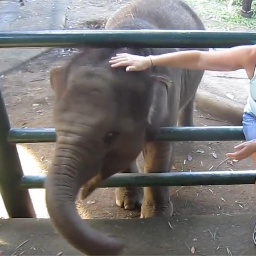}}~
 \subfloat{\includegraphics[width=.12\textwidth]{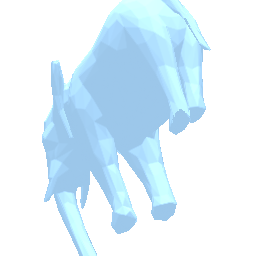}}~
 \subfloat{\includegraphics[width=.12\textwidth]{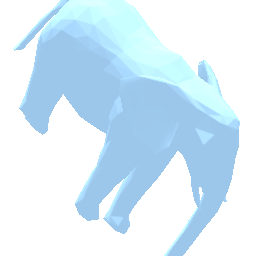}}~
 \subfloat{\includegraphics[width=.12\textwidth]{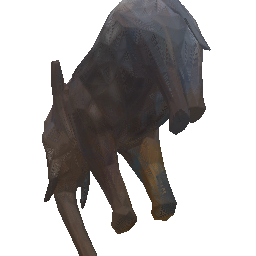}}~\\ 
 \vspace{-0.35cm}
 \subfloat{\includegraphics[width=.12\textwidth]{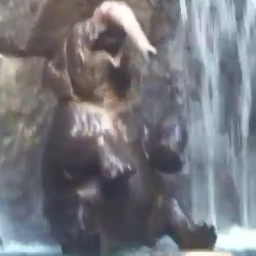}}~
 \subfloat{\includegraphics[width=.12\textwidth]{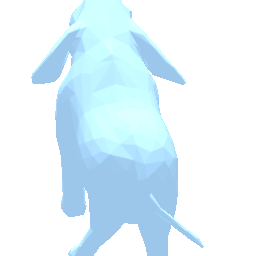}}~
 \subfloat{\includegraphics[width=.12\textwidth]{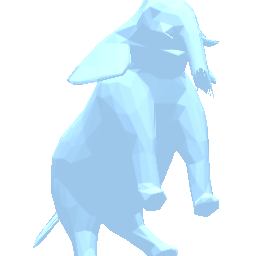}}~
 \subfloat{\includegraphics[width=.12\textwidth]{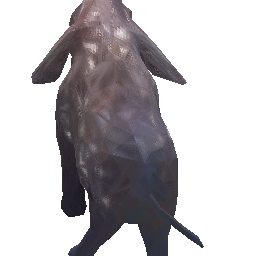}}~
 \subfloat{\includegraphics[width=.12\textwidth]{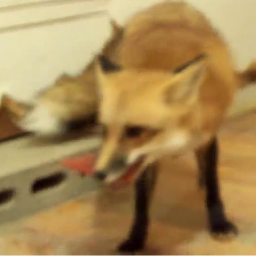}}~
 \subfloat{\includegraphics[width=.12\textwidth]{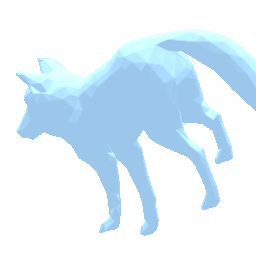}}~
 \subfloat{\includegraphics[width=.12\textwidth]{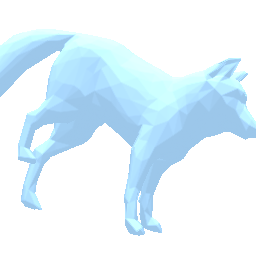}}~
 \subfloat{\includegraphics[width=.12\textwidth]{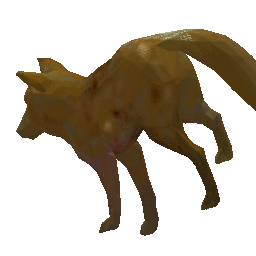}}~\\ 

 \caption{\textbf{Failure Cases:} We visualize some failure modes of our method. The columns present the input image, 3D reconstruction from the predicted viewpoint and a different one and the predicted texture .\label{fig:1}}
 \end{centering}
 \end{figure*}

We visualize some failure cases of the proposed method in Figure~\ref{fig:1}. Common failure cases are related to the inability to predict a good camera pose and the inference of simplistic textures.
 
{\small
\bibliographystyle{ieee_fullname}
\bibliography{egbib}
}